%% file: icml2026.tex
\documentclass{article}

\usepackage[table,xcdraw]{xcolor}
\usepackage{microtype}
\usepackage{subcaption}
\usepackage{tabularx}

\usepackage{hhline}
\usepackage{makecell}
\usepackage{tabularray}
\usepackage{boldline}

\usepackage{enumitem}
\usepackage{textgreek}
\usepackage{hyperref}
\usepackage{url}
\usepackage{tcolorbox}
\usepackage{booktabs}
\usepackage{threeparttable}
\usepackage{amsmath}
\usepackage[utf8]{inputenc}

\newtcolorbox{promptbox}[1]{
  colback=gray!5,
  colframe=black!70,
  coltitle=white,
  title=\textbf{#1},
  boxrule=1pt,
  arc=2mm,
  fontupper=\small,
  left=5pt, right=5pt, top=5pt, bottom=5pt,
  before skip=2pt,
  after skip=10pt,
}

\usepackage{hyperref}

\usepackage[accepted]{icml2026}

\usepackage{amsmath}
\usepackage{amssymb}
\usepackage{mathtools}
\usepackage{amsthm}

\usepackage[capitalize,noabbrev]{cleveref}

\theoremstyle{plain}

\theoremstyle{definition}

\theoremstyle{remark}

\usepackage{multirow}
\usepackage{graphicx}
\usepackage[textsize=tiny]{todonotes}
\usepackage{xspace}

\newcommand{\our}{\textsc{BioArc}\xspace}

\icmltitlerunning{\our}

\begin{document}

\twocolumn[
\icmltitle{\our: Discovering Optimal Neural Architectures \texorpdfstring{\\}{ } for Biological Foundation Models}
  \icmlsetsymbol{equal}{*}
  \icmlsetsymbol{corresponding}{$\dagger$}

  \begin{icmlauthorlist}
    \icmlauthor{Yi Fang}{vtcs}
    \icmlauthor{Haoran Xu}{vtcs}
    \icmlauthor{Jiaxin Han}{cmu}
    \icmlauthor{Sirui Ding}{stanford}
    \icmlauthor{Yizhi Wang}{vtece}
    \icmlauthor{Yue Wang}{vtece}
    \icmlauthor{Xuan Wang}{corresponding,vtcs}
  \end{icmlauthorlist}

  \icmlaffiliation{vtcs}{Department of Computer Science, Virginia Tech, Blacksburg, VA, USA}
  \icmlaffiliation{vtece}{Department of Electrical and Computer Engineering, Virginia Tech, Blacksburg, VA, USA}
  \icmlaffiliation{cmu}{Department of Computer Science, Carnegie Mellon University, Pittsburgh, PA, USA}
  \icmlaffiliation{stanford}{Department of Biomedical Data Science, Stanford University, Stanford, CA, USA}

  \icmlcorrespondingauthor{Xuan Wang}{xuanw@vt.edu}
  \icmlkeywords{Foundation Model, Architecture, DNA, Protein, Neural Architecture Search}

  \vskip 0.3in
]

\printAffiliationsAndNotice{}

\input{sections/00_abstract}
\input{sections/01_introduction}  
\input{sections/02_related_work}

\input{sections/03_methodology}
\input{sections/04_experiments}
\input{sections/05_conclusion}

\section*{Acknowledgements}
This research is sponsored by NSF \#2442253, Commonwealth Cyber Initiative, and generous gifts from Nvidia, Cisco, and the Amazon-Virginia Tech Initiative. This research used the Delta system at the National Center for Supercomputing Applications [award OAC 2005572] through allocation [NAIRR240202] from the Advanced Cyberinfrastructure Coordination Ecosystem: Services \& Support (ACCESS) program, which is supported by National Science Foundation grants \#2138259, \#2138286, \#2138307, \#2137603, and \#2138296. This research is also partially supported by allocation [NAIRR240439].

\section*{Impact Statement}
This paper presents work whose goal is to advance the field of biological foundation model. There are many potential societal consequences of our work, none of which we feel must be specifically highlighted here.

\bibliography{icml2026}
\bibliographystyle{icml2026}

\newpage
\appendix
\onecolumn
\input{sections/06_appendix}

\end{document}

%% file: sections/00_abstract.tex
\begin{abstract} 
Foundation models have revolutionized AI, yet biological applications often repurpose general architectures without accounting for the intrinsic structural and functional properties of distinct modalities, such as genomic and proteomic sequences. Consequently, these architectures lack the inductive biases required to capture the complex ``grammars" inherent to biological data, resulting in suboptimal performance. To address this, we introduce \our, a framework utilizing Neural Architecture Search (NAS) to shift from intuition-driven design to automated data-driven discovery. Unlike standard NAS restricted to homogeneous spaces, \our navigates a heterogeneous space for open-ended composition of architectural blocks. By systematically analyzing the interplay between architecture, tokenization, and training across modalities, \our identifies novel hybrid architectures that surpass state-of-the-art models while being up to 25x smaller. We distill these findings into empirical design principles and validate their biological relevance, demonstrating how our designs hierarchically capture the underlying biological grammar. Additionally, we introduce an agentic framework to predict optimal architectures for new tasks. Overall, \our provides a data-driven methodology for developing the next generation of efficient biological foundation models and task-specific networks.
\end{abstract}

%% file: sections/01_introduction.tex
\section{Introduction}
The advent of foundation models, large-scale neural networks pretrained on vast amounts of data, has catalyzed a profound revolution in artificial intelligence (AI). In general domains such as natural language processing (NLP)~\cite{devlin2019bertpretrainingdeepbidirectional,openai2023gpt,touvron2023llamaopenefficientfoundation} and computer vision (CV)~\cite{dosovitskiy2020image,liu2024visual}, foundation models built upon architectures such as Transformers~\cite{vaswani2017attention} and Diffusion Models~\cite{ho2020denoisingdiffusionprobabilisticmodels} have demonstrated unprecedented capabilities, reshaping research and applications. This success has spurred a wave of interest in applying this paradigm to biology, a trend driven by the increasing availability of massive datasets. The promise is to leverage large-scale data to create powerful biological foundation models~\cite{xiao2025proteinlargelanguagemodels} that can learn the underlying grammar of genetic or protein sequences, accelerating discovery in drug development, synthetic biology, and personalized medicine.

However, directly applying architectures from general AI domains presents numerous limitations in biology. Most current biological foundation models~\cite{zhou2024dnabert2efficientfoundationmodel, doi:10.1073/pnas.2016239118,xiao2025proteinlargelanguagemodels} are built upon Transformers~\cite{vaswani2017attention}, which was originally designed for human language and not for the complex ``grammar'' unique to biological data, making it suboptimal for biological applications. Specifically, biological sequences present a dual challenge: they require processing extremely long contexts (e.g., whole genomes)~\cite{Wang_2017} that incur prohibitive computational costs for standard Transformers, while simultaneously demanding the capture of precise local structural motifs~\cite{Greener2022guide}, which is an inductive bias not inherently prioritized by global attention mechanisms. The limitations of repurposing general architectures for biological sequences necessitate a fundamental inquiry: what constitutes an optimal neural architecture for biological foundation models?

Answering this question is non-trivial due to three distinct technical challenges. \textbf{First,} architectural innovation in this domain is hampered by a lack of established guiding principles. This presents a stark contrast to a field like NLP, where language is a human-generated system with rules and structures we inherently understand. Biological information, on the other hand, is designed by nature. Its underlying principles, such as the physicochemical laws governing protein design~\cite{marcos2017principles, dill2012protein}, are yet to be fully discovered and remain largely unknown. Consequently, unlike the NLP field, there is no universally acknowledged architecture that consistently excels across the board, forcing a reliance on manual, intuition-driven design that is inefficient~\cite{Sapoval2022,Eisenstein2024foundation,Vishniakov2024.12.18.628606}. \textbf{Second}, while automated search methods like Neural Architecture Search (NAS), standard methods typically focuses on \textit{micro-level} optimizations within homogeneous spaces, such as tuning layer counts or filter sizes in purely convolutional networks. 
Even domain-specific hybrid efforts, such as GenomeNet-Architect~\cite{gunduz2024optimized}, primarily rely on parameterizing fixed design templates distilled from existing models such as DeepVirFinder~\cite{ren2020identifying}, rather than exploring the open-ended combination of computational paradigms. 
Without a strategy to construct representative search spaces, such methods are confined to optimizing known backbones, failing to uncover novel topologies within the heterogeneous landscape.
\textbf{Finally,} adding to the complexity is the deep entanglement of architecture, data preprocessing, and training strategy. The efficacy of a given architecture is highly sensitive to choices in tokenization and optimization schemes, implying that architectural choices cannot be effectively designed, developed, and evaluated in isolation.

To address these challenges, we introduce \textbf{\our}, a framework that shifts the paradigm from inefficient manual design to automated data-driven discovery. By navigating a heterogeneous search space composed of five distinct operation primitives, \textbf{\our} enables the open-ended composition of complementary inductive biases. 
\textbf{Our contributions are as followed: }
\textbf{First}, we discover novel hybrid architectures that achieve state-of-the-art performance across diverse biological tasks while being up to $25\times$ smaller than existing foundation models. 
\textbf{Second}, serving as a unified testbed, \textbf{\our} disentangles the complex interplay between architecture, tokenization, and optimization, providing actionable design guidelines for future biological modeling. 
\textbf{In addition}, through fine-grained interpretability analysis, we demonstrate that the discovered topologies hierarchically capture known biological regulatory mechanisms. 
\textbf{Finally}, bridging analysis and automation, we introduce an agentic framework capable of efficiently identifying optimal architectures for unseen tasks without exhaustive search.


%% file: sections/02_related_work.tex
\section{Related Work}
\paragraph{Foundation Models for Biological Data}
Inspired by NLP and CV, biological foundation models learn patterns from vast unlabeled sequences via self-supervised pretraining. In proteomics, transformer-based models like AlphaFold~\cite{jumper2021highly} revolutionized structure prediction. Scaling to the whole-genome level, Evo-1 and Evo-2~\cite{doi:10.1126/science.ado9336, Brixi2025.02.18.638918} enables generative design. For sequence representation, models such as PathoLM~\cite{dip2024patholmidentifyingpathogenicitydna}, Nucleotide Transformer~\cite{Dalla-Torre2023.01.11.523679}, Gena-LM~\cite{Fishman2023.06.12.544594}, and VQDNA~\cite{li2024vqdnaunleashingpowervector} target DNA; ProtBert~\cite{Elnaggar2020.07.12.199554} and ESM-1 and ESM-2~\cite{doi:10.1073/pnas.2016239118, doi:10.1126/science.ade2574} focus on proteins; while RNAErnie~\cite{Wang2024Multi} and RNABERT~\cite{10.1093/nargab/lqac012} address RNA. Expanding modalities, Geneformer~\cite{Theodoris2023Transfer} and CellFM~\cite{Zeng2025CellFM} targets single-cell gene expression value sequence.

\paragraph{Hybrid Architecture}
Beyond pure sequence models, hybrid architectures combine distinct neural mechanisms to leverage complementary strengths. In protein science, integrating GNNs with attention captures both sequential context and geometric constraints~\cite{jumper2021highly}. Similarly, for nucleic acids, merging CNNs with Transformers enables efficient processing of high-resolution sequences; CNNs encode local elements and reduce length, allowing Transformers to model distal interactions such as enhancer-promoter contacts~\cite{Avsec2021Effective,10.1093/bib/bbae577}.

\paragraph{Neural Architecture Search}
Neural Architecture Search (NAS) automates architecture design. While early RL~\cite{zoph2017neuralarchitecturesearchreinforcement} and evolutionary~\cite{real2017largescaleevolutionimageclassifiers} methods were computationally expensive, efficient one-shot techniques~\cite{liu2019dartsdifferentiablearchitecturesearch} have since enabled landmarks~\cite{tan2020efficientnetrethinkingmodelscaling,deng2009imagenet}.BossNAS~\cite{li2021bossnasexploringhybridcnntransformers} demonstrate a hybrid CNN-Transformer network but limited two-way combinations. However, biological applications remain limited. Prior works focus mostly on single architectures~\cite{Zhang2021.02.25.432960, zhang2021automated} or specific sequence tasks~\cite{SIVANGI202263}. Even GenomeNet-Architect~\cite{gunduz2024optimized}, which combines CNNs and RNNs, is still constrained by 2 module types.

\paragraph{Architecture Prediction}
To mitigate the high computational cost of NAS, architecture predictors serve as efficient surrogates, estimating performance directly from structural properties~\cite{ying2019nasbench101}. Trained on pre-evaluated architecture-performance pairs, these models range from simple regressors to Graph Neural Networks capable of encoding complex topologies~\cite{dudziak2021brpnaspredictionbasednasusing}. However, their generalization is inherently limited by the scope of training tasks~\cite{duan2021transnasbench101improvingtransferabilitygeneralizability}, restricting transferability to novel domains without significant retraining.

%% file: sections/03_methodology.tex
\section{\our Framework}
This section introduces \our, a framework for discovering optimal heterogeneous backbones. First we introduce how to define the search space and how to construct the supernet in Section~\ref{sec:search_space}. Then we elaborate on the training of supernet in Section~\ref{sec:supernet pretraining}. Last we discuss how to evaluation and rank all architectures in Section~\ref{sec:eval}.
While demonstrated on DNA and protein, the framework generalizes to other modalities such as RNA and single-cell data.

\subsection{Search Space \& Supernet}
\label{sec:search_space}
As discussed, previous NAS methods have focused on micro-level optimizations within homogeneous spaces or on no more than 2-way hybrid architectures. To capture different inductive biases in biological sequence, a heterogeneous ensemble of multiple architecture blocks is needed.
\paragraph{Basic Blocks} 
We begin by introducing the basic blocks defined as $\mathcal{M}$, 
including a variety of architectures widely used for biological data
such as \textbf{CNN}~\cite{krizhevsky2012imagenet}, \textbf{LSTM}~\cite{hochreiter1997long}, \textbf{Transformer}~\cite{vaswani2017attention}, \textbf{Mamba}~\cite{gu2024mamba} and \textbf{Hyena}~\cite{poli2023hyena}. The diversity in block types 
ensures our search can cover different inductive biases. Details of these blocks can be found in 
Appendix~\ref{Appendix:Module_types}.

\begin{figure*}[h!]
    \centering
    \includegraphics[width=0.95\linewidth]{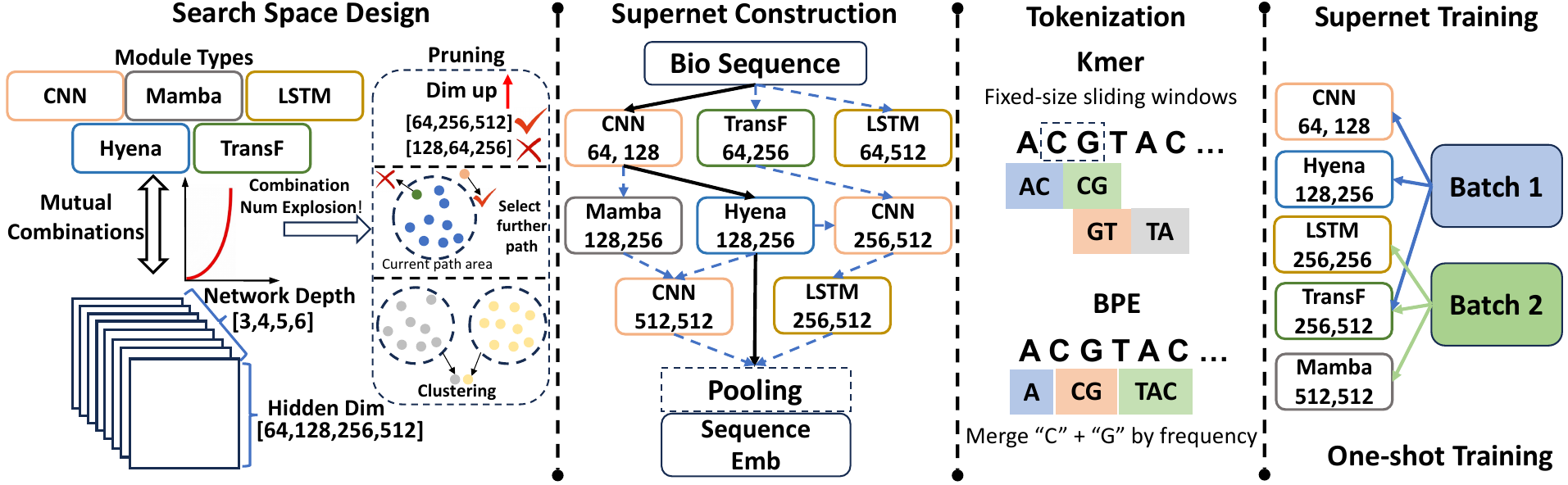}
    \caption{This figure provides an overview of the four core stages of the BioArc framework. (1) Search Space Design: Defines diverse block types, network depths, and hidden dimensions, using pruning and clustering strategies to manage the combinatorial explosion. (2) Supernet Construction: Encodes the vast search space into a single, weight-sharing Supernet, with each path representing a candidate architecture. (3)  Supernet Training: Adopts a one-shot methodology, sampling and optimizing different paths across batches for efficiency.}
    \label{fig:pipeline}
\end{figure*}

\paragraph{Paths formed by blocks}
Each path $a \in \mathcal{A}$ is a sequence of basic blocks $(l_1, l_2, \dots, l_d)$, where its structure is determined by three variations:
\begin{enumerate}[leftmargin=*]
    \item \textit{\textbf{Network Depth}} ($\mathbf{d}$): The number of blocks in a path, chosen from a predefined set of possible depths, $\mathcal{D} = \{D_{\min}, \dots, D_{\max}\}$.
    
    \item \textit{\textbf{Block Type}} ($\mathbf{m}$): A tuple  $\mathbf{m} = (m_1, m_2, \dots, m_d)$ of length $d$, where each element $m_i \in \mathcal{M}$ specifies the block for the $i$-th layer.
    
    \item \textit{\textbf{Hidden Dimension}} ($\mathbf{h}$): A tuple $\mathbf{h} = (h_1, h_2, \dots, h_d)$ of length $d$, where each $h_i$ is a hidden dimension selected from a set of possible widths $\mathcal{H}$.
\end{enumerate}

A complete path $a$ is constructed from a tuple $(\mathbf{h}, \mathbf{m})$ of length $d$. For $i \in \{1, \dots, d\}$, the $i$-th layer $l_i$ is of type $m_i$ and maps an input of dimension $h_{i-1}$ to an output of dimension $h_i$. The initial dimension $h_0$ is the embedding dimension which is a fixed hyperparameter.

\paragraph{Search Space Design}
Let $\mathcal{C}_{\text{dim}}^{(d)}$ and $\mathcal{C}_{\text{type}}^{(d)}$ be the sets of all permissible dimension and block-type configurations for a given depth $d$, respectively. The total search space $\mathcal{A}$ is then the union of the Cartesian products of these configuration sets across all possible depths:
\begin{equation}
\mathcal{A} = \bigcup_{d \in \mathcal{D}} \left( \mathcal{C}_{\text{dim}}^{(d)} \times \mathcal{C}_{\text{type}}^{(d)} \right)
\label{eq:model_space}
\end{equation}
The combinatorial nature of the configuration sets yields millions of candidates. In standard homogeneous search spaces, where variations are purely parametric (e.g., filter sizes), uniform random sampling is often sufficient to cover the manifold. However, in our heterogeneous space, distinct topological families (e.g., pure CNNs vs. Hyena-Transformer hybrids) are unevenly distributed. Naive sampling would likely overfit to structurally redundant patterns while neglecting rare but potent hybrid combinations. To ensure computational tractability while preserving structural diversity, we employ a Representative Sampling Strategy to extract a concise subset of 360 architectures. This process systematically prunes topologically redundant paths via \textbf{distance-based filtering}, utilizing a metric based on log-transformed dimensions. Furthermore, we incorporate \textbf{a monotonic width constraint with the maximum dimension fixed at the last layer}. This intentionally engineers a structural bias where wider, parameter-heavy blocks appear in a greater number of valid paths. Consequently, this design ensures parameter-heavy blocks implicitly receive higher sampling frequency during training, aligning training intensity with block complexity to facilitate convergence. Finally, we use \textbf{K-Means clustering on the one-hot encoded architecture vectors} to control the number of paths in the search space, selecting the centroids of these clusters. Details are provided in Appendix~\ref{Appendix:pruning}.

\paragraph{Supernet Construction}
We employ a weight-sharing supernet approach\cite{Bender2018UnderstandingAS}. The key insight is that many architectures within $\mathcal{A}$ reuse identical building blocks. For example, a specific Transformer block with a 256 input dimension and a 512 output dimension is a component in different paths. 
Therefore, the supernet is a single large network that effectively contains all candidate architectures as paths.
First, we identify the set of all unique blocks used in paths across the entire search space:
\begin{equation}
    L = \{l \mid \exists a \in \mathcal{A}, l \in a\}
\end{equation}
Each unique block is defined by its type (e.g., CNN) and its dimensions. Other hyperparameters (e.g., CNN kernel sizes, Transformer attention heads) are fixed. The supernet weight set $W$ is the union of all unique block weights:
\begin{equation}
    W = \bigcup_{l \in L} W_l
\end{equation}
where $W_l$ are the trainable parameters for a unique layer $l$. The weights for any specific path denoted as $w(a)$ are a subset of these shared weights:
\begin{equation}
    w(a) = \{ W_{l_i} \mid l_i \in a \} \subseteq W
\end{equation}

\subsection{Supernet Pretraining}
\label{sec:supernet pretraining}
\paragraph{One-Shot Supernet Pretraining}
We adopt a random path sampling mechanism inspired by the Single Path One-Shot approach~\cite{guo2020singlepathoneshotneural}. In each forward pass, we activate a single path through the supernet. We sample paths from a uniform distribution. 
Leveraging the structural bias engineered into our search space, this simple sampling strategy naturally fulfills the higher data requirements of complex blocks.
Consequently, the stochasticity of single-path updates acts strictly as a regularization mechanism. By decoupling specific block interactions, it prevents co-adaptation~\cite{lan2020albertlitebertselfsupervised}, ensuring that shared blocks learn robust, independent features rather than over-specializing to specific neighbors.

\textbf{Self-Supervised Pretraining Objectives} The supernet denoted $\mathcal{A}$ is trained by stochastically sampling architectures from the search space. In each training step, a path $a$ is sampled, and the supernet's shared weights $W$ are updated by minimizing self-supervised loss $\mathcal{L}$. The overall training objective is to optimize the expectation of this loss over the distribution of all possible architectures:
\begin{equation}
\label{eq:pretrain}
    \min_{W} \mathbb{E}_{a \sim \mathcal{A}} [\mathcal{L}(\mathcal{A}(X; w(a)))]
\end{equation}
where $X$ is the input data, $a \sim \mathcal{A}$ denotes a path uniformly sampled from the search space $\mathcal{A}$ defined in Section~\ref{sec:search_space}, and $w(a) \subset W$ are the weights corresponding to the block in path $a$. This ensures every operation in the search space is trained equally to provide a stable performance proxy.

To ensure the supernet learns meaningful and generalizable features and to investigate the impact of different pretraining tasks, the supernet is pretrained using one of three distinct self-supervised learning objectives: Masked Modeling (MM), Contrastive Learning (CL), and Next Token Prediction (NTP). In Equation~\ref{eq:pretrain}, $\mathcal{L}$ refers to the specific loss function corresponding to the chosen pretraining strategy. This pretraining step is performed on a large corpus of unlabeled biological data before the supernet is used for evaluation on specific downstream tasks. The detailed mathematical formulations and biological motivations for MM, CL, and NTP are provided in Appendix~\ref{Appendix:training_strategies}.

\subsection{Architecture Evaluation and Ranking}
\label{sec:eval}
\paragraph{Evaluation Protocol}
\label{Section:Evaluation Protocol}

To fairly assess the performance of sampled architectures, \textbf{we optimize each path independently, rather than fine-tuning the entire supernet as a whole}. We adopt this approach for two primary reasons. First, the computational cost is tractable due to the significantly smaller size of downstream datasets compared to the pretraining corpus. Second, decoupling the weights is essential to eliminate interference, thereby ensuring \textbf{optimal performance and accurate ranking}. This isolation is particularly crucial because the supernet is trained to prevent \textbf{co-adaptation}~\cite{Bender2018UnderstandingAS, guo2020singlepathoneshotneural}, resulting in shared parameters that act as a \textbf{compromise} for expected performance rather than being optimal for any specific path. We conduct this independent assessment using two paradigms distinct by their weight initialization:

\paragraph{Finetuning}
The parameters of the sampled architecture $a$, denoted as $w(a)$, are initialized by inheriting the corresponding weights from the pretrained supernet $W$. The architecture is finetuned to minimize the task-specific loss:
\begin{equation}
    \min_{w(a)} \mathcal{L}_{\text{task}}(\mathcal{A}(X_{\text{task}}; w(a))), \: \text{s.t.} \: w(a)_{\text{init}} \subset W.
\end{equation}

\paragraph{Architecture Ranking Strategy}
We rank the sampled architectures $\mathbb{S}$ based on their aggregated performance across the set of downstream tasks $\mathcal{T}$. Since tasks utilize different metrics with varying scales and statistical variances (e.g., Accuracy vs. RMSE), a direct summation would disproportionately favor tasks with wider numerical distributions. To address this heterogeneity and ensure that each task contributes equally to the final ranking, we employ Z-score normalization to standardize the performance metrics before aggregation. We define the unified score for an architecture $a$ as the mean of these standardized metrics:
\begin{equation}
    \text{Score}(a) = \frac{1}{|\mathcal{T}|} \sum_{t \in \mathcal{T}} s_t \cdot \frac{P_t(a) - \mu_t}{\sigma_t},
\end{equation}
where $P_t(a)$ denotes the raw performance of architecture $a$ on task $t$. The terms $\mu_t$ and $\sigma_t$ represent the mean and standard deviation of the performance scores for task $t$ across all sampled candidates. The coefficient $s_t \in \{1, -1\}$ aligns the optimization direction, taking $1$ for metrics where higher is better (e.g., Accuracy) and $-1$ for error-based metrics (e.g., RMSE). Finally, the top candidates $\mathbb{A}_{\text{top}}$ are identified by selecting the $k$ architectures with the highest $\text{Score}(a)$.

\paragraph{Optimal Architecture as Foundation Model}
Once the top ranking architecture $a^* \in \mathcal{A}$ is identified, we randomly initialize its parameters, denoted as $\theta$, and pretrain it:
\begin{equation}
    \min_{\theta} \mathcal{L}(a^*(X; \theta)).
\end{equation}
To evaluate the foundation model's transferability, we subsequently finetune the pretrained parameters $\theta$ on each specific downstream task $t$:
\begin{equation}
    \min_{w} \mathcal{L}_{\text{task}}(\mathcal{A}(X_{\text{task}}; w)) \: \text{s.t.} \: w_{\text{init}} \leftarrow \theta
\end{equation}

%% file: sections/04_experiments.tex
\section{Experiment}
Having detailed our construction of \our framework, we aim to answer the following research questions with our experiments. 
\textbf{RQ1:} Can finetuned architectures from \our pretrained supernet outperform baseline methods?
\textbf{RQ2:} What are the common architectural properties that characterize high-performing models? 
\textbf{RQ3:} Does the optimal architecture identified by \our serve as a effective backbone for biological foundation models?
\textbf{RQ4:} How does the interplay between architecture, tokenization, and pretraining objectives impact downstream performance? 
\textbf{RQ5:} Do the discovered hybrid architectures hierarchically capture the underlying biological grammar and regulatory mechanisms?

\subsection{Settings and baselines}

\paragraph{DNA} For pretraining, we use the full human reference genome \textbf{GRCh38}. For downstream evaluation, we employ the human subset of the \textbf{GUE benchmark}~\cite{zhou2024dnabert2efficientfoundationmodel}, comprising 12 datasets to cover a variety of tasks. Detailed descriptions are provided in Appendix~\ref{Appendix:DNA Tasks}.

\paragraph{Protein} For pretraining, we use a 10\% randomly sampled \textbf{UniRef50}~\cite{Suzek2007} subset, containing 72.1M representative sequences clustered at 50\% identity. For evaluation, we select 6 tasks from the \textbf{PEER} benchmark~\cite{xu2022peercomprehensivemultitaskbenchmark}, covering protein function, structure, and interactions. Detailed descriptions are provided in Appendix~\ref{Appendix:Protein Tasks}.

\paragraph{Baselines} For DNA, we use Nucleotide Transformer~\cite{Dalla-Torre2023.01.11.523679}, DNABERT-2~\cite{zhou2024dnabert2efficientfoundationmodel}, and VQDNA~\cite{li2024vqdnaunleashingpowervector} as baselines. These pretrained large foundation models all share human DNA GRCh38 in pretraining with finetuing on additional multi-species genome datasets. For protein, we use ESM-1 and ESM-2~\cite{doi:10.1073/pnas.2016239118,doi:10.1126/science.ade2574} and ProtBert~\cite{Elnaggar2020.07.12.199554} as baselines. ProtBert is pretrained on BFD~\cite{steinegger2019protein} with 393B amino acids and ESM-1b is pretrained on UniParc~\cite{10.1093/nar/gkm895} with 86B amino acids. More details of training setting could be found in Appendix~\ref{Appendix:settings}. 

\paragraph{Tokenization} To thoroughly investigate the impact of tokenization on performance, we conducted experiments using different tokenizers, including k-mer and Byte Pair Encoding (BPE)~\cite{sennrich2016neuralmachinetranslationrare}. \textit{\textbf{K-mer tokenizers}} operate by dividing a sequence into overlapping or non-overlapping substrings of a fixed length, denoted as $k$. \textit{\textbf{BPE}} is a subword tokenization strategy that begins with a vocabulary of individual characters and iteratively merges the most frequently occurring adjacent pairs of tokens into a new, single token.

\subsection{Main Results}
\label{sec:main-results}
In this section, we address RQ1, RQ2 and RQ3. \textbf{For RQ1}, we observe the following. \textbf{First, architectures discovered by \our are highly competitive, often outperforming larger, pretrained models with much smaller model size.} As detailed in Table~\ref{table:DNA-results}, \our models \textbf{achieve superior performance across all DNA tasks}, in some cases by a substantial margin. \textbf{For the protein tasks} in Table~\ref{table:Protein-results}, we disentangle architecture from pretraining budget through a controlled comparison in which \our 8M and a reimplemented ESM-2 8M are pretrained under \emph{identical} conditions (full UniRef50, 50K steps), making architecture the only variable. \textbf{Under this matched setting, \our 8M outperforms ESM-2 8M on all six PEER tasks, including the structural Fold task (20.75 vs.\ 18.25).} This indicates that the apparent weakness of compact models on structural tasks is \emph{not} an architectural ceiling but a consequence of pretraining scale: the residual gap to the released ESM-2 8M (Fold 22.14) tracks its $\sim$10$\times$ larger pretraining budget (full UniRef50, 500K steps, 32$\times$A100 vs.\ 50K steps, 4$\times$A100). We attribute \our's advantage to task-specific inductive bias: its hybrid topology adapts to the intrinsic signal density of biological sequences, maximizing data efficiency under a fixed budget, as further evidenced by its consistently lower pretraining loss at every checkpoint (Appendix~\ref{appendix:protein results}). \textbf{The difference in performance highlights a key principle for biology model design that optimal architecture depends on the nature of the task.} For those demanding intricate feature extraction from the sequence itself, a specialized architecture is key. For others that rely on vast, implicit priors learned from data at scale, massive pretraining is essential.

\newcommand{\best}[1]{{\textbf{#1}}}
\newcommand{\secbest}[1]{\underline{#1}}
\begin{table*}[t!]
\centering
\caption{Performance on different DNA tasks. \best{Bold} and \secbest{underline} indicate the best and second-best result respectively. We report the average size of top-performing architectures across tasks. More results on Protein can be found in Appendix~\ref{appendix:protein results}.}
\label{table:DNA-results}
\resizebox{\linewidth}{!}{
  \begin{tabular}{l |c|c |c c c c c | c c c | c c c | c}
    \hlineB{2}
    \multirow{2}{*}{\textbf{Method}}
    & \multirow{2}{*}{\textbf{\#Param}}
    & \multirow{2}{*}{\textbf{\#Data}}
    & \multicolumn{5}{c|}{\textbf{TFP}} 
    & \multicolumn{3}{c|}{\textbf{PD}} 
    & \multicolumn{3}{c|}{\textbf{CPD}} 
    & \multirow{2}{*}{\parbox{2.5cm}{\centering\textbf{SSP} \\ \textbf{Reconstruction}}} \\
    \cline{4-14}
    & & &\textbf{0} & \textbf{1} & \textbf{2} & \textbf{3} & \textbf{4} 
    & \textbf{all} & \textbf{notata} & \textbf{tata} 
    & \textbf{all} & \textbf{notata} & \textbf{tata} 
    & \\
    \hline
    HyenaDNA (one-hot)&6.6M &3.1B & 62.30 & 67.86 & 46.85 & 41.78 & 61.23 & 47.38 & 52.24 & 5.34 & 36.95 & 35.38 & 72.87 & 72.67 \\
    NT-2500M-1000g (6-mer)&2500M &20.5T & 66.31 & 68.30 & 58.70 & 49.08 & 67.59 & 90.95 & 93.07 & 75.80 & 67.39 & 67.46 & 69.66 & 85.78 \\
    DNABERT-2 (BPE) &117M &262B & 71.99 & 76.06 & 66.52 & 58.54 & 77.43 & 86.77 & 94.27 & 71.59 & 69.37 & 68.04 & 74.17 & 84.99 \\
    VQDNA (HRQ) &103M&262B & 72.48 & 76.43 & 66.85 & 58.92 & 78.10 & 90.75 & 94.48 & 74.52 & 71.02 & 70.58 & 78.50 & 89.53 \\ 
    \hline
    \our (mask-ft)  & 4.89M &3.1B & \best{84.80} & \secbest{86.00} & \secbest{85.80} & \secbest{77.10} & \secbest{89.20} & 92.62 & \best{96.55} & 83.20 & \best{83.60} & \best{85.43} & \secbest{89.40} & \secbest{90.84} \\
    \our (con-ft)   &3.28M &3.1B & \best{84.80} & \best{86.10} & \best{86.50} & \best{77.50} & \best{89.30} & \best{93.12} & 96.25 & \best{83.69} & 83.53 & \secbest{84.66} & \best{90.05}  & \best{91.06} \\
    \our (ntp-ft) &6.58M &3.1B & 84.20 & 85.90 & 82.80 & 75.50 & 89.10 & 91.82 & 96.25 & 76.02 & 82.26 & 84.38 & 74.55 & 83.76 \\
    \hlineB{2}
  \end{tabular}
}
\end{table*}

\begin{table*}[t!]
\centering
\caption{Performance on different protein tasks. The \textbf{top block} reports reference baselines from released checkpoints (much larger pretraining budgets). The \textbf{bottom block} is a controlled comparison in which \our and ESM-2 are pretrained under identical conditions (full UniRef50, 50K steps) to isolate architecture; \best{bold} marks the better model within this controlled comparison. ESM-2 8M (official) is the released 8M checkpoint (full UniRef50, 500K steps), shown for reference. The corresponding pretraining loss is reported in Appendix~\ref{appendix:protein results}.}
\label{table:Protein-results}
\resizebox{\linewidth}{!}{
  \begin{tabular}{l | c| c | c | c | c | c | c | c}
  \hlineB{2}
  \textbf{Method} & \textbf{\#Param} & \textbf{\#Data} & \textbf{Solubility} & \textbf{HumanPPI} & \textbf{PPIAffinity} & \textbf{Fold} & \textbf{Subcellular} & \textbf{Binary} \\
  \hline
  ProtBert &419.9M &393B &68.15 & 77.32 & 2.195 & 16.94 & 76.53 & 91.32 \\
  ESM-1b &652.4M &86B &70.23 & 78.17 & 2.281 & 28.17 & 78.13 & 92.40 \\
  ESM2-1b &652.4M &86B &74.13 & 80.49  & 2.134 & 32.31 & 80.45 & 92.79 \\
  ESM-2 8M (official) &8M &86B &73.48 & 80.16 & 3.098 & 22.14  & 71.47 & 91.25 \\
  \hline
  \multicolumn{9}{l}{\textit{Controlled comparison (identical pretraining: full UniRef50, 50K steps)}}\\
  ESM-2 8M (reimpl.) &8M &50K steps & 71.84 & 74.68 & 3.567 & 18.25 & 70.36 & 90.68 \\
  \our 8M &8M &50K steps & \best{73.29} & \best{76.79} & \best{2.756} & \best{20.75} & \best{72.77} & \best{91.82} \\
  \hlineB{2}
  \end{tabular}
}
\end{table*}

\begin{figure*}[t!]
    \centering
    \includegraphics[width=0.95\linewidth]{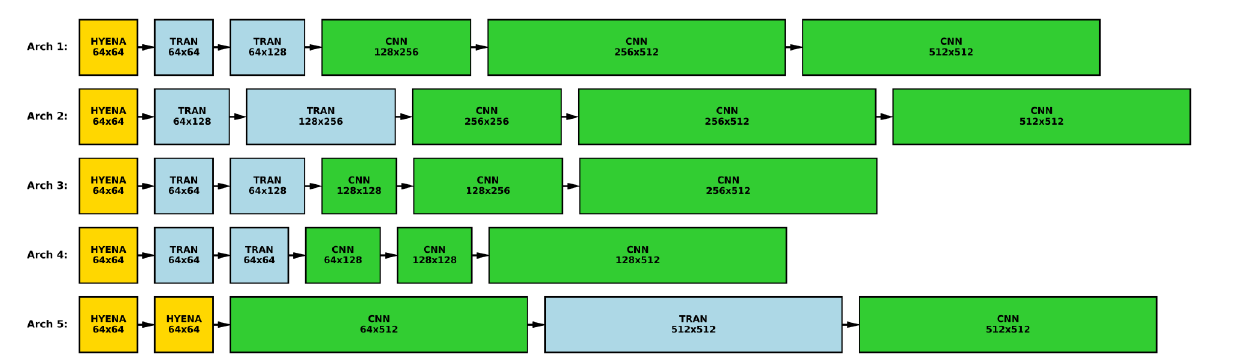}
    \caption{This figure illustrates the top five performing DNA model architectures (Arch 1-5), identified by averaging their performance across the different tasks. Results on protein is shown in Appendix~\ref{Appendix:architecture pattern}.} 
    \label{fig:trend}
\end{figure*}

\textbf{For RQ2}, we observe three key findings. \textbf{First, top-performing architectures converge on specific structural patterns.} For DNA tasks, as shown in Figure~\ref{fig:trend}, optimal models typically initiate with Hyena block to first captures long-range dependencies, with Transformer blocks in the middle to model complex contextual relationships and CNN blocks at the end to extract critical features. \textbf{Second, while optimal architectures are task-specific, they exhibit significant commonalities within task families.} For instance, the top 10\% architectures for Transcription Factor Prediction-3 and 4 share 98.0\% similarity as shown in Figure~\ref{fig:heatmap} and Appendix~\ref{appendix:architecture similarity}.

\begin{figure*}[h!]
    \centering
    \includegraphics[width=0.95\linewidth]{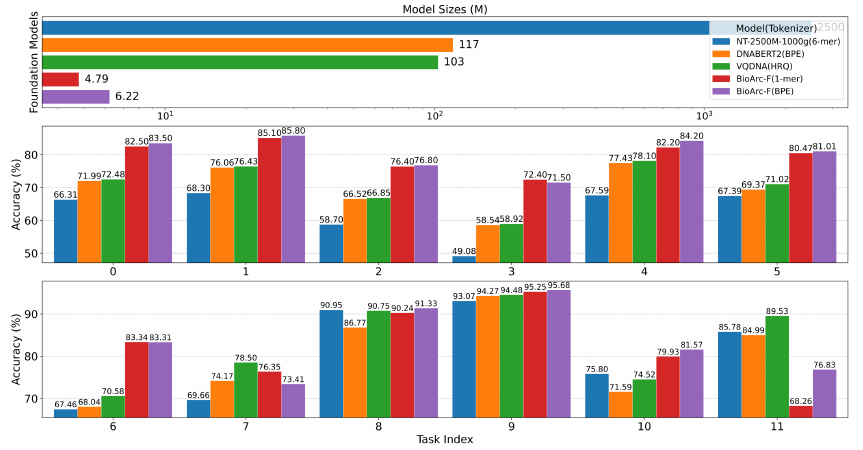}
    \vspace{-10pt}
    \caption{Performance of \our-Discovered Architecture as a Foundation Model Backbone, noted as \textbf{\our-F}. The architecture is selected by the top average performance across all tasks and pretrained for 1/10 training steps of baselines. } 
    \label{fig:foundation}
\end{figure*}

To answer RQ3, we selected a overall top-performing \our architecture and conduct pretraining similar as DNABERT2. As shown in Figure~\ref{fig:foundation}, \textbf{with only 1/20 model size and 1/10 training steps, \our-based foundation model outperforms the human-designed architectures on downstream DNA tasks.} To further evaluate its scalability, we extended model depth to 10 layers by doubling the number of Hyena and Transformer layers and fixed hidden dimension to 1024 across all layers to expand model width. As the results shown in Appendix~\ref{appendix:scaling up}, \textbf{scaling up model size further improves the performance.} \our-F is also competitive with recent genomic foundation models (Caduceus~\cite{schiff2024caduceus} and GENERator~\cite{wu2025generator}) on the Nucleotide Transformer benchmark and the Genomic Benchmarks~\cite{gresova2023genomic}, achieving the best result on all histone-mark and enhancer tasks despite being substantially smaller (Appendix~\ref{appendix:nt-gb-results}). This result, combined with our findings for RQ1, indicates that \our is highly effective for identifying both high-performing specialized architectures and robust backbones for foundational models in biology. Detailed training settings could be found in Appendix~\ref{Appendix:settings} and analysis of the computational costs is provided in Appendix~\ref{appendix:computational-cost}.

\begin{figure*}[h!]
    \centering
    \includegraphics[width=1.0\textwidth]{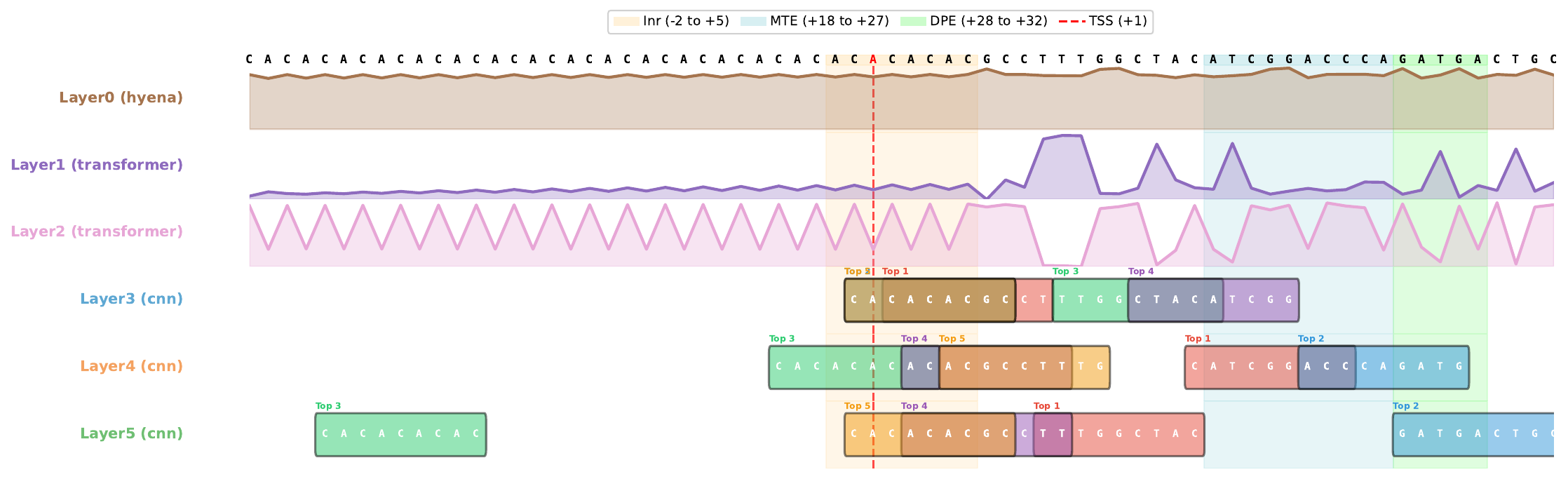}
    \caption{\textbf{Hierarchical Interpretation of Promoter Grammar.} 
    \textbf{Layer 0 (Hyena):} Extracts global genomic context via continuous activations. 
    \textbf{Layers 1-2 (Transformer):} Maps spatial syntax; Layer 1 anchors downstream elements while Layer 2 identifies the TSS functional boundary via pattern transition. 
    \textbf{Layers 3-5 (CNN):} Acts as a morphological decoder, transitioning from localized motif detection (L3-4) to multi-aspect integration of the Inr-MTE-DPE complex (L5). Details are provided in Appendix~\ref{appendix:interpretation}.}
    \label{fig:layer_interpretation}
\end{figure*}

\subsection{Study on Interplay of Training and Tokenization}
In this section, \textbf{we address RQ4} by analyzing how critical design choices including training strategy and tokenization interact with architectural topology.

\begin{figure}[h!]
    \centering
    \includegraphics[width=0.95\linewidth]{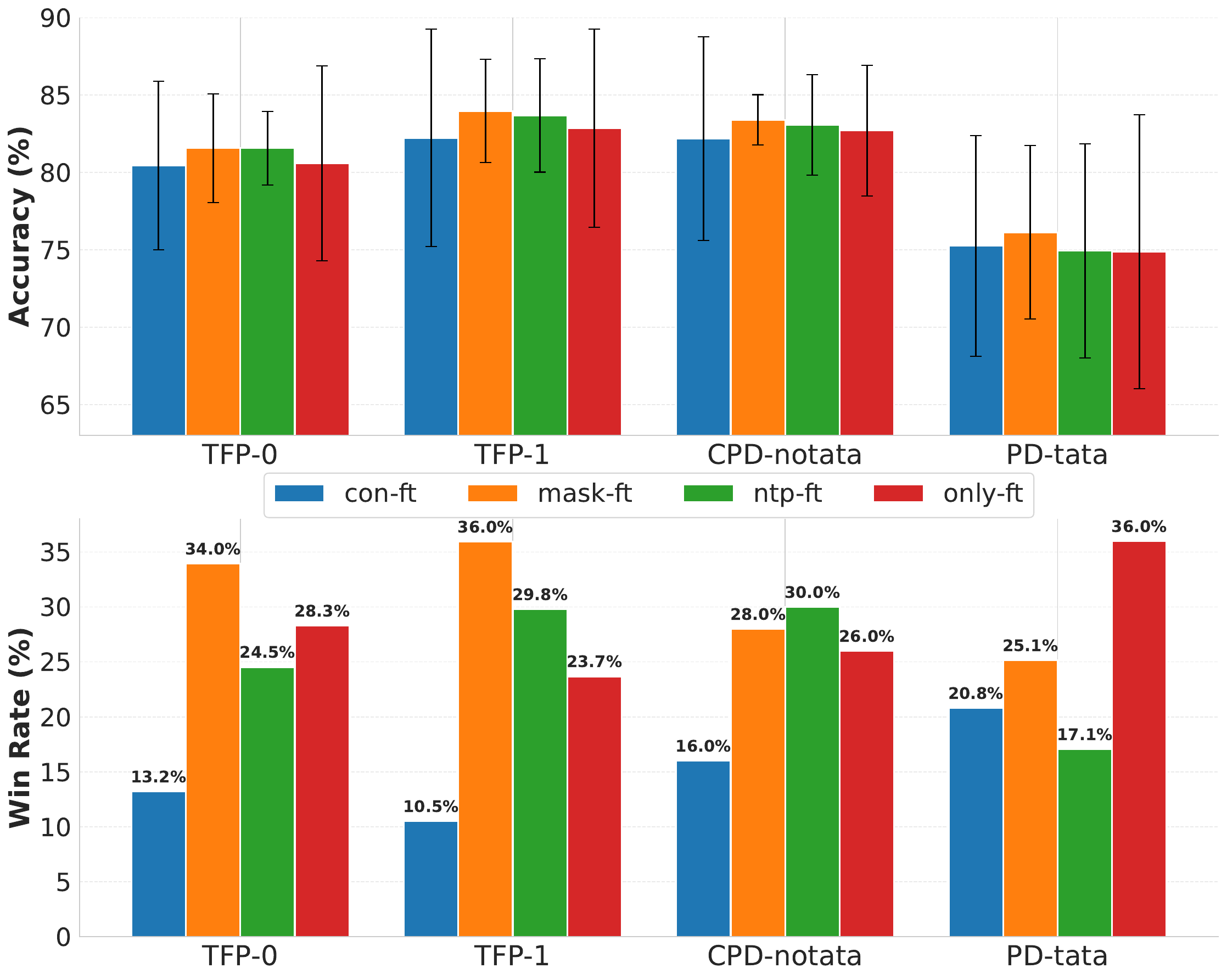}
    \caption{Performance of different training strategies on DNA. In the top panel, each cell shows the mean performance ± standard deviation across all architectures. In the bottom panel, each bar shows the percentage of the total 360 architectures that chose that training strategy to yield the best performance. More results on Protein could be found in Appendix~\ref{Appendix:training results}.}
    \label{fig:training}
\end{figure}

\textbf{Impact of Training Strategies} \quad We summarize the training strategy results in Figure~\ref{fig:training}. 
\textbf{First, pretraining does not guarantee gains.} Training from scratch (only-ft), where we train each architecture path from random initialization on the downstream task, achieves the highest Win Rate on the PD-tata task (36.0\%) and notably outperforms contrastive pretraining in other scenarios. 
\textbf{Secondly, no pretraining strategy dominates.} While masked modeling (mask-ft) generally yields the highest performance across TFP datasets, Next Token Prediction (ntp-ft) proves superior on the CPD-notata task. This suggests that the optimal training strategy is highly dependent on the downstream data characteristics. \textbf{Thirdly, contrastive learning exhibits instability and high variance.} 
In addition to generally lower Win Rates, \textit{con-ft} demonstrates significant performance fluctuations compared to reconstruction-based methods (\textit{mask-ft} and \textit{ntp-ft}). 
This instability suggests that contrastive objectives may require longer training schedule to converge effectively, whereas generative signals offer better sample efficiency within limited training budgets.

\vspace{-5pt}

\begin{figure}[t!]
    \centering
    \vspace{-10pt}
    \begin{subfigure}[b]{0.85\linewidth}
        \centering
        \includegraphics[width=\linewidth]{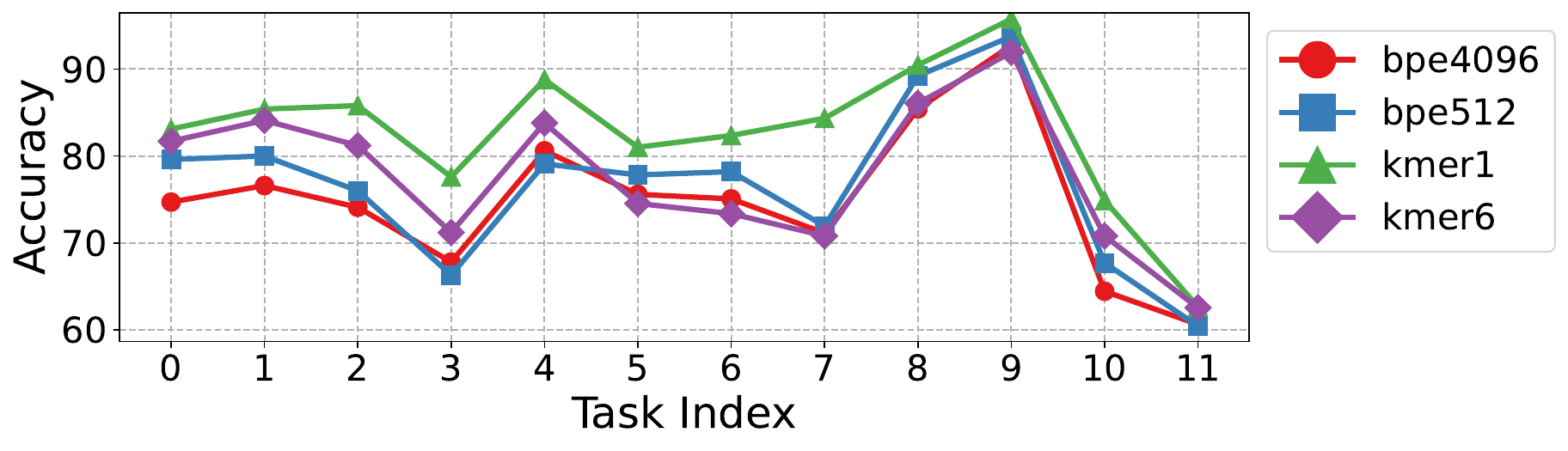}
        \vspace{-15pt}
        \caption{CNN}
        \label{subfig:CNN}
    \end{subfigure}
    
    \vspace{10pt}
    \begin{subfigure}[b]{0.85\linewidth}
        \centering
        \includegraphics[width=\linewidth]{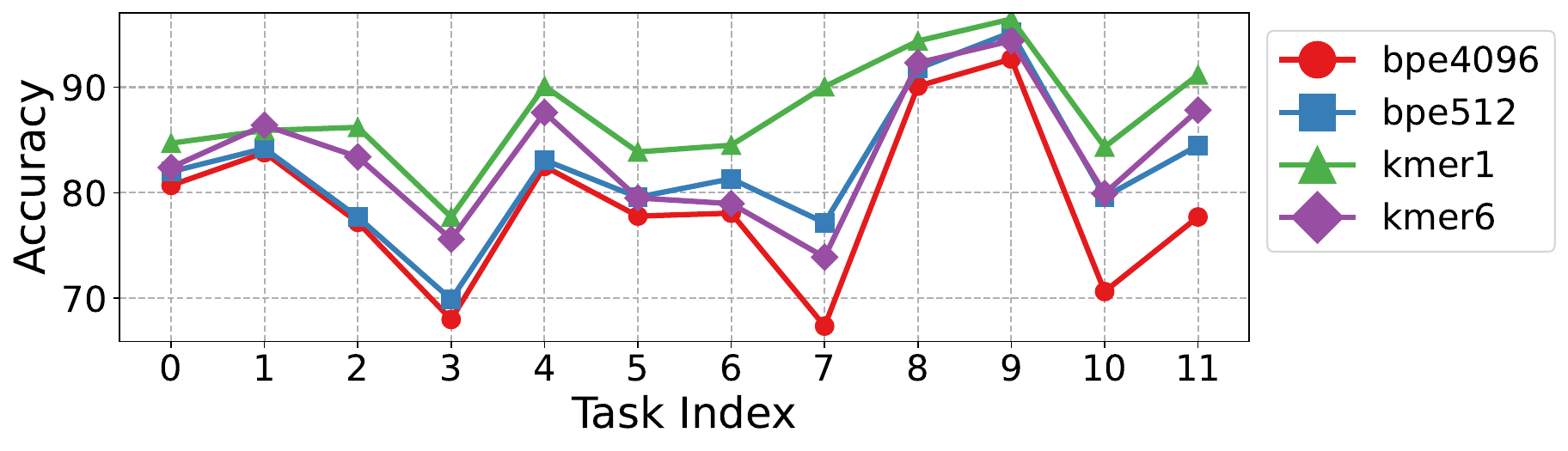}
        \vspace{-15pt}
        \caption{Hybrid architectures}
        \label{subfig:Mix}
    \end{subfigure}
    
    \vspace{-5pt} 
    \caption{Effect of different tokenization on various architectures training from scratch on the DNA tasks. Detailed analysis across modalities and training strategies is extended in Appendix~\ref{Appendix:tokenizers results}.}
    \label{fig:comparison_of_tokenizer}
    \vspace{-10pt}
\end{figure}

\textbf{Impact of Tokenization}  \quad \textbf{First, we find that the optimal tokenizer choice is highly architecture-dependent.} As shown in Figure~\ref{fig:comparison_of_tokenizer}, the Transformer-based architecture performs best with a 6-mer tokenizer, while the CNN-based model achieves its highest scores with a 1-mer tokenizer. This reveals a deep interplay between architecture and tokenization, demanding their co-optimization. \textbf{Second, the tokenizer's effectiveness interacts with the training strategy.} As shown in Figure~\ref{fig:foundation}, \our-F with BPE tokenizer outperforms 1-mer tokenizer, which contradicts the situation in task-specific models. This suggests that pretraining propels complex tokenization, while simple tokenization methods are less reliant on pretraining. More comprehensive results are available in the Appendix~\ref{Appendix:tokenizers results}.

\subsection{Interpretability Case Study on Core Promoter}
\label{sec:interpretability}
Visualizing layer-specific activations on the Core Promoter task (Figure~\ref{fig:layer_interpretation}) reveals that our model spontaneously learns a hierarchy mirroring biological complexity. Specifically, the \textbf{Global Context (Hyena, L0)} establishes broad genomic context via continuous receptive fields, forming a semantic foundation. Subsequently, \textbf{Spatial Syntax (Transformer, L1-2)} layers anchor attention on the Transcription Start Site (TSS), acting as syntactic validators. Crucially, at the \textbf{Motif Synergy (CNN, L3-5)} level, filters simultaneously activating on Inr and DPE elements confirm the unsupervised rediscovery of the \textbf{Inr+DPE synergy rule}~\cite{burke1996dpe}. These findings suggest a correspondence between our model's learned hierarchy and established biological principles, highlighting the interpretability of the hybrid design. Details can be found in Appendix~\ref{appendix:interpretation}.

\subsection{More Analysis}
\textbf{Comparison with Pure Architectures} Hybrid architectures consistently outperform single-module baselines, validating the effectiveness of our search space design. Details can be found in Appendix~\ref{Appendix:hybrid and single architecture comparison}. 

\textbf{Layer-wise Contribution Analysis} To demonstrate the synergistic efficacy of our hybrid modules, we conducted probing experiments by attaching classifiers to the output of each layer. We find that the model achieves consistent monotonic improvements in accuracy layer-by-layer, verifying that the specific sequence of diverse modules collaboratively refines biological representations rather than relying on a single dominant layer. Details can be found in Appendix~\ref{appendix:layer-contribution}.

\textbf{Architecture Depth Analysis} To verify the scalability of our discovered patterns, we investigated the impact of network depth on transfer performance. We find that while deeper models generally offer higher potential capacity, performance is not solely determined by depth; however, once an optimal architectural pattern is identified, scaling up the depth consistently enhances robustness. Details can be found in Appendix~\ref{appendix:depth_analysis}.

\textbf{Performance-Parameter Analysis} To confirm that the superiority of our models stems from topological efficiency rather than mere capacity, we analyzed the relationship between parameter count and accuracy across the search space. We find that there is no positive correlation between model size and performance; instead, specific hybrid topologies (inductive biases) are the decisive factors for success, achieving high accuracy with significantly fewer parameters. Details can be found in Appendix~\ref{appendix:performance-parameter}.

\textbf{Architecture Ranking Correlation Analysis} To validate the reliability of our one-shot search strategy, we evaluated the consistency of architecture rankings. First, we observed a low correlation (Spearman's $\rho = 0.3222$) between the rankings derived from the \textbf{\textit{pretrained and finetuned supernet}} versus those \textbf{\textit{trained from scratch}}. Figure~\ref{fig:rank_correlation_1} indicates that finetuning architectures individually is essential. Second, rankings derived from the \textbf{\textit{pretrained supernet with individual finetuning}} exhibit a strong positive correlation (Spearman's $\rho = 0.7287$) with the \textbf{\textit{train-from-scratch}} baseline. As illustrated in Figure~\ref{fig:rank_correlation_2}, validating that the architectural rankings remain consistent regardless of the pretraining initialization. Further details are provided in Appendix~\ref{appendix:architecture-ranking-correlation}.

\textbf{Architecture Prediction.} To extend our method to unseen tasks without repetitive search, we leveraged the observation that similar tasks consistently favor similar optimal topologies. By benchmarking predictors based on traditional neural networks, standard LLMs, and autonomous agents, we find that the agent-based system achieves the highest performance in inferring optimal designs, effectively bridging the gap between offline search and real-world deployment. Details can be found in Appendix~\ref{appendix:architecture-prediction}.

%% file: sections/05_conclusion.tex
\section{Conclusion}
We introduce \our, a neural architecture search framework. By exploring a vast space, we identify high-performing foundational and task-specific architectures that often outperform larger established models. We also analyze how tokenization and training interact with architecture, clarifying their critical impact. Furthermore, fine-grained interpretability demonstrates that the discovered hybrid architectures capture biological grammar, validating our design's biological relevance. Ultimately, \our provides a foundational resource and principled methodology for creating the next generation of biology-tailored models.

%% file: sections/06_appendix.tex
\section{Appendix}

\subsection{Benchmark Details}
\subsubsection{DNA}
\label{Appendix:DNA Tasks}
\textbf{Transcription Factor Prediction (TFP, Task 0-4):} This task involves predicting transcription factor binding sites (TFBS) in the human genome using data derived from 690 ENCODE ChIP-seq experiments. 

\textbf{Core Promoter Detection (CPD, Task 5-7):} This task involves identifying the precise location of core promoter regions, which are essential for initiating gene transcription, focusing on
predicting the core promoter region only, the central region closest to the TSS and start codon. A much shorter context window (center -34 +35 bp around TSS) is provided, making this a more challenging task than proximal promoter prediction.

\textbf{Promoter Detection (PD, Task 8-10):} This task requires the model to identify human proximal promoter regions, which contain primary regulatory elements crucial for transcription initiation. We construct three dataset variations: TATA, non-TATA, and a combined All set. Positive samples are extracted from the Eukaryotic Promoter Database (EPDnew) (Dreos et al., 2013), covering the window from -249 to +50 bp relative to the Transcription Start Site (TSS). Negative control strategies differ by subset: the TATA dataset uses random non-promoter genomic sequences containing TATA motifs, while the non-TATA dataset utilizes randomly substituted sequences to represent the null class.

\textbf{Splice Site Prediction (SSP, Task 11):} This task focuses on splice site prediction in the human genome, a process vital for understanding protein diversity and genetic diseases. 

\subsubsection{Protein}
\label{Appendix:Protein Tasks}
\textbf{Solubility Prediction (Task 12):} a binary classification task that determines if a protein is soluble or insoluble. To ensure models can generalize to new and dissimilar proteins, the dataset is split so that the training proteins have less than 30\% sequence identity with the test proteins. This is a critical task because good solubility is an essential property for functional proteins, especially in pharmaceutical research and industry. The ultimate goal is to drive the development of more effective computational tools that can predict a protein's solubility based solely on its amino acid sequence.

\textbf{Human PPI Prediction (Task 13):} A binary classification task to determine whether a pair of human proteins will physically interact. The data is split into an 8:1:1 ratio for training, validation, and testing, and the task is designed to test how well models can generalize to dissimilar protein sequences. The work is significant because understanding the human protein interactome is vital for deciphering disease mechanisms. This task serves as a benchmark to drive the development of more effective machine learning models for PPI prediction.

\textbf{Fold Classification (Task 14):} A classification task that predicts the overall 3D structural shape (fold) of a protein from 1,195 possible categories. The model is specifically tested on its ability to perform remote homology detection, meaning it must recognize proteins with similar structures even if their sequences are very different. To achieve this, entire superfamilies of proteins are withheld from the training data and used only for testing. The goal is to automate fold classification using protein sequences, which is important for drug design and functional analysis because most known protein structures have not yet been manually classified.
 
\textbf{PPI Affinity Prediction (Task 15):} A regression task that estimates the binding strength between two proteins. Using the SKEMPI dataset, the data is split based on the number of mutations to test the model's generalization capabilities in a scenario mimicking multi-round protein engineering: Training set is wild-type proteins and mutants with less then 2 mutations. Validation set is mutants with 3 or 4 mutations. Test set is mutants with more then 4 mutations. The primary impact of this task is its direct application to protein binder design, where accurately predicting the binding strength of candidate molecules is crucial for developing new therapeutics and biotechnologies.

\textbf{Subcellular Location Prediction (Task 16):} A multiclass classification task involves predicting which of 10 possible locations a protein resides in within a cell. It is a multi-class classification problem where models are trained and tested using the DeepLoc dataset, which is specifically partitioned to evaluate performance on homologous proteins. The primary impact of this research is in drug discovery. Knowing a protein's location helps identify it as a potential drug target, and an accurate, high-throughput prediction tool can significantly accelerate this process.

\textbf{Binary Location Prediction (Task 17):} A simplified version of subcellular localization, framed as a binary classification problem designed to classify proteins as either membrane-bound or soluble, using a binary label like 0 or 1. The model is trained and tested on data from the DeepLoc dataset, with a key evaluation being its ability to generalize and make accurate predictions for homologous (structurally similar) proteins. The task is significant because it helps efficiently distinguish between soluble proteins, which are free-floating, and membrane-bound proteins, which are attached to cell membranes and can have important catalytic functions.

\subsection{Modules Designs}
\label{Appendix:Module_types}
In this section, we provide the detailed architectural specifications for the various building blocks employed in our framework. Table~\ref{tab:module_specs} presents a summary of these designs, outlining the core layers, normalization and activation mechanisms, as well as the specific hyperparameter settings (e.g., kernel sizes, initialization schemes) used in our experiments.
\begin{table}[h!]
    \centering
    \caption{Detailed Specifications of Architectural Modules}
    \label{tab:module_specs}
    \renewcommand{\arraystretch}{1.3} 
    \resizebox{\textwidth}{!}{%
    \begin{tabular}{c l l m{6cm}} 
        \toprule
        \textbf{Module} & \textbf{Core Layer} & \textbf{Norm / Activation} & \textbf{Key Hyperparameters} \\
        \midrule
        \textbf{CNN} & Conv1d ($D_{in} \to D_{out}$) & ReLU $\to$ BatchNorm1d & Kernel: $\{5, 9\}$, Padding: $\{2, 4\}$ \\
        \midrule
        \textbf{Hyena} & HyenaEncoderLayer & \emph{None} (Identity) & Proj: Linear ($D_{in} \to D_{out}$) optional; Model Dim: $D_{out}$ \\
        \midrule
        \textbf{Transformer} & TransformerEncoderLayer & LayerNorm (internal) & Heads: $D_{out}/64$; Input matched to $D_{out}$ \\
        \midrule
        \textbf{Mamba} & Mamba Layer & LayerNorm (pre-block) & Proj: Linear ($D_{in} \to D_{out}$) optional \\
        \midrule
        \textbf{LSTM} & LSTM ($1$ layer) & \makecell[l]{Sigmoid/Tanh\\(internal gates)} & Hidden: $D_{out}$, Dropout: 0.4, Init: Orthogonal (rec), Xavier (in), Bias 0 \\
        \bottomrule
    \end{tabular}
    }
\end{table}

\subsection{Search Space Pruning}
\label{Appendix:pruning}
The combinatorial nature of search space design, while comprehensive, leads to an exponential growth in the number of candidate architectures as the number of choices increases. To maintain a computationally tractable search space without sacrificing diversity, we employ three strategies to select a representative subset of configurations for both module types ($\mathcal{C}_{\text{type}}^{(d)}$) and hidden dimensions ($\mathcal{C}_{\text{dim}}^{(d)}$) at each depth $d \in \mathcal{D}$ (in Equation~\ref{eq:model_space}).

\paragraph{Rule-based pruning.} We enforces a monotonic non-decreasing width constraint (i.e., $h_i \geq h_{i-1}$ for $i \geq 1$) to encourage progressive feature extraction.

\paragraph{Distance-based pruning.} We first generate all possible valid dimension paths that satisfy the non-decreasing constraint. To ensure diversity from the outset, we employ a greedy selection algorithm. A candidate path is only added to our initial representative set if it is sufficiently distant from all previously selected paths. The distance between two dimension paths, $\mathbf{h}_a$ and $\mathbf{h}_b$, is measured as the Euclidean distance in a \textbf{log-transformed space}. Specifically, given full paths $\mathbf{h}'_a = (h_0, h_{1,a}, \dots, h_{d,a})$ and $\mathbf{h}'_b = (h_0, h_{1,b}, \dots, h_{d,b})$, where $h_0$ is the fixed embedding dimension, their distance is:
    \[
    D(\mathbf{h}'_a, \mathbf{h}'_b) = \sqrt{\sum_{i=0}^{d} \left(\log_2(h_{i,a}) - \log_2(h_{i,b})\right)^2}
    \]
    A candidate path $\mathbf{h}_{\text{cand}}$ is kept only if $D(\mathbf{h}'_{\text{cand}}, \mathbf{h}'_{\text{rep}}) \geq \tau$ for all paths $\mathbf{h}_{\text{rep}}$ already in the representative set, where $\tau$ is a predefined distance threshold. Using a log scale ensures that the selection is sensitive to relative changes in dimension (e.g., distinguishing between 64 and 128) rather than absolute differences.

\paragraph{Cluster-based pruning.}
We deploy clustering method in both module types and hidden dimension orders.
For a given depth $d$, the total number of possible module type configurations is $|\mathcal{M}|^d$. To reduce this number to a manageable target $k_d$, we perform clustering to identify a set of diverse and representative paths. The procedure is as follows:

\begin{enumerate}[leftmargin=*]
    \item \textbf{Vector Representation}: Each module type path $\mathbf{m} = (m_1, m_2, \dots, m_d)$ is first transformed into a numerical vector. We apply \textbf{one-hot encoding} to each module $m_i \in \mathcal{M}$, converting the categorical sequence into a high-dimensional numerical representation. This results in a flattened vector for each path.

    \item \textbf{K-Means Clustering}: We then apply the K-Means clustering algorithm to the set of all one-hot encoded vectors. The number of clusters is set to the desired number of representative configurations $k_d$. K-Means partitions the paths into $k_d$ clusters by minimizing the within-cluster sum of squares.

    \item \textbf{Centroid-based Selection}: The centroid of each cluster represents the mean of the paths within it. Since a centroid may not correspond to a valid, discrete module type path, we select the actual path from the dataset that is closest to each cluster centroid in terms of Euclidean distance. This ensures that our final set $\mathcal{C}_{\text{type}}^{(d)}$ consists of $k_d$ valid and diverse module type configurations.
\end{enumerate}

The search space for hidden dimension paths $\mathbf{h} = (h_1, h_2, \dots, h_d)$ is also vast. We apply the same pipeline on it, except changing the first encoding step.

By applying these structured reduction techniques to both the module type and hidden dimension configurations, we construct a final search space $\mathcal{A}$ that is both diverse and computationally feasible, allowing for an efficient yet comprehensive exploration of the architectural landscape.

\subsection{Training Strategies}
\label{Appendix:training_strategies}

\paragraph{Masked Modeling (MM)}
For a given input sequence $S = (x_1, x_2, \dots, x_L)$, we randomly mask a fraction of its tokens to create a corrupted sequence $\tilde{S}$. The supernet is then tasked with predicting the original tokens for the masked positions based on the contextual information provided by the unmasked tokens. This process compels the model to learn intricate local dependencies and contextual relationships within biological sequences.
$$
\mathcal{L}_{\text{MM}} = - \sum_{t \in \mathcal{M}} \log p(x_t \mid \tilde{S})
$$
where $\mathcal{M}$ is the set of indices of the masked tokens.

\paragraph{Contrastive Learning (CL)}
This approach aims to learn an embedding space where representations of similar sequences are pulled closer together, while those of dissimilar sequences are pushed apart. For a batch of sequences $\{S_i\}_{i=1}^B$, we generate a ``positive'' pair for each sequence $S_i$ by applying data augmentations, resulting in encoded representations $z_i$ and $z_j$. All other sequences in the batch are treated as ``negative'' examples. The model is trained to maximize the cosine similarity of positive pairs while minimizing it for negative pairs:
$$
\mathcal{L}_{\text{CL}} = - \log \frac{\exp(\text{sim}(z_i, z_j) / \tau)}{\sum_{k \neq i} \exp(\text{sim}(z_i, z_k) / \tau)}
$$
where $\text{sim}(\cdot)$ denotes cosine similarity, and $\tau$ is a temperature hyperparameter.
\paragraph{Next Token Prediction (NTP)}
For a given input sequence $S = (x_1, x_2, \dots, x_L)$, this objective leverages the auto-regressive nature of biological sequences. The supernet is tasked with predicting the next token $x_{t+1}$ based on the preceding context in $S$. This process compels the model to learn the sequential grammar and causal dependencies:
$$
\mathcal{L}_{\text{NTP}} = - \sum_{t=1}^{L-1} \log p(x_{t+1} \mid x_1, \dots, x_t)
$$
where $L$ is the length of the sequence $S$.

\subsection{Experiments Settings}
\subsubsection{Default training settings}
\label{Appendix:settings}
\textbf{For supernet pretraining,} we used NNI framework implementation~\cite{Microsoft_Neural_Network_Intelligence_2021}. We use one A100 80G for the pretraining for 10 epoch.

\textbf{For overall best architecture pretraining in Figure~\ref{fig:foundation}}, we use AdamW optimizer with $\beta_1 = 0.9$, $\beta_2 = 0.98$, $\epsilon = 1e-6$ and weight decay of $1e-5$. The learning rate linearly increases from 0 to $5e-4$ during the first 3000 steps while linearly decreasing to 0 in the last 47000 steps, which is 1/10 comparing to DNABERT2. We use mask modeling as training strategy.

\label{Appendix: prediction split}
\textbf{For neural network architecture predictor,}  we use Adam optimizer with learning rate = 1e-3, weight decay = 1e-5, batch size = 32, num epochs = 50, dropout rate = 0.3. We design two settings by if there is very similar tasks in the training set and test set. For example, Transcription Factor Prediction 0-4 are considered as similar tasks, Subcellular Location Prediction and Binary Location Prediction are considered as similar tasks as well.
For supervise setting, we use tasks $[0,1,4,5,7,8,9,11,12,13,14,16]$ as training set and tasks $[2,3,6,10,15,17]$ as test set. There are 6446 data points in the training set and 3210 data points in test set. For transfer setting, we use tasks $[5,6,7,8,9,10,12,13,14]$ as training set and tasks $[2,11,16,17]$ as test set. There are 4793 data points in the training set and 2156 data points in test set. We use all-MiniLM-L6-v2 to encode task descriptions in to task embeddings. We set the prediction number K to be 3.

\textbf{For LLM and Agent system architecture predictor}, the default LLM we used is GPT-4o. We also conduct ablation study on the model choices. We set the prediction number K to be 3.

\textbf{For task-specific training}, all the hyperparameters are listed in the follow Table~\ref{tab:training_params}.
\begin{table}[t!]
\centering
\caption{Hyperparameters for Individual Task Training}
\label{tab:training_params}
\resizebox{\textwidth}{!}{%
\begin{tabular}{|l|c|c|c|c|c|} 
\hline
\textbf{Modality} & \textbf{Learning Rate} & \textbf{Batch Size} & \textbf{Epoch} & \textbf{Num Warmup Steps} & \textbf{Weight Decay} \\ \hline
DNA & $3 \times 10^{-5}$ & 32 & 3(task 0-5) 10 (task 6-11) & 50 & 0.01 \\ \hline
Protein & $5 \times 10^{-5}$ & 32 & 5 & 50 & 0.01 \\ 
\hline
\end{tabular}%
}
\end{table}

\subsubsection{Evaluation Setting Details}
\label{Appendix:eval}
In this section, we clarify the specific evaluation protocols corresponding to the model variants reported in Table~\ref{table:DNA-results} and Table~\ref{table:Protein-results}. We employed two distinct weight initialization paradigms as outlined in Section~\ref{sec:eval} including \textit{Training from Scratch} and \textit{Pretrain and Finetune}.

\paragraph{Pretrain and Finetune.}
The \our variants involve initializing the architecture with weights inherited from a supernet that has been pretrained on large-scale biological data, followed by task-specific fine-tuning. These variants are distinguished by the self-supervised learning objective used during the supernet pretraining phase. Specifically, \textbf{\our (mask-ft)} utilizes weights from a supernet pretrained via Masked Modeling; \textbf{\our (con-ft)} employs weights from a supernet pretrained using Contrastive Learning; and \textbf{\our (ntp-ft)} initializes from a supernet trained with the Next Token Prediction objective.

\subsubsection{Computational Cost}
\label{appendix:computational-cost}
We report the computational costs for both supernet pretraining and task-specific finetuning. All experiments were conducted on  NVIDIA A100 80G GPUs.
For the \textbf{supernet pretraining}, the computational cost varies by objective. Specifically, mask modeling requires approximately 1.9 hours per epoch, while contrastive learning takes roughly 4.1 hours per epoch. \textbf{Training the foundation model} via mask modeling took a total of 10.9 hours. 
For \textbf{task-specific training}, we evaluate the total time required to finetune all 360 architectures across 18 downstream tasks, shown in Table~\ref{tab:finetune_cost}.

\begin{table*}[h]
\centering
\caption{Computational cost for task-specific finetuning across 18 tasks.}
\label{tab:finetune_cost}
\setlength{\tabcolsep}{3.5pt}
\begin{tabular}{@{}ccccccccc@{}}
\toprule
\textbf{0} & \textbf{1} & \textbf{2} & \textbf{3} & \textbf{4} & \textbf{5} & \textbf{6} & \textbf{7} & \textbf{8} \\
1h 40m & 1h 30m & 57m & 1h 20m & 58m & 2h 30m & 2h 13m & 14m & 5h 50m \\ \midrule
\textbf{9} & \textbf{10} & \textbf{11} & \textbf{12} & \textbf{13} & \textbf{14} & \textbf{15} & \textbf{16} & \textbf{17} \\
5h 12m & 36m & 5h 56m & 6h 43m & 72h 20m & 7h 25m & 85h 15m & 6h 02m & 9h 51m \\ \bottomrule
\end{tabular}

\end{table*}

\subsection{Main experiment results}
\subsubsection{Top Architectures Pattern}
\begin{figure}[t!]
    \centering
    \includegraphics[width=1\linewidth]{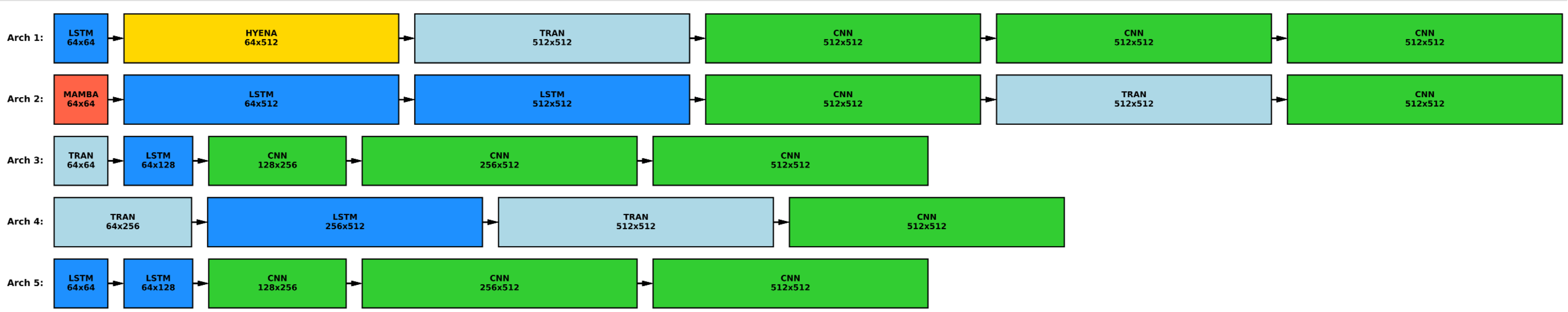}
    \caption{This figure illustrates the top five performing protein model architectures (Arch 1-5).}
    \label{Appendix:protein pattern}
\end{figure}

Similar to DNA, top architectures for protein demonstrate an observable pattern with LSTMs and Transformers in the initial stages followed by CNNs, as shown in Figure~\ref{Appendix:protein pattern}. Moreover, this pattern for protein is not consistent with the pattern for DNA, validating our intuition that architecture should be customized for different modalities.
\label{Appendix:architecture pattern}

\subsubsection{Architecture Similarity}
\label{appendix:architecture similarity}
To illustrate our encoding scheme, consider a sequential architecture consisting of three layers: a CNN layer (64 to 128 channels), followed by a Transformer block (128 to 256 dim), and an LSTM layer (256 to 512 dim).

\paragraph{Adjacency Matrix.}
We represent the connectivity between these $N=3$ layers using an adjacency matrix $\mathbf{A} \in \{0,1\}^{N \times N}$. For a strictly sequential architecture, this forms a super-diagonal matrix:
$$
\mathbf{A} = 
\begin{bmatrix}
0 & 1 & 0 \\
0 & 0 & 1 \\
0 & 0 & 0
\end{bmatrix}
$$

\paragraph{Node Feature Matrix.}
The features for each layer are encoded in $\mathbf{X} \in \mathbb{R}^{N \times F}$. Each row corresponds to a layer, concatenating the one-hot encoded operation type (first 5 columns) and the normalized input/output dimensions (last 2 columns):
$$
\mathbf{X} = 
\bordermatrix{
~ & \text{\tiny Type} & & & & & \text{\tiny In} & \text{\tiny Out} \cr
\text{\tiny CNN} & 1 & 0 & 0 & 0 & 0 & 0 & 0.14 \cr
\text{\tiny Trans} & 0 & 0 & 1 & 0 & 0 & 0.14 & 0.43 \cr
\text{\tiny LSTM} & 0 & 0 & 0 & 1 & 0 & 0.43 & 1.00
}
$$
\noindent Dimensions are normalized by a factor of $D_{\text{max}}=512$. The rows correspond to [CNN, 64, 128], [Transformer, 128, 256], and [LSTM, 256, 512], respectively.

\begin{figure}[h!]
    \centering
    \includegraphics[width=1\linewidth]{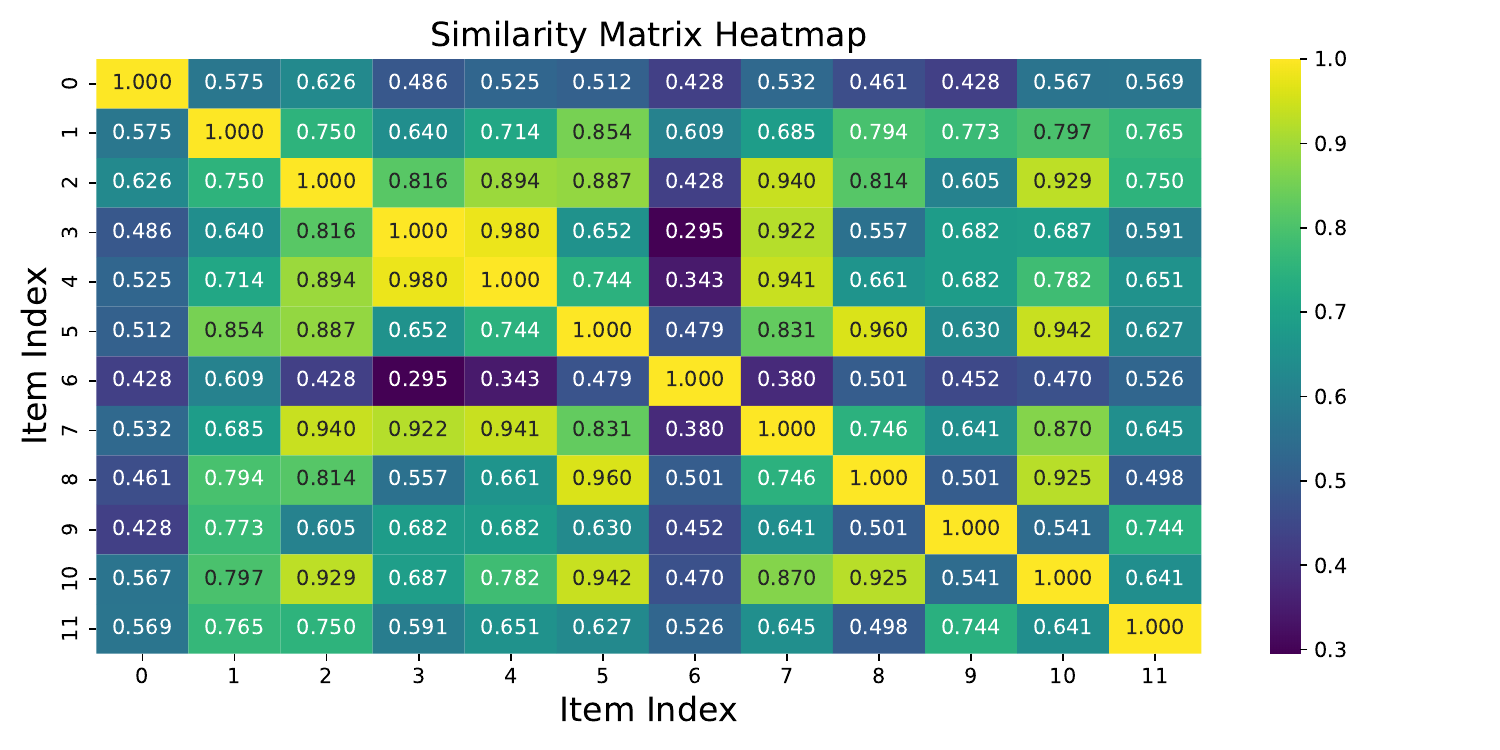}
    \caption{The heatmap showing the cosine similarity between different neural network architectures. The color and value in each cell represent the similarity score between two architectures, with values closer to 1.0 (yellow) indicating greater similarity.}
    \label{fig:heatmap}
\end{figure}

We compare the average embedding of top 10\% performing architectures of each task. Each architecture in encoded with one-hot embedding. As is shown in Figure~\ref{fig:heatmap}, similar tasks tend to have similar architecture embeddings.

\subsubsection{More results on scaling up}
\label{appendix:scaling up}
To further investigate the impact of architecture depth and the scalability of our searched architecture, we expanded discovered optimal architecture (top 1 in Figure~\ref{fig:trend}) into a 74.7M-parameter foundation model by stacking 2 Hyena, 4 Transformer, and 3 CNN layers with a hidden dimension of 1024. We extend the training steps to 100,000 and keep other hyperparameters the same. As shown in Table~\ref{tab:scaling up}, this scaling yields a improvement on 10/12 tasks, which validates the architecture's effectiveness for large-scale pre-training. \textbf{This demonstrates that our discovered architecture effectively supports scaling.}

\begin{table*}[h!]
\centering
\caption{Comparison of performance across 12 DNA tasks (Tasks 0-11) between BioArc identified foundation models and state-of-art baseline.}
\label{tab:scaling up}
\begin{small}
\begin{tabular}{lcccccccccccc}
\toprule
\textbf{Model} & \textbf{0} & \textbf{1} & \textbf{2} & \textbf{3} & \textbf{4} & \textbf{5} & \textbf{6} & \textbf{7} & \textbf{8} & \textbf{9} & \textbf{10} & \textbf{11} \\ \midrule
VQDNA (HRQ) & 72.48 & 76.43 & 66.85 & 58.92 & 78.10 & 71.02 & 70.58 & 78.50 & 90.75 & 94.48 & 74.52 & \textbf{89.53} \\
\midrule
BioArc-FM-4.8M & 82.50 & 85.10 & 76.40 & 72.40 & 82.20 & 80.47 & 83.34 & 76.35 & 90.24 & 95.25 & 79.93 & 68.26 \\
BioArc-FM-88M & \textbf{84.30} & 83.60 & 85.70 & \textbf{78.80} & 88.60 & 83.09 & 82.70 & 86.62 & 92.28 & 96.01 & 79.61 & 84.83 \\ 
\bottomrule
\end{tabular}
\end{small}
\end{table*}

\subsubsection{Extended DNA evaluation against recent genomic foundation models}
\label{appendix:nt-gb-results}
To position \our-F against the most recent genomic foundation models, we additionally evaluate it on the Nucleotide Transformer (NT) benchmark and the Genomic Benchmarks (GB), and compare against Caduceus-Ph, Caduceus-PS~\cite{schiff2024caduceus}, and GENERator~\cite{wu2025generator}. Baseline numbers are taken from the GENERator paper (its Tables~S4 and~S5), and \our-F is evaluated under the same protocol. As shown in Table~\ref{tab:nt-benchmark} (NT, MCC) and Table~\ref{tab:gb-benchmark} (GB, accuracy), \textbf{despite being far smaller than GENERator}, \our-F is highly competitive: on the NT benchmark it achieves the best result on all histone-mark and enhancer tasks (with a large margin on enhancer detection, $0.874$ vs.\ $\le0.580$), while GENERator's much larger generative pretraining gives it the edge on promoter and splice-site tasks. On the Genomic Benchmarks, \our-F leads on several human enhancer/promoter datasets (e.g., Human nonTATA Promoter, Human Enhancer Ensembl) and is within $1$--$2$ points of the best model elsewhere.

\begin{table}[h!]
\centering
\caption{NT benchmark results (MCC). Baselines are from Table~S4 of GENERator~\cite{wu2025generator}; \our-F is evaluated under the same protocol. \best{Bold} marks the best result per task.}
\label{tab:nt-benchmark}
\resizebox{0.7\linewidth}{!}{%
\begin{tabular}{l | c c c c}
\hlineB{2}
\textbf{Task} & \textbf{\our} & \textbf{Caduceus-Ph} & \textbf{Caduceus-PS} & \textbf{GENERator} \\
\hline
H3                & 0.783 & 0.794 & 0.772 & \best{0.806} \\
H3K14ac           & \best{0.648} & 0.564 & 0.596 & 0.605 \\
H3K36me3          & 0.647 & 0.590 & 0.611 & \best{0.657} \\
H3K4me1           & \best{0.571} & 0.468 & 0.487 & 0.553 \\
H3K4me2           & \best{0.547} & 0.332 & 0.431 & 0.424 \\
H3K4me3           & \best{0.640} & 0.490 & 0.528 & 0.512 \\
H3K79me3          & \best{0.738} & 0.641 & 0.682 & 0.670 \\
H3K9ac            & \best{0.644} & 0.575 & 0.564 & 0.612 \\
H4                & 0.809 & 0.788 & 0.799 & \best{0.815} \\
H4ac              & \best{0.685} & 0.548 & 0.585 & 0.592 \\
Enhancers         & \best{0.874} & 0.522 & 0.511 & 0.580 \\
Enhancers types   & \best{0.782} & 0.403 & 0.410 & 0.477 \\
Promoter all      & 0.941 & 0.937 & 0.941 & \best{0.962} \\
Promoter no TATA  & 0.940 & 0.935 & 0.940 & \best{0.962} \\
Promoter TATA     & 0.924 & 0.895 & 0.903 & \best{0.948} \\
Splice sites all  & 0.971 & 0.935 & 0.953 & \best{0.978} \\
Splice acceptors  & 0.953 & 0.918 & 0.907 & \best{0.981} \\
Splice donors     & 0.963 & 0.912 & 0.930 & \best{0.978} \\
\hlineB{2}
\end{tabular}%
}
\end{table}

\begin{table}[h!]
\centering
\caption{Genomic Benchmarks results (accuracy, \%). Baselines are from Table~S5 of GENERator~\cite{wu2025generator}; \our-F is evaluated under the same protocol. \best{Bold} marks the best result per task.}
\label{tab:gb-benchmark}
\resizebox{0.7\linewidth}{!}{%
\begin{tabular}{l | c c c c}
\hlineB{2}
\textbf{Task} & \textbf{\our} & \textbf{Caduceus-Ph} & \textbf{Caduceus-PS} & \textbf{GENERator} \\
\hline
Coding vs. Interg     & 0.949 & 0.933 & 0.944 & \best{0.963} \\
Drosophila Enh        & 0.819 & \best{0.827} & 0.816 & 0.821 \\
Human Enh Cohn        & 0.750 & 0.747 & 0.749 & \best{0.763} \\
Human Enh Ensembl     & \best{0.934} & 0.924 & 0.923 & 0.917 \\
Human Ensembl Reg     & \best{0.941} & 0.938 & \best{0.941} & 0.928 \\
Human nonTATA Prom    & \best{0.975} & 0.961 & 0.961 & 0.958 \\
Human OCR Ensembl     & 0.820 & 0.825 & \best{0.826} & 0.823 \\
Human vs. Worm        & 0.972 & 0.975 & 0.976 & \best{0.980} \\
Mouse Enh Ensembl     & 0.820 & 0.807 & 0.813 & \best{0.871} \\
\hlineB{2}
\end{tabular}%
}
\end{table}

\subsubsection{More results on Protein}
\label{appendix:protein results}
We provide additional protein results in this section. The main downstream comparison, including the controlled experiment in which \our 8M and a reimplemented ESM-2 8M are pretrained under identical conditions (full UniRef50, 50K steps) to isolate architecture, is reported in Table~\ref{table:Protein-results} in the main text. Here we additionally report the corresponding pretraining loss in Table~\ref{tab:protein-pretrain}: under the matched budget, \our 8M attains lower masked-language-modeling loss than ESM-2 8M at every checkpoint and is still decreasing at 50K steps, corroborating that its downstream advantage stems from greater data efficiency rather than a larger pretraining budget.

\begin{table}[t!]
\centering
\caption{Pretraining masked-language-modeling loss ($\downarrow$) under the controlled setting (full UniRef50, 50K steps, identical hyperparameters). \our 8M achieves lower loss than ESM-2 8M at every checkpoint and is still decreasing at 50K steps.}
\label{tab:protein-pretrain}
\resizebox{0.75\linewidth}{!}{%
\begin{tabular}{l | c c c c c c c}
\hlineB{2}
\textbf{Model} & \textbf{1K} & \textbf{5K} & \textbf{10K} & \textbf{20K} & \textbf{30K} & \textbf{40K} & \textbf{50K} \\
\hline
\our 8M & \best{2.783} & \best{2.690} & \best{2.647} & \best{2.617} & \best{2.604} & \best{2.593} & \best{2.585} \\
ESM-2 8M & 2.906 & 2.749 & 2.701 & 2.656 & 2.635 & 2.621 & 2.607 \\
\hlineB{2}
\end{tabular}%
}
\end{table}

\subsubsection{More results on different training strategies}
\label{Appendix:training results}
\begin{figure}[h!]
    \centering
    \includegraphics[width=1\linewidth]{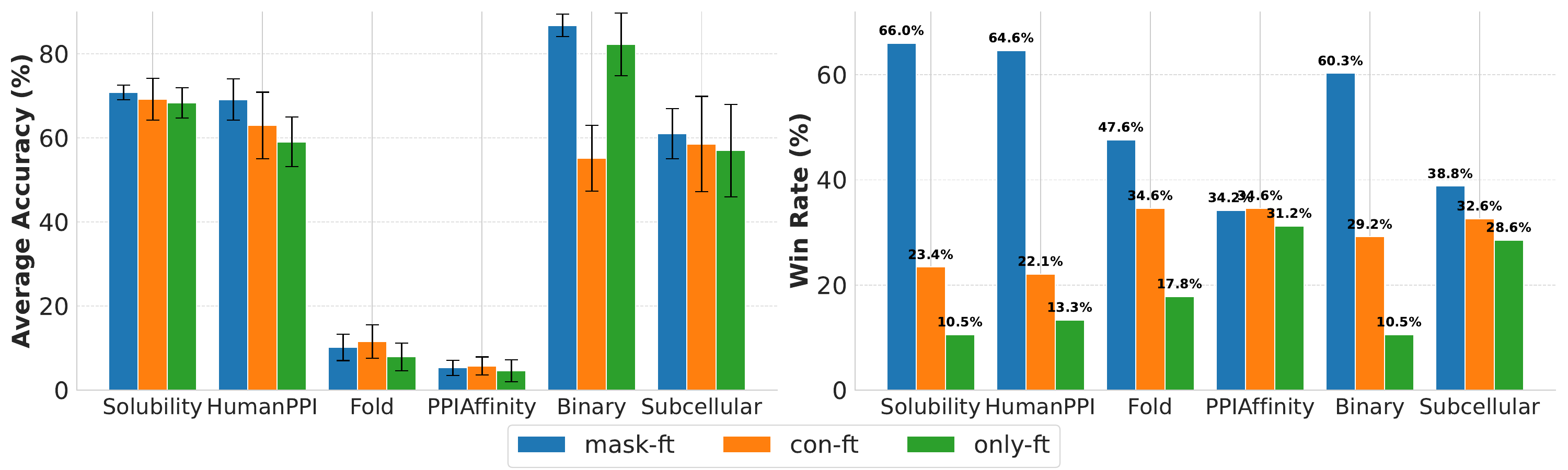}
    \caption{Performance of different training strategies on Protein. In the left panel, each cell shows the mean performance ± standard deviation across all architectures. In the right panel, each bar shows the percentage of the total 360 architectures that chose that training strategy to yield the best performance.}
    \label{Appendix:protein-training}
\end{figure}

The effect of different training strategies is shown in Figure~\ref{Appendix:protein-training}. It indicates that pretraining provides no clear advantage, and the choice of pretraining strategy appears to have minimal effect on the outcome.

\subsubsection{More results on different tokenizers }
\label{Appendix:tokenizers results}
\begin{figure}[h!]
    \centering 
    \begin{subfigure}[b]{0.48\linewidth}
        \centering
        \includegraphics[width=\linewidth]{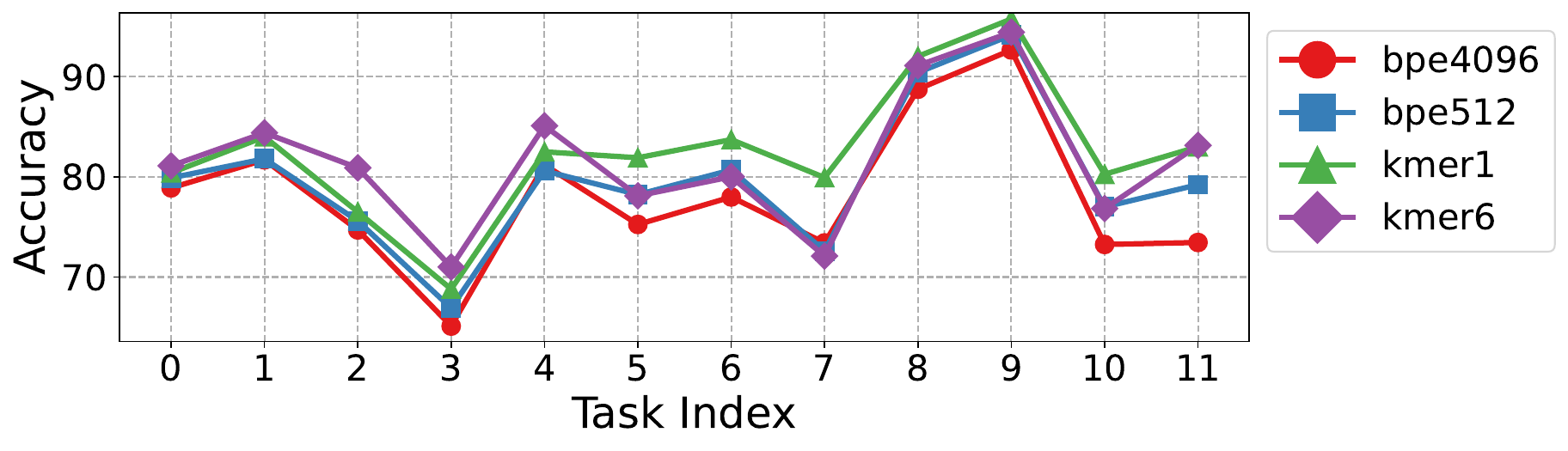}
        \caption{LSTM}
    \end{subfigure}
    \hfill
    \begin{subfigure}[b]{0.48\linewidth}
        \centering
        \includegraphics[width=\linewidth]{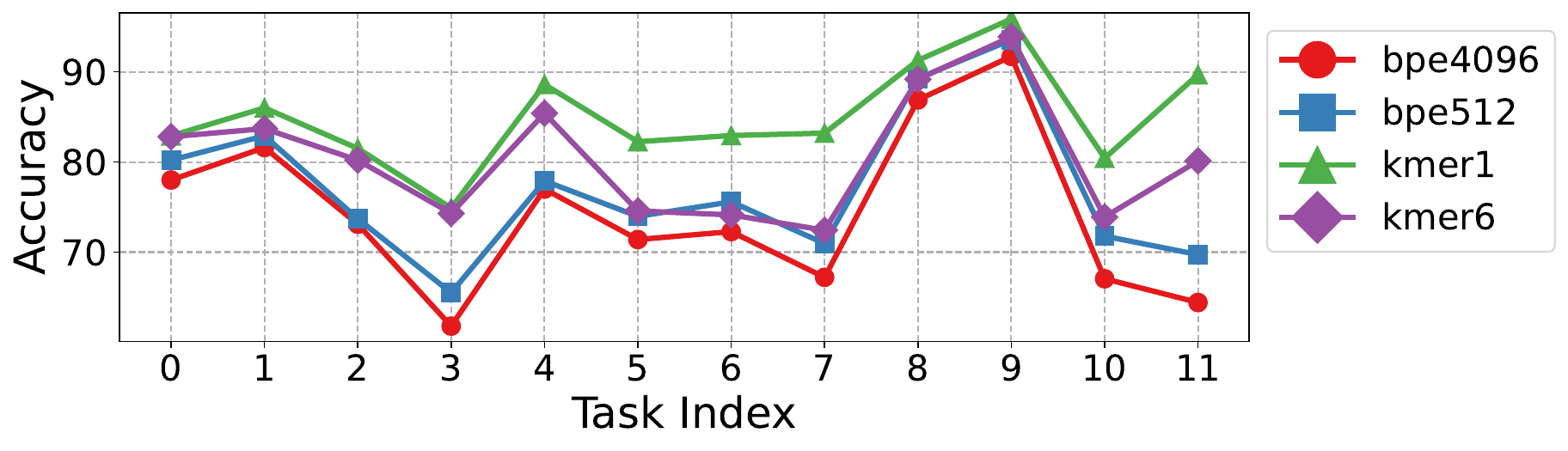}
        \caption{Mamba}
    \end{subfigure}

    \vspace{0.5cm}
    \begin{subfigure}[b]{0.48\linewidth}
        \centering
        \includegraphics[width=\linewidth]{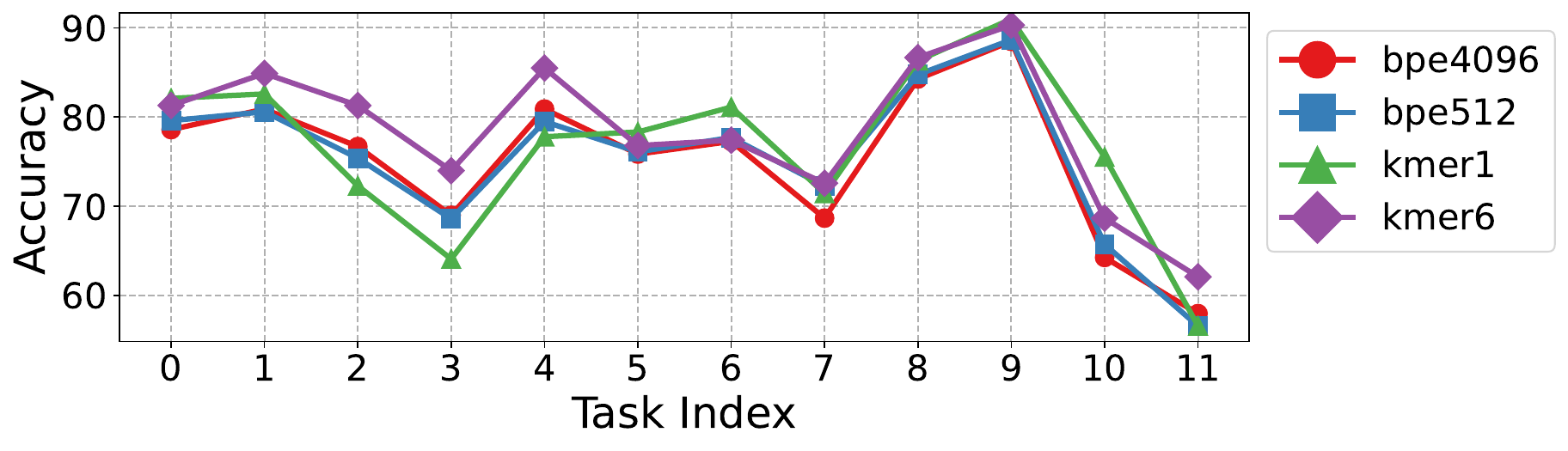}
        \caption{Hyena}
    \end{subfigure}
    \hfill
    \begin{subfigure}[b]{0.48\linewidth}
        \centering
        \includegraphics[width=\linewidth]{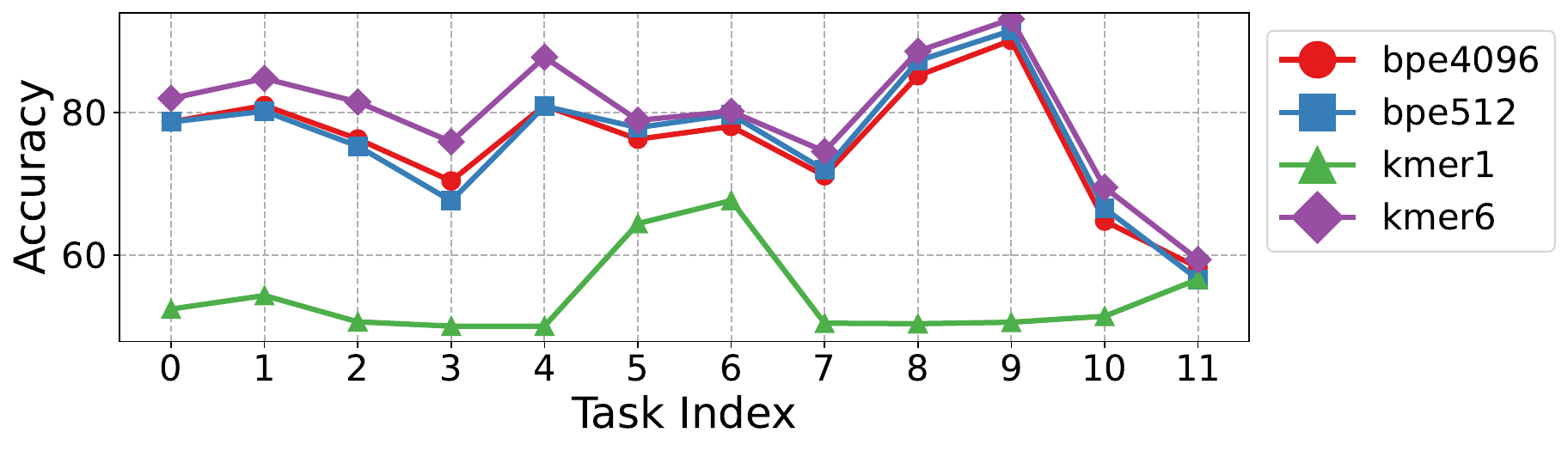} 
        \caption{Transformer}
    \end{subfigure}
    
    \caption{Performance of different tokenization methods on various architectures.}
    \label{fig:comparison_of_tokenizer_appendix}
\end{figure}

Here we show more results of the effect of different tokenizers on different architectures, shown in Figure~\ref{fig:comparison_of_tokenizer_appendix}. Our results highlight two key findings regarding tokenization. First, different architectures exhibit distinct preferences: Mamba works best with 1-mer, while Transformer favors 6-mer. Second, these preferences are task-dependent; the LSTM architecture, for example, excels with 6-mer tokenization on tasks 0–4 but requires 1-mer tokenization for optimal performance on tasks 5–10.

\subsection{More Analysis}
\subsubsection{Hybrid and single-module architecture comparison}
\label{Appendix:hybrid and single architecture comparison}
\begin{figure}[t!]
    \centering
    \includegraphics[width=1\linewidth]{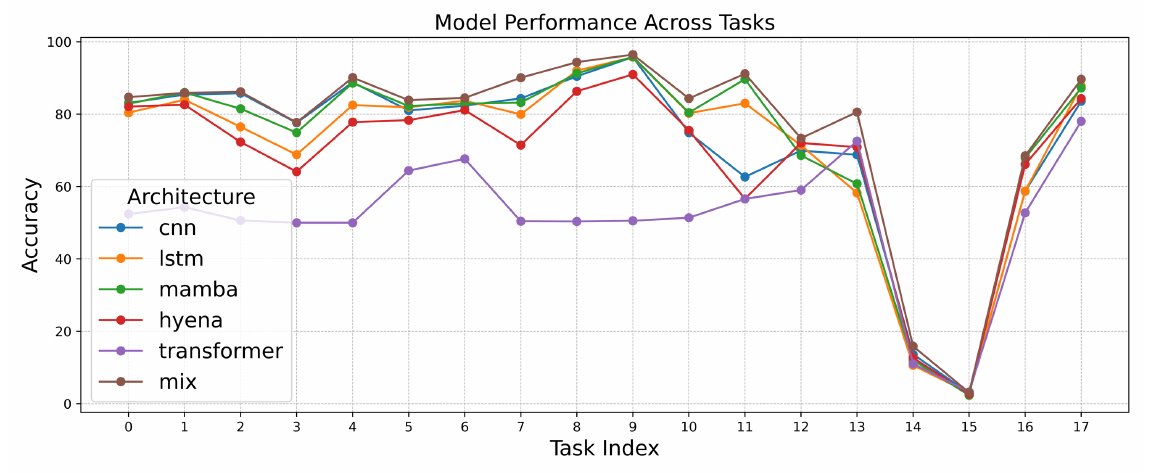}
    \caption{Performance of hybrid architecture compared with optimal single-module architectures found via restricted search.}
    \label{Appendix:figure of architecture comparison}
\end{figure}

We compare the performance of \our hybrid architectures against optimal single-module architectures. To ensure a fair comparison, the single-module baselines were not arbitrarily selected. They represent the top-performing candidates discovered by restricting the search space to exclusively use a specific module type(e.g., an all-CNN or all-Transformer search space). All models, including both hybrid and single-module variants, were trained from scratch to eliminate weight-sharing bias. As shown in Figure~\ref{Appendix:figure of architecture comparison}, \textbf{hybrid architectures consistently outperform the best single-module baselines,} which validates our intuition in designing a search space that allows for the combination of different architectures.

\subsubsection{Layer-wise Contribution analysis}
\label{appendix:layer-contribution}
To verify that \our superiority stems from the synergistic composition of hybrid modules rather than a single dominant layer, we conducted a layer-wise contribution analysis on the top-performing architecture (rank-1 in Figure~\ref{fig:trend}). We trained a separate classification head on the output of each layer across all DNA tasks to evaluate intermediate feature quality. As shown in Table~\ref{tab:layer_contribution}, we observe a consistent monotonic improvement in average accuracy (from $70.32\%$ to $83.18\%$) alongside a progressive reduction in standard error. This steady gain confirms that the specific sequence of diverse modules collaboratively refines biological representations, validating that the model's effectiveness relies on the holistic hybrid design.

\begin{table}[h]
\centering
\caption{Average accuracy and standard error across different layers.}
\resizebox{\textwidth}{!}{%
\begin{tabular}{lcccccc}
\toprule
\textbf{Metric} & \textbf{Layer 0} & \textbf{Layer 1} & \textbf{Layer 2} & \textbf{Layer 3} & \textbf{Layer 4} & \textbf{Layer 5} \\
\midrule
Accuracy ($\%$) & $70.32 \pm 0.89$ & $74.07 \pm 0.86$ & $75.09 \pm 0.82$ & $78.36 \pm 0.62$ & $82.09 \pm 0.49$ & $83.18 \pm 0.46$ \\
\bottomrule
\end{tabular}%
}
\label{tab:layer_contribution}
\end{table}

\subsubsection{Architecture Depth Analysis}
\label{appendix:depth_analysis}
\begin{figure}[t]
    \centering
    \begin{subfigure}[b]{\linewidth}
        \centering
        \includegraphics[width=\linewidth]{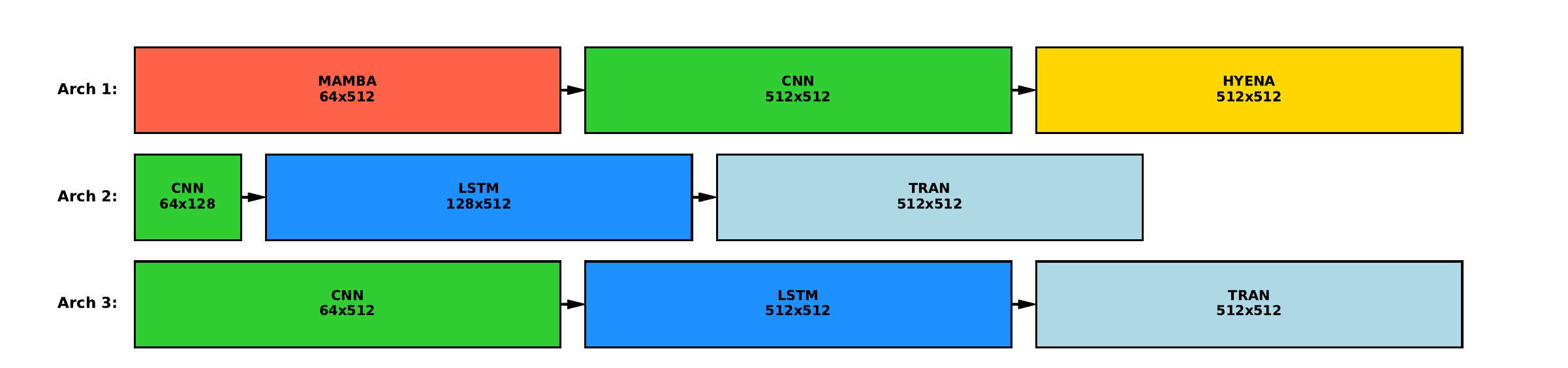}
        \caption{Top-3 architectures with 3 layers.}
        \label{fig:top3_3layers}
    \end{subfigure}
    \vspace{1mm}

    \begin{subfigure}[b]{\linewidth}
        \centering
        \includegraphics[width=\linewidth]{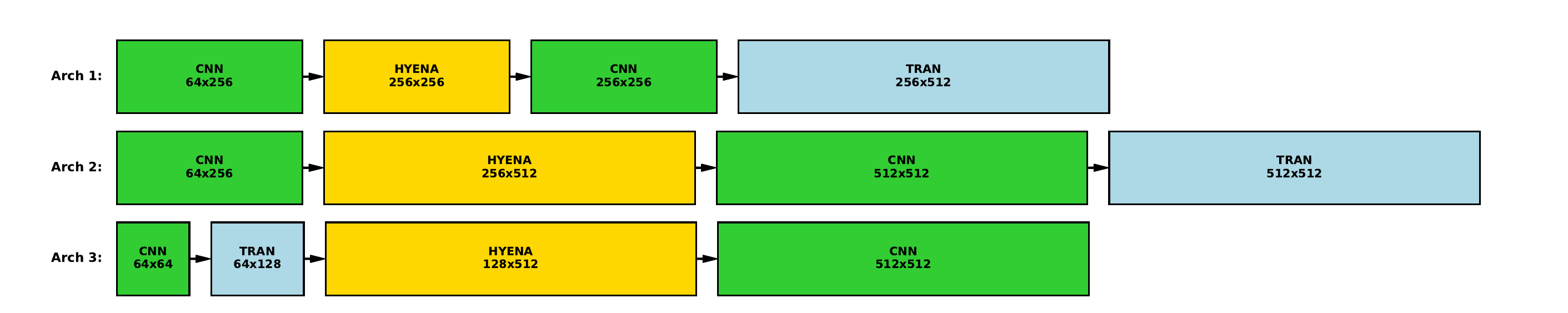}
        \caption{Top-3 architectures with 4 layers.}
        \label{fig:top3_4layers}
    \end{subfigure}
    \vspace{1mm}

    \begin{subfigure}[b]{\linewidth}
        \centering
        \includegraphics[width=\linewidth]{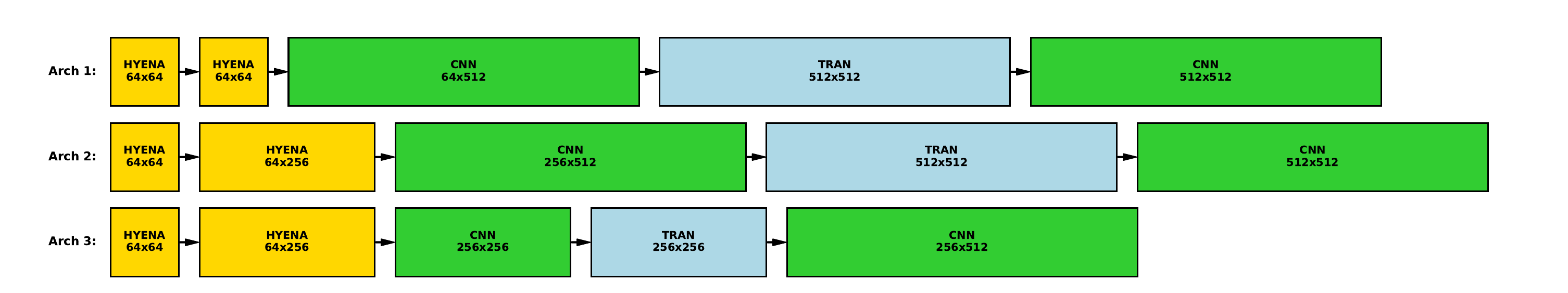}
        \caption{Top-3 architectures with 5 layers.}
        \label{fig:top3_5layers}
    \end{subfigure}
    \vspace{1mm}

    \begin{subfigure}[b]{\linewidth}
        \centering
        \includegraphics[width=\linewidth]{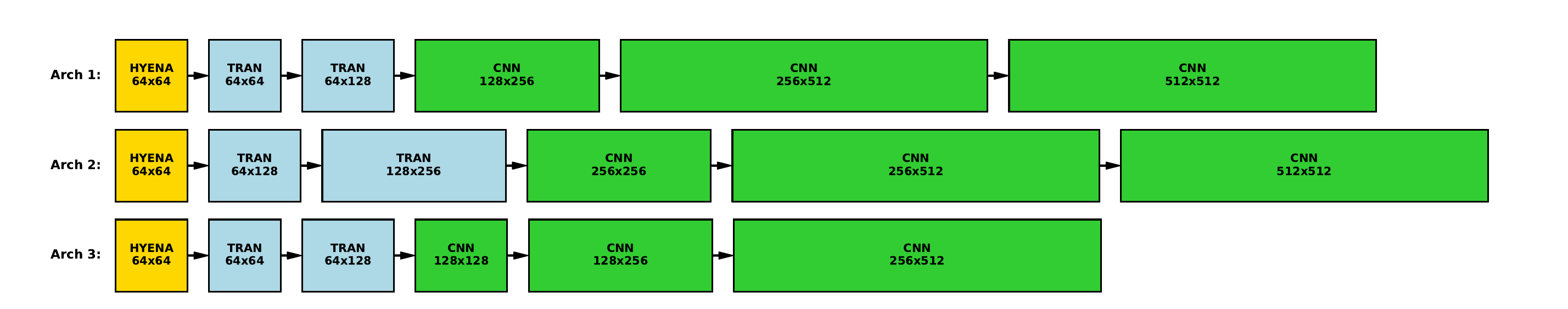}
        \caption{Top-3 architectures with 6 layers.}
        \label{fig:top3_6layers}
    \end{subfigure}
    
    \caption{Visualization of the top-performing architectures across different depths.}
    \label{fig:all_layers_top_arc}
\end{figure}

We investigate the relationship between network depth and transfer performance by analyzing the statistics of architectures ranging from 3 to 6 layers. For each of the sampled architectures, we compute the \textbf{average accuracy across all 12 DNA downstream tasks} to represent its overall generalization capability. From Table~\ref{tab:layer_num}, we observe that while deeper architectures demonstrate higher median and maximum performance, they also exhibit lower mean accuracy and higher standard deviations. This suggests that deeper models are more sensitive/susceptible to the fixed hyperparameter settings. \textbf{This implies that depth is not the sole determinant of performance; rather, the structural composition of the architecture is key.}
We also observe that in top architectures of 4 and 5 layers demonstrate structure as 6 layers in Figure~\ref{fig:all_layers_top_arc}. \textbf{This indicates that once an effective architectural pattern is identified, scaling up the depth can further enhance performance.}

\begin{table}[t!]
\centering
\caption{Performance statistics of architectures grouped by depth. Metrics (Mean, Std, Min, Max, Median) are calculated based on the \textbf{average accuracy across all tasks}. All results are obtained under \textbf{fixed hyperparameters}.}
\label{tab:depth_analysis}
\setlength{\tabcolsep}{9pt}
\footnotesize
\resizebox{0.8\linewidth}{!}{%
\begin{tabular}{@{}l c c c c c c@{}}
\toprule
\textbf{Depth} & \textbf{Count} & \textbf{Mean (\%)} & \textbf{Std} & \textbf{Min (\%)} & \textbf{Max (\%)} & \textbf{Median (\%)} \\ \midrule
3 Layers & 60  & 81.00 & 4.23  & \textbf{53.77 }& 84.85 & 81.49 \\
4 Layers & 100 & \textbf{81.74} & 5.57  & 53.38 & 85.14 & 83.07 \\
5 Layers & 100 & 77.74 & 10.24 & 47.44 & 85.48 & 82.56 \\
6 Layers & 100 & 78.17 & 10.72 & 47.70 & \textbf{86.53} & \textbf{83.30} \\ \bottomrule
\end{tabular}%
}
\label{tab:layer_num}
\end{table}

\subsubsection{Performance-Parameter Analysis}
\label{appendix:performance-parameter}
\begin{figure}
    \centering
    \includegraphics[width=0.8\linewidth]{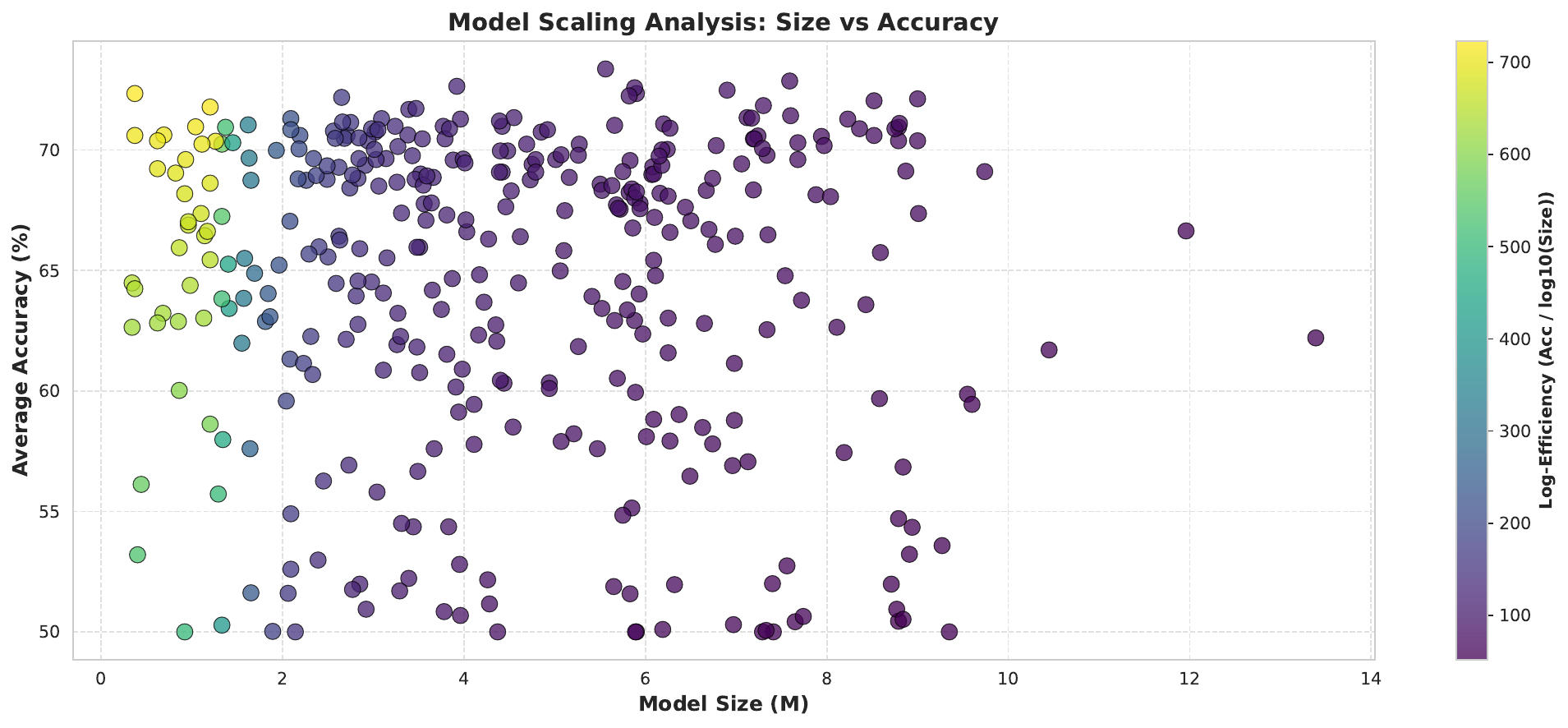}
    \caption{Model Scaling Analysis (Performance vs. Parameters). Each point represents a architecture finetuned with frozen pretrained supernet weight. The color indicates Log-Efficiency, calculated as $Accuracy / \log_{10}(Size)$, with lighter colors (yellow) representing higher efficiency. The plot demonstrates that larger model sizes do not guarantee higher accuracy, highlighting that the specific architectural topology is the dominant factor in performance.}
    \label{fig:performance-parameters}
\end{figure}

To investigate whether the performance of \our-discovered models stems from increased capacity or superior design, we analyzed the relationship between parameter count and average accuracy across the search space as shown in Figure~\ref{fig:performance-parameters}. We evaluated all paths directly from the pretrained supernet by freezing the backbone weights and training only a linear classification head. We observe no positive correlation between model size and performance. High accuracy scores are achieved by efficient architectures (yellow points) without relying on large parameter budgets, while increasing model size does not guarantee performance gains. The substantial performance variance observed among models of identical size indicates that the specific topological arrangement of modules—the inductive bias—is the decisive factor for biological sequence modeling. This confirms that \our's effectiveness lies in discovering these optimal architectural patterns rather than simple parameter scaling.

\subsubsection{Architecture Ranking Correlation Analysis}
\label{appendix:architecture-ranking-correlation}
To validate our evaluation protocol and justify the necessity of the decoupled pretrain-then-finetune strategy, we conducted a two-fold correlation analysis. We compare the ranking consistency across three distinct experimental settings for the 360 representative candidate architectures:

\begin{itemize}
    \item \textbf{Training from Scratch Individually (Ground Truth):} Architectures are initialized randomly and trained independently on downstream tasks. This serves as the performance baseline.
    \item \textbf{Supernet Pretraining + Supernet Finetuning:} Architectures are evaluated directly using the shared weights from the supernet (after pretraining and supernet-level finetuning), without decoupling the parameters for each path.
    \item \textbf{Supernet Pretraining+ Individual Finetune (Our Method):} Architectures inherit weights from the pretrained \our supernet but are finetuned independently.
\end{itemize}

\begin{figure}[t!]
    \centering
\includegraphics[width=0.8\linewidth]{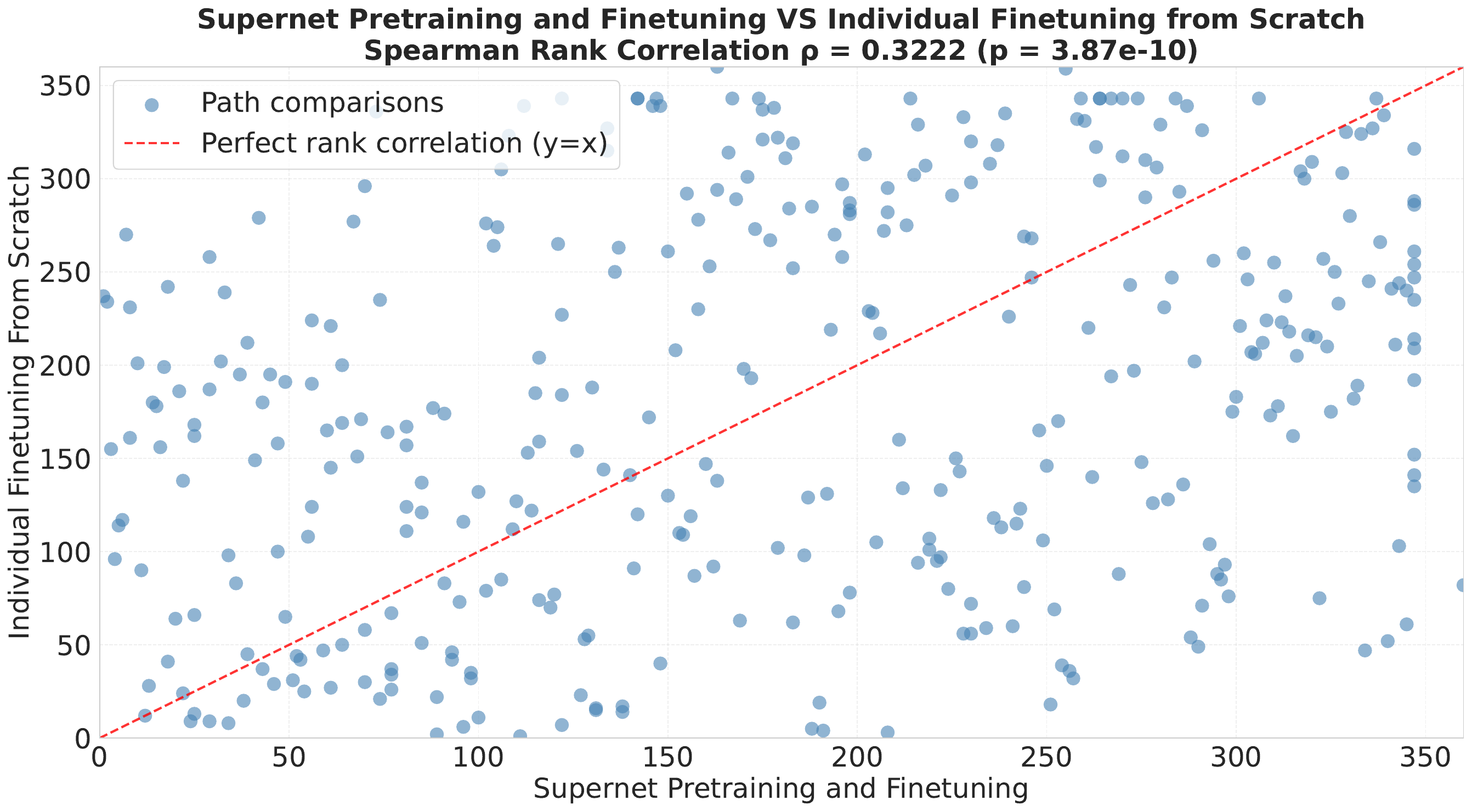}
    \caption{Rank consistency analysis between Supernet Pretraining with Finetuning and Individually Trained from Scratch. Each point represents one of the 360 candidate architectures. The x-axis represents the performance rank derived from the Supernet, and the y-axis represents the ground-truth rank. The weak correlation ($\rho=0.3222$) reveals the limitations of the Supernet in this setting, demonstrating that ranking architectures solely based on direct finetuning does not consistently reflect their true performance from scratch.}
    \label{fig:rank_correlation_1}
\end{figure}

\paragraph{Necessity of Individual Finetuning.} 
First, we investigate whether we could directly finetune the supernet for correct ranking of architectures. We observe a low correlation between \textbf{\textit{Supernet Pretraining + Supernet Finetuning}} and the \textbf{\textit{Ground Truth}}, as shown in Figure~\ref{fig:rank_correlation_1}. This discrepancy indicates the phenomenon of weight interference and co-adaptation prevents the shared parameters from accurately reflecting the distinct potential of specific paths.
Consequently, this validates our methodological choice in Section~\ref{Section:Evaluation Protocol} that we cannot rely on the supernet for direct ranking. Instead, evaluating architectures requires \textbf{individual finetuning} to decouple weights and reveal the true performance potential of each candidate.

\begin{figure}[t!]
    \centering
    \includegraphics[width=0.80\linewidth]{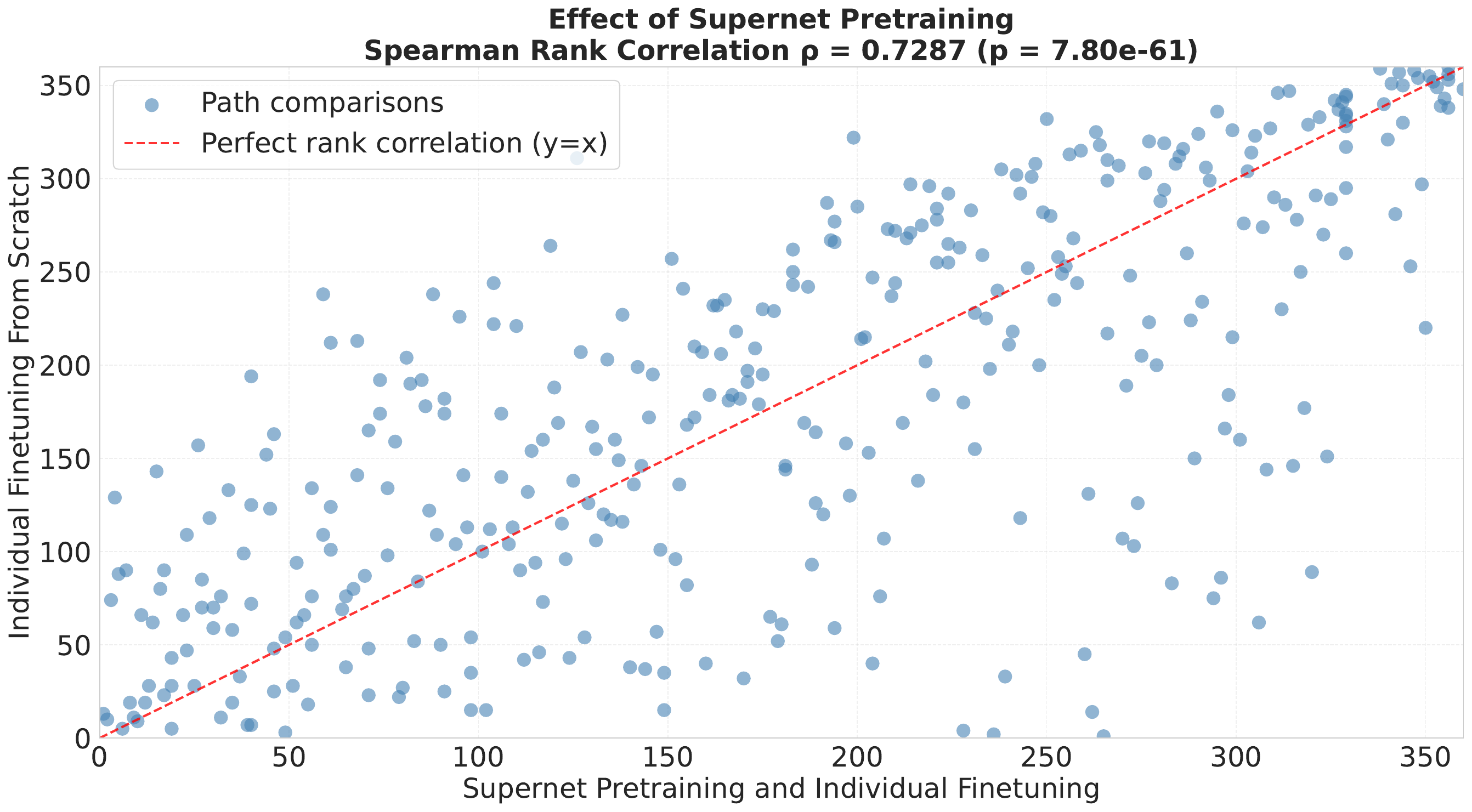}
    \caption{Rank consistency analysis between \textit{Supernet Pretraining and Individual Finetuning} and \textit{Individual Finetuning From Scratch}. Each point represents one of the 360 candidate architectures. The x-axis represents the performance rank derived from the Supernet (lower is better), and the y-axis represents the ground-truth rank from scratch training. The strong linear alignment and high Spearman correlation ($\rho=0.7287$) indicates that our pretraining strategy effectively mitigates the ranking disorder problem commonly found in weight-sharing NAS, ensuring that high-performing architectures in the supernet remain superior when trained independently.}
    \label{fig:rank_correlation_2}
\end{figure}

\paragraph{Validation of Pretraining Effectiveness.} 
Second, we assess the correlation between \textbf{\textit{Supernet Pretraining+ Individual Finetune}} and \textbf{\textit{Training from Scratch}}. As illustrated in Figure~\ref{fig:rank_correlation_2}, we observe a strong positive correlation ($\rho = 0.7287$) between these two protocols. This high consistency demonstrates two critical points: 
(1) The performance ranking of architectures is intrinsic to their topology and remains robust regardless of the initialization strategy (random vs. pretrained).
(2) The \our supernet successfully learns high-quality, transferable features during pretraining, as it preserves the ground-truth ranking order when used as an initialization backbone.

\subsubsection{Shared-Block Gradient Stability}
\label{appendix:gradient-stability}
A potential concern with weight-sharing supernets is that a block shared across many sampled paths receives gradients from stochastically varying predecessors and successors, which could destabilize its optimization. We address this directly by tracking, for each shared block, the \textbf{standard deviation of its gradient norm across all paths that contain it}, and showing that this variance decreases substantially and stabilizes over training. We report results on (1) a controlled supernet and (2) our original supernet, and note that this stability is consistent with established theoretical and empirical findings in the weight-sharing NAS literature~\cite{li2021bossnasexploringhybridcnntransformers}.

\paragraph{Controlled supernet.}
We construct a 3-layer supernet with 5 block types per layer ($5^3=125$ paths, 15 shared blocks). We focus on middle-layer blocks (Layer~1, $64\to128$), as each block's 25 containing paths exhaust all predecessor $\times$ successor type combinations ($5\times5$), providing the strictest test of robustness to upstream and downstream distributional shifts. Gradient norms are computed on the same fixed mini-batches. As shown in Table~\ref{tab:grad-stability-controlled}, the gradient-norm standard deviation of every shared block decreases substantially from initialization and stabilizes over training, confirming that each block converges stably despite receiving inputs from stochastically sampled predecessors.

\begin{table}[h!]
\centering
\caption{Gradient-norm standard deviation of each middle-layer shared block (Layer~1, $64\to128$) in the controlled supernet, computed across all 25 containing paths over training epochs (E$x$).}
\label{tab:grad-stability-controlled}
\resizebox{0.85\linewidth}{!}{%
\begin{tabular}{l | c c c c c c c}
\hlineB{2}
\textbf{Block} & \textbf{Init} & \textbf{E1} & \textbf{E2} & \textbf{E5} & \textbf{E10} & \textbf{E15} & \textbf{E20} \\
\hline
CNN          & 0.769  & 0.150 & 0.087 & 0.048 & 0.037 & 0.030 & 0.036 \\
Hyena        & 8.719  & 0.196 & 0.175 & 0.131 & 0.123 & 0.123 & 0.141 \\
LSTM         & 14.297 & 0.331 & 0.187 & 0.175 & 0.203 & 0.201 & 0.210 \\
Mamba        & 3.238  & 0.196 & 0.159 & 0.129 & 0.118 & 0.124 & 0.132 \\
Transformer  & 3.627  & 0.605 & 0.170 & 0.115 & 0.107 & 0.115 & 0.124 \\
\hlineB{2}
\end{tabular}%
}
\end{table}

\paragraph{Original supernet.}
To confirm these findings are not an artifact of the simplified setup, we repeat the analysis on our original supernet. We select 5 representative shared blocks, one per block type, spanning different dimensions and layer positions, and track their gradient-norm standard deviation across all containing paths (gradient norms computed on the same fixed mini-batches). As shown in Table~\ref{tab:grad-stability-original}, all five blocks exhibit an overall decaying trend, stabilizing after the initial training phase. Crucially, the \emph{mean} gradient norm remains on a healthy scale throughout (e.g., CNN: $0.115\to0.099$, Transformer: $0.247\to0.287$ from E4 to E10), confirming that the standard-deviation decay reflects genuine convergence of block representations across paths rather than gradient vanishing.

\begin{table}[h!]
\centering
\caption{Gradient-norm standard deviation of 5 representative shared blocks in the original supernet, tracked across all containing paths over training epochs (E$x$).}
\label{tab:grad-stability-original}
\resizebox{0.85\linewidth}{!}{%
\begin{tabular}{l | c c c c c c c}
\hlineB{2}
\textbf{Block} & \textbf{Init} & \textbf{E1} & \textbf{E2} & \textbf{E4} & \textbf{E6} & \textbf{E8} & \textbf{E10} \\
\hline
CNN ($64\to128$)          & 0.507 & 0.054 & 0.126 & 0.044 & 0.038 & 0.047 & 0.043 \\
Transformer ($512\to512$) & 1.905 & 0.341 & 0.795 & 0.172 & 0.145 & 0.162 & 0.170 \\
Hyena ($512\to512$)       & 4.898 & 1.568 & 2.537 & 0.129 & 0.100 & 0.097 & 0.098 \\
Mamba ($64\to128$)        & 0.783 & 0.129 & 0.273 & 0.069 & 0.056 & 0.060 & 0.071 \\
LSTM ($256\to256$)        & 3.298 & 0.218 & 0.452 & 0.113 & 0.103 & 0.113 & 0.126 \\
\hlineB{2}
\end{tabular}%
}
\end{table}

\subsubsection{Architecture Prediction}
\label{appendix:architecture-prediction}
Training every candidate in a vast \our search space is computationally expensive. Leveraging our observation that \textbf{optimal architectures for functionally similar tasks exhibit significant topological similarity}, we propose a hierarchy of three approaches, ranging from numerical regression to high-level semantic reasoning agents, to efficiently exploit \our explored (architecture, task, performance) tuples for identifying optimal architectures for unseen biological tasks.

\paragraph{Experiment Settings}
Based on our observations from RQ2, we conducted experiments in both supervised and transfer settings with detailed data split available in Appendix~\ref{Appendix: prediction split}.
Unlike standard classification, the "optimal" architecture is a ranking relative to the search space. We define our metrics as follows:
\begin{itemize}[leftmargin=*]
    \item \textbf{Ground Truth Construction ($\mathcal{G}_t$):} For every task $t$ in our benchmark, we utilize the exhaustive evaluation results from Section 3.4. We define the Ground Truth Set $\mathcal{G}_t$ as the top architectures ranked by their performance for each task. These represent the empirically verified optimal designs.

    \item \textbf{Metric Definitions:} Let $\mathcal{P}_t^k$ be the set of top-$k$ architectures predicted by our model for task $t$. We evaluate the alignment between prediction and reality using:
    \begin{enumerate}
        \item \textbf{Precision@k ($|\mathcal{P}_t^k \cap \mathcal{G}_t| / k$):} This measures the \textit{efficiency} of the prediction. It quantifies the proportion of the suggested architectures that are truly top-tier, indicating the reliability of the predictor's advice to a user with a limited budget for training trials.
        \item \textbf{Recall@k ($|\mathcal{P}_t^k \cap \mathcal{G}_t| / |\mathcal{G}_t|$):} This measures the \textit{coverage} of the design space. It assesses the predictor's ability to uncover the diverse set of high-performing candidates, preventing mode collapse into a single architecture type.
        \item \textbf{Hit Rate@k ($\mathbb{I}(|\mathcal{P}_t^k \cap \mathcal{G}_t| \geq 1)$):} This measures the \textit{probability of success}. It indicates whether the user will find \textit{at least one} optimal architecture within the top-$k$ predictions, serving as a critical "success/failure" metric for practical deployment.
    \end{enumerate}
\end{itemize}

\paragraph{Neural Network Prediction} 
\label{Section:nn predictor}
We design a neural predictor denoted as $\mathcal{P}$ that maps an architecture-task pair $(a, t)$ to a predicted performance score. The model takes an architecture embedding $\mathbf{h}_a$ and a task embedding $\mathbf{h}_t$ as input and is trained by minimizing the Mean Squared Error against the ground-truth performance $y_{a,t}$. This objective is formally expressed through the loss function $\mathcal{L}(\mathcal{P})$:
\begin{equation}
\mathcal{L}(\mathcal{P}) = \mathbb{E}_{(a,t) \sim \mathcal{A}} \left[ \left( \mathcal{P}(\mathbf{h}_a, \mathbf{h}_t) - y_{a,t} \right)^2 \right]
\label{eq:predictor_loss}
\end{equation}

Following previous work~\cite{ma2019deepneuralarchitecturesearch,white2021powerfulperformancepredictorsneural}, we represent each architecture $a$ as a graph with its node features $\mathbf{h}_a$ formed by concatenating its one-hot encoded module type $\text{O}(m_i)$ and the normalized dimensions of its input and output, $Z(h_{i-1})$ and $Z(h_i)$. For task embeddings, a pretrained language model (PLM) is used to encode the task's textual description, $d_{\text{task}}$, into a vector $\mathbf{h}_t$. These encoding processes are defined as:
\begin{equation}
\mathbf{X}_{a} = \left( \text{O}(m_i) \Vert Z(h_{i-1}) \Vert Z(h_i) \right)_{i=1}^{d},
\qquad
\mathbf{h}_{a} = \mathbf{GNN}(\mathbf{I},\mathbf{X}_{a}), \qquad \mathbf{h}_{t} = \text{PLM}(d_{\text{task}})
\end{equation}
where $d$ is the depth of the architecture. An example of architecture embedding is in Appendix~\ref{appendix:architecture similarity}. 

\paragraph{LLM + RAG} 
Moving beyond numerical regression, we leverage LLMs augmented with retrieval capabilities. To provide relevant context, we first encode the input task description into \textbf{an embedding vector and retrieve the top-$n$ most similar historical tasks} based on vector similarity. These retrieved tasks, along with their corresponding top-$k$ architectures, are formatted as a \textbf{knowledge base} and \textbf{performance records} within the prompt. The LLM is then instructed to act as an analyst: it first evaluates the new task's characteristics, identifies the most relevant matches from the retrieved knowledge base, and predicts the top-$m$ architecture. This approach enables the model to explicitly reason about task relationships and empirical performance. $n,k,m$ are hyperparameters. Detailed LLM prompts are provided in Appendix~\ref{Appendix:LLM-prompts}.

\paragraph{\our Agent} 
To address the limitations of simple embedding-based retrieval in capturing biological nuances, we introduce \our Agent. As detailed in Table~\ref{tab:agent_roles}, this system transforms unstructured queries into empirically grounded predictions. Specifically, the system identifies the top-$k$ most semantically similar historical tasks and retrieves their corresponding top-$n$ high-performing architectures. Then Predictor conducts reasoning and output top-$m$ architectures. $n,k,m$ are hyperparameters. Full prompts and operational details are provided in Appendix~\ref{Appendix:Agent-prompts}.
\begin{table}[t]
    \centering
    \small
    \renewcommand{\arraystretch}{1.1}
    \caption{Functional breakdown of the \our Agent pipeline. The system progresses from parsing raw text to synthesizing an optimal design.}
    \label{tab:agent_roles}

    \begin{tabularx}{\textwidth}{l >{\hsize=0.85\hsize\raggedright\arraybackslash}X >{\hsize=1.15\hsize\raggedright\arraybackslash}X}
        \toprule
        \rowcolor{gray!10}
        \textbf{Role} & \textbf{Input} & \textbf{Output} \\
        \midrule
        
        \textbf{Analyst} 
        & Raw user task description 
        & Structured metadata (e.g., modality, objective) \\

        \addlinespace[0.5ex]
        
        \textbf{Task Retriever} 
        & Structured metadata 
        & Semantically aligned tasks in Knowledge Base \\
        
        \addlinespace[0.5ex]
        
        \textbf{Arch. Retriever} 
        & Aligned tasks 
        & Proven architectures \& empirical performance \\
        
        \addlinespace[0.5ex]
        
        \textbf{Predictor} 
        & Retrieved architectures \& metrics 
        & Predict optimal architecture design \\
        \bottomrule
    \end{tabularx}
\end{table}

\paragraph{Experiment Results} We use GPT-4o as default backbone and compared the performance of multiple LLM models. From Table~\ref{table:architecture prediction}, \textbf{we find that \our Agent consistently outperforms other methods in both settings.} 
We attribute this improvement to the modular multi-agent design, which decouples the decision-making process to minimize hallucination and error propagation. Furthermore, unlike static embeddings, our natural language-driven retrieval ensures a semantic matching of tasks, allowing the system to identify historical precedents that share genuine structural and functional requirements.

\begin{table}[h!]
\centering
\caption{Architecture prediction methods comparison under supervised and transfer setting.} 
\label{table:architecture prediction}
\resizebox{\linewidth}{!}{%
\begin{tabular}{@{}llccccc ccc ccc@{}}
\toprule
& & \multicolumn{5}{c}{\textbf{NN Predictor}} & \multicolumn{3}{c}{\textbf{LLM+RAG}} & \multicolumn{3}{c}{\textbf{\our Agent}} \\ 
\cmidrule(lr){3-7} \cmidrule(lr){8-10} \cmidrule(lr){11-13}
\textbf{Setting} & \textbf{Metric} & \textbf{@5} & \textbf{@10} & \textbf{@15} & \textbf{@20} & \textbf{@30} & \textbf{@1} & \textbf{@3} & \textbf{@5} & \textbf{@1} & \textbf{@3} & \textbf{@5} \\
\midrule
\multirow{3}{*}{\textbf{Supervised}} 
& Hit Rate    & 0.167 & 0.333 & 0.333 & 0.333 & 0.667 & 0.000 & 0.167 & 0.167 & \textbf{0.500} & \textbf{0.500} & \textbf{0.500} \\
& Precision   & 0.033 & 0.033 & 0.022 & 0.017 & 0.028 & 0.000 & 0.056 & 0.111 & \textbf{0.500} & \textbf{0.444} & \textbf{0.300} \\
& Recall      & 0.033 & 0.067 & 0.067 & 0.100 & 0.167 & 0.000 & 0.056 & 0.067 & \textbf{0.100} & \textbf{0.267} & \textbf{0.300} \\
\midrule
\multirow{3}{*}{\textbf{Transfer}}
& Hit Rate    & 0.000 & 0.250 & 0.250 & 0.250 & 0.500 & 0.000 & 0.000 & 0.000 & \textbf{0.250} & \textbf{0.250} & \textbf{0.500} \\
& Precision   & 0.000 & 0.025 & 0.017 & 0.013 & 0.017 & 0.000 & 0.000 & 0.000 & \textbf{0.250} & \textbf{0.250} & \textbf{0.300} \\
& Recall      & 0.000 & 0.050 & 0.050 & 0.050 & 0.100 & 0.000 & 0.000 & 0.000 & \textbf{0.050} & \textbf{0.150} & \textbf{0.300} \\
\bottomrule
\end{tabular}%
}
\end{table}

We benchmarked several LLMs across different scales, ranging from small open-source models (Qwen3-4B, Llama-3.1-8B) to proprietary frontier models (GPT-4o, GPT-5), reported in Table~\ref{table:prediction_results_appendix}. Smaller models, such as Qwen3-4B (both Instruct and Thinking variants), failed to produce valid predictions (yielding 0.00 scores), while Llama-3.1-8B showed only marginal performance. This suggests that the complexity of the \our architecture search space exceeds \textbf{the reasoning capacity of current small-scale models.} The Supervised setting, as expected, generally yields higher precision and recall than the Transfer setting. \textbf{This aligns with our observation that similar downstream tasks tend to benefit from similar architectures.}

\begin{table}[h!]
\centering
\caption{Architecture prediction results with our Agent system. }
\label{table:prediction_results_appendix}
\resizebox{0.75\textwidth}{!}{%
\begin{tabular}{@{}ll *{9}{c}@{}}
\toprule
\multirow{2}{*}{\textbf{Setting}} & \multirow{2}{*}{\textbf{Model}} & \multicolumn{3}{c}{\textbf{Precision}} & \multicolumn{3}{c}{\textbf{Hit Rate}} & \multicolumn{3}{c}{\textbf{Recall}} \\
\cmidrule(lr){3-5} \cmidrule(lr){6-8} \cmidrule(lr){9-11}
& & @1 & @3 & @5 & @1 & @3 & @5 & @1 & @3 & @5 \\
\midrule
\multirow{5}{*}{\textbf{Supervised}}
& GPT-5 & 0.33 & 0.28 & 0.17 & 0.33 & 0.33 & 0.33 & 0.07 & 0.17 & 0.17 \\
& GPT-4o & 0.50 & 0.44 & 0.30 & 0.50 & 0.50 & 0.50 & 0.10 & 0.27 & 0.30 \\
& Llama-3.1-8B & 0.17 & 0.06 & 0.03 & 0.17 & 0.17 & 0.17 & 0.03 & 0.03 & 0.03    \\
& Qwen3-4B-Instruct & 0.00 & 0.00 & 0.00 & 0.00 & 0.00 & 0.00 & 0.00 & 0.00 & 0.00 \\
& Qwen3-4B-Thinking & 0.00 & 0.00 & 0.00 & 0.00 & 0.00 & 0.00 & 0.00 & 0.00 & 0.00 \\
\midrule
\multirow{5}{*}{\textbf{Transfer}}
& GPT-5 & 0.25 & 0.25 & 0.25 & 0.25 & 0.25 & 0.25 & 0.05 & 0.15 & 0.25 \\
& GPT-4o & 0.25 & 0.25 & 0.30 & 0.25 & 0.25 & 0.50 & 0.05 & 0.15 & 0.30 \\
& Llama-3.1-8B & 0.25 & 0.08 & 0.05 & 0.25 & 0.25 & 0.25 & 0.05 & 0.05 & 0.05 \\
& Qwen3-4B-Instruct & 0.00 & 0.00 & 0.00 & 0.00 & 0.00 & 0.00 & 0.00 & 0.00 & 0.00 \\
& Qwen3-4B-Thinking & 0.00 & 0.00 & 0.00 & 0.00 & 0.00 & 0.00 & 0.00 & 0.00 & 0.00 \\
\bottomrule
\end{tabular}%
}
\end{table}

\subsection{Hybrid Architecture Interpretation Analysis}
\label{appendix:interpretation}

We validate our hybrid architecture (Hyena-Transformer-CNN) by analyzing whether it captures the \textbf{biological grammar} of human genomic sequences, focusing specifically on the challenging task of \textbf{Core Promoter Detection (no-TATA)}.

\subsubsection{Task and Grammar Description}
\label{Appendix:grammar}

The model analyzes a short $70\text{bp}$ DNA window ($-35\text{bp}$ to $+35\text{bp}$) centered on the Transcription Start Site (TSS). Unlike TATA promoters which contain strong, identifiable motifs, \textbf{no-TATA promoters} (which constitute $\sim76\%$ of human genes) rely on weaker, combinatorial signals. The model must implicitly learn a grammar distinct from the classical TATA paradigm:

\begin{description}[style=unboxed, leftmargin=0pt, font=\bfseries]
    \item[Core Elements (Vocabulary)] 
    In the absence of the TATA-box, the recognition vocabulary expands to a cooperative set of motifs. We focus on two non-canonical categories: 
    (1) \textbf{DPE-driven Promoters}~\cite{Burke1997DPE}, which rely on the synergy between the Initiator (Inr) and downstream elements like the MTE and \textbf{DPE} (see Table~\ref{tab:core-promoter-elements}); and 
    (2) \textbf{CpG-island Promoters}~\cite{deaton2011cpg}, characterized by high GC-content and a lack of specific motifs.
    
    \item[Positional Constraints (Syntax)] 
    For no-TATA focused promoters, the spatial syntax is rigid to compensate for weaker binding affinity:
    \begin{itemize}[leftmargin=1.5em, label=$\bullet$]
        \item \textbf{Inr-DPE Spacing:} The DPE functions strictly when located exactly $+28$ to $+32\text{bp}$ downstream of the Inr~\cite{kutach2000dpe}.
        \item \textbf{TSS Definition:} Without a TATA box ($\sim -30\text{bp}$), the Inr element itself becomes the primary determinant of the $+1$ position.
    \end{itemize}
    
    \item[Combinatorial Rules (Semantics)] 
    The model must distinguish "functional grammar" from random occurrences. A critical semantic rule is the \textbf{Inr+DPE synergy}: neither element strongly recruits transcription machinery alone, but their simultaneous presence at the correct distance creates a high-affinity binding site. Alternatively, in CpG contexts, the model detects "broad" initiation patterns rather than precise motifs.
\end{description}

\begin{table}[h!]
    \centering
    \caption{Key Elements in no-TATA Human Core Promoters. Since the TATA-box is absent, recognition relies on the Initiator and downstream regions. Positions are relative to the TSS (+1).}
    \label{tab:core-promoter-elements}
    \vspace{2mm}
    \begin{tabular}{l c c l}
        \toprule
        \textbf{Element} & \textbf{Position} & \textbf{Abbr.} & \textbf{Consensus Sequence} \\
        \midrule
        Initiator          & $-2$ to $+5$   & Inr     & \texttt{YYANWYY}~\cite{Javahery1994DNA} \\
        Motif Ten Element    & $+18$ to $+27$ & MTE     & \texttt{CSARCSSAACGS} \\
        Downstream Element   & $+28$ to $+32$ & DPE     & \texttt{RGWYV} \\
        \bottomrule
    \end{tabular}
    \begin{flushleft}
        \footnotesize
        \textit{Note:} \textbf{R}=Purine, \textbf{Y}=Pyrimidine, \textbf{W}=Weak (A/T), \textbf{S}=Strong (G/C), \textbf{N}=Any.
        \textbf{These canonical consensus sequences are defined primarily in \textit{Drosophila} and are often more degenerate in humans.}
    \end{flushleft}
\end{table}

\subsubsection{Interpretation Methodology}
To decode how distinct layers process biological information, we employ layer-specific visualization techniques tailored to the mathematical operations of each module.

\paragraph{Hyena: Norm-based Activation Profiling}
Since Hyena operators process long-range context via implicit convolutions, individual dimensions do not necessarily correspond to discrete features. Instead, we measure the \textbf{information density} at each position $t$ by computing the $L_2$ norm of the hidden state vector $h_t \in \mathbb{R}^d$:
\begin{equation}
    A_{\text{Hyena}}(t) = ||h_t||_2 = \sqrt{\sum_{i=1}^{d} (h_{t,i})^2}
\end{equation}
Peaks in $A_{\text{Hyena}}(t)$ indicate regions where the model aggregates significant global context, highlighting semantically rich sequence segments.

\paragraph{Transformer: Attention Accumulation}
To understand syntactic relationships, we analyze the self-attention matrix $\alpha$. Rather than examining pairwise weights in isolation, we calculate the \textbf{Total Attention Received} for each token $j$ by summing the attention weights from all query positions $i$:
\begin{equation}
    S_{\text{Attn}}(j) = \sum_{i=1}^{L} \alpha_{i,j}
\end{equation}
Tokens with high $S_{\text{Attn}}(j)$ act as "Attention Magnets," representing structural pivots. \textbf{For visualization, $S_{\text{Attn}}$ is min-max normalized to the range $[0, 1]$ to highlight relative structural importance independent of absolute magnitude.}

\paragraph{CNN: Importance Ranking and Maximal Activation}
Given the large number of convolutional kernels, we first prioritize them based on their contribution to the model's decision. We compute the \textbf{gradient-weighted importance} for each kernel $k$ (following the Grad-CAM protocol) and select the top-$K$ filters with the highest total contribution. For these top-ranked filters, we identify the specific input sequence window $x_{[\hat{t}:\hat{t}+w]}$ that triggers the maximum activation:
\begin{equation}
    \hat{t} = \operatorname*{argmax}_{t} |(W_k * X)_t|
\end{equation}
By visualizing the subsequence at position $\hat{t}$, we reconstruct the specific DNA vocabulary (motifs) that the network deems most critical for prediction.

\subsubsection{Validation of Grammar Capture: A Layer-wise Analysis}
To interpret how the hybrid architecture identifies no-TATA promoters, we visualize the activation patterns across different layers (Figure~\ref{fig:layer_interpretation}). The analysis reveals a hierarchical processing strategy that transitions from global sequence context to localized motif integration.

\paragraph{Hyena (Layer 0): Global Contextual Embedding.}
The Hyena layer acts as a global semantic encoder. Its activation profile constitutes a continuous, non-sparse baseline across the entire sequence window. This broad pattern suggests that the long-convolution mechanism captures dispersed sequence properties, such as GC-content trends or structural propensities, establishing the global genomic context required for subsequent fine-grained validation.

\paragraph{Transformer (Layers 1--2): Spatial Syntax and Transition Mapping.}
The Transformer layers serve as syntactic validators, capturing the spatial transitions and anchor points critical for promoter architecture:
\begin{itemize}
    \item \textbf{Layer 1} displays sharp, sparse attention peaks that primarily target the downstream MTE/DPE zones. This suggests the layer is functioning as a positional anchor, marking key regulatory elements relative to the TSS.
    \item \textbf{Layer 2} exhibits a distinct shift in activation patterns centered at the TSS (red dashed line). Upstream of the TSS, the layer maintains a high-frequency, regular oscillation, likely characterizing the repetitive sequence background of the promoter region. In contrast, downstream of the TSS, the wave pattern expands and aligns with the coding/regulatory transition zones. This sharp bifurcation indicates that Layer 2 is mapping the \textbf{functional boundary} between the upstream regulatory environment and the downstream initiation complex, a key step in identifying the precise coordinate of transcription start.
\end{itemize}

\paragraph{CNN Stack (Layers 3--5): Local Feature Extraction and Integration.}
The CNN layers function as a multi-scale morphological decoder, specializing in the detection of localized regulatory motifs. Rather than processing global dependencies, this stack focuses on capturing the precise signatures of the promoter core:
\begin{itemize}
    \item \textbf{Localized Signal Detection:} Throughout the CNN stack, filters demonstrate high specificity for known biological landmarks. For instance, in Layer 3 and Layer 4, filters (e.g., Top 1) precisely align with the \textbf{Inr region} ($-2$ to $+5$) and the \textbf{MTE/DPE zones}, acting as computational sensors for these conserved DNA sequences.
    \item \textbf{Multi-element Coverage:} A notable observation in the deeper CNN layers (e.g., Layer 5) is the emergence of multi-aspect activation. The filter sets (Top 1, 2, and 5) collectively cover the entire span of the Inr-MTE-DPE architecture. This simultaneous activation suggests that the CNN stack integrates isolated motif signals into a coordinated functional unit, capturing the synergistic interaction required for accurate promoter prediction.
\end{itemize}

\subsection{LLM Prompts}
\label{Appendix:LLM-prompts}
\begin{promptbox}{LLM+RAG:}
You are an expert in machine learning architecture selection. Your task is to analyze a given machine learning task and predict the best architectures based on similar tasks and their performance.

\textbf{You have access to:}
\begin{enumerate}[leftmargin=*, nosep]
    \item A knowledge base of similar tasks with their characteristics.
    \item Performance data showing which architectures work best for each task.
\end{enumerate}

\vspace{5pt}
\textbf{Your goal is to:}
\begin{enumerate}[leftmargin=*, nosep]
    \item Analyze the given task description.
    \item Find the most similar tasks from the knowledge base.
    \item Recommend the best architectures based on performance data.
    \item Provide clear reasoning for your predictions.
\end{enumerate}

\vspace{5pt}
\hrule 
\vspace{5pt}

\textbf{Available Tasks in Knowledge Base:}
\begin{quote}
\small\texttt{\{retrieve datas\}}
\end{quote}

\textbf{Performance Data:}
\begin{quote}
\small\texttt{\{performance data\}}
\end{quote}

\vspace{5pt}
\textbf{Please analyze the query task and provide your recommendations in the following format:}

\begin{tcolorbox}[colback=white, colframe=gray!50, boxrule=0.5pt, arc=0mm, left=2pt, right=2pt, top=2pt, bottom=2pt, fontupper=\footnotesize\ttfamily]
Task Analysis:\\
- Dataset characteristics: [analyze the input task]\\
- Problem type: [classification/regression/etc.]\\
- Modality: [DNA/Protein/etc.]\\
- Key requirements: [identify important constraints]

Similar tasks identified:\\
- Task Index: [X] - [reasoning for similarity]\\
- Task Index: [Y] - [reasoning for similarity]

Architecture Recommendations:\\
1. BEST CHOICE: [architecture-experiment-model\_name] - [reasoning]\\
2. SECOND BEST: [architecture-experiment-model\_name] - [reasoning]\\
3. THIRD BEST: [architecture-experiment-model\_name] - [reasoning]

Reasoning:\\
Detailed explanation of why these architectures are recommended based on similar tasks
\end{tcolorbox}

\vspace{5pt}
\textbf{Instruction:} Focus on recommending architectures that have shown good performance on similar tasks. Use the exact architecture names from the performance data.
\end{promptbox}
\newpage

\subsection{Agents Prompts}
\label{Appendix:Agent-prompts}
\subsubsection{Analyst Agent}
\begin{promptbox}{Analyst Agent}
\textbf{Role Description:}
You are an Analyst Agent specialized in understanding machine learning tasks and datasets. Your \textbf{ONLY} responsibility is to analyze the user's input and extract key parameters for architecture search.

\textbf{Workflow:}
\begin{enumerate}[leftmargin=*, nosep]
    \item Analyze the user's dataset and task description.
    \item Extract and summarize the key information needed for architecture search.
    \item Identify the modality (DNA, Protein, etc.) and problem type.
    \item Provide a clear summary for the Retriever Agent to use for searching.
\end{enumerate}

\vspace{5pt}
\textbf{You should NOT:}
\begin{itemize}[leftmargin=*, nosep]
    \item Search for architectures yourself.
    \item Make recommendations.
    \item Call any search tools.
\end{itemize}

\vspace{5pt}
\textbf{Output Format (Response Template):}
\begin{tcolorbox}[colback=white, colframe=gray!50, boxrule=0.5pt, arc=0mm, left=2pt, right=2pt, top=2pt, bottom=2pt, fontupper=\footnotesize\ttfamily]
Task Summary:\\
- Dataset characteristics (size, type, format)\\
- Problem type (classification, regression, clustering, etc.)\\
- Modality (DNA, Protein, etc.)\\
- Key constraints and requirements\\
- Performance objectives

Search Parameters:\\
- Modality: [DNA/Protein/etc.]\\
- Problem Type: [classification/regression/etc.]\\
- Task Description: [Clear summary for search]

Context:\\
- Any additional context that might be relevant for architecture selection\\
- Special requirements or constraints
\end{tcolorbox}

\vspace{5pt}
\textbf{Instruction:} Format your output clearly for the Retriever Agent to process.
\end{promptbox}

\newpage
\subsubsection{Task Retriever Agent}
\begin{promptbox}{Task Retriever Agent}
\textbf{Role Description:}
You are a Task Retriever Agent. Your responsibility is to identify the most similar tasks or top \texttt{\{top\_k\}} tasks from the knowledge base using your natural language understanding capabilities. Must give task index.

\textbf{Workflow:}
\begin{enumerate}[leftmargin=*, nosep]
    \item Receive a message from the Analyst containing task summary and search parameters.
    \item Extract the task description, problem type, modality, and key characteristics from the Analyst's output.
    \item Use your LLM reasoning to identify which tasks in the knowledge base are most similar to the query task.
    \item Consider multiple dimensions of similarity:
    \begin{itemize}[nosep]
        \item Problem type (classification, regression, clustering, etc.)
        \item Modality (DNA, Protein, text, image, etc.)
        \item Dataset characteristics (size, complexity, format)
        \item Task objectives and constraints
        \item Domain-specific requirements
    \end{itemize}
    \item Select the top \texttt{\{top\_k\}} most relevant tasks based on your understanding.
    \item Format the retrieved similar tasks clearly for the Architecture Retriever Agent.
\end{enumerate}

\vspace{5pt}
\textbf{You MUST:}
\begin{itemize}[leftmargin=*, nosep]
    \item Use your natural language understanding to identify similar tasks.
    \item \textbf{NOT} search for architectures yourself - that's the Architecture Retriever's job.
    \item \textbf{NOT} perform any scoring or evaluation of architectures.
    \item Focus \textbf{ONLY} on task similarity identification using LLM reasoning.
    \item Consider semantic similarity, not just keyword matching.
    \item Return \textbf{EXACTLY} \texttt{\{top\_k\}} similar tasks (no more, no less).
\end{itemize}

\vspace{5pt}
\textbf{Output Structure Template:}
\begin{tcolorbox}[colback=white, colframe=gray!50, boxrule=0.5pt, arc=0mm, left=2pt, right=2pt, top=2pt, bottom=2pt, fontupper=\footnotesize\ttfamily]
Conclusion:\\
Task Index: [Index]

Task similarity analysis:\\
- Query Task Description: [from Analyst]\\
- Analysis Strategy: Using LLM reasoning to identify semantically similar tasks

Top similar tasks:

1. Task Index: [Index]\\
- Similarity Reasoning: [Your explanation of why this task is similar]\\
- Task Description: [Description from knowledge base]\\
- Problem Type: [Type from knowledge base]\\
- Modality: [Modality from knowledge base]\\
- Dataset Characteristics: [Details from knowledge base]\\
- Key Similarities: [What makes this task similar to the query]

(Continue for all \{top\_k\} similar tasks identified)

Summary:\\
Brief summary of the most relevant tasks identified and why they are good matches for the query task

Next step:\\
Pass this information to the Architecture Retriever Agent to find suitable architectures for these similar tasks.
\end{tcolorbox}
\end{promptbox}

\newpage
\subsubsection{Architecture Retriever Agent.} 
\begin{promptbox}{Architecture Retriever Agent}
\textbf{Role Description:}
You are an Architecture Retriever Agent. Your responsibility is to find suitable architectures based on the similar tasks identified by the Task Retriever.

\textbf{Workflow:}
\begin{enumerate}[leftmargin=*, nosep]
    \item Receive a message from the Task Retriever containing similar tasks and their details.
    \item \textbf{IMMEDIATELY} call the \texttt{architecture\_retrieval\_tool} with the Task Retriever's output text.
    \item The tool will automatically:
    \begin{itemize}[nosep]
        \item Extract task indices from the Task Retriever's output.
        \item Look up architectures that were successful on the identified similar tasks.
        \item Return detailed architecture information including performance metrics.
        \item Provide architecture descriptions, performance summaries, and statistics.
    \end{itemize}
    \item Return the tool's result directly - do not format or modify it.
\end{enumerate}

\vspace{5pt}
\textbf{Critical Requirements:}
\begin{itemize}[leftmargin=*, nosep]
    \item You \textbf{MUST} call the tool and return its EXECUTION RESULT.
    \item \textbf{NEVER} return tool call parameters or JSON strings like \texttt{\{"name": "tool\_name", "parameters": \{...\}\}}.
    \item \textbf{NEVER} return the raw tool call format.
    \item \textbf{ALWAYS} return the actual tool execution output.
\end{itemize}

\vspace{5pt}
\textbf{You MUST:}
\begin{itemize}[leftmargin=*, nosep]
    \item \textbf{IMMEDIATELY} call \texttt{architecture\_retrieval\_tool} with the Task Retriever's output text.
    \item Return the tool's execution result directly (the actual data, not the call format).
    \item \textbf{NOT} perform any scoring or evaluation yourself - that's the Reviewer's job.
    \item \textbf{NOT} search for tasks - that's the Task Retriever's job.
    \item \textbf{NOT} format or modify the tool's output.
    \item Focus \textbf{ONLY} on architecture retrieval based on task performance.
\end{itemize}

\vspace{5pt}
\textbf{Example of correct behavior:}
\begin{itemize}[leftmargin=*, nosep]
    \item Input: Task Retriever output with task indices.
    \item Action: Call \texttt{architecture\_retrieval\_tool(input\_text=task\_retriever\_output)}.
    \item Output: The actual JSON result from the tool execution (not the tool call format).
\end{itemize}

\vspace{5pt}
\textbf{Important:} Simply call the tool and return its result. Do not add any additional formatting or text.
\end{promptbox}

\newpage
\subsubsection{Predictor Agent}
\begin{promptbox}{Predictor Agent}
\textbf{Role Description:}
You are a Predictor Agent responsible for evaluating ML architectures and creating final recommendations.

\textbf{Workflow:}
\begin{enumerate}[leftmargin=*, nosep]
    \item Receive a message from the Architecture Retriever Agent containing detailed information for each architecture and the overall problem type.
    \item For each architecture, you \textbf{MUST} call the \texttt{evaluate\_architecture} function with the \texttt{architecture\_name} and \texttt{problem\_type}.
    \item Consolidate the scores returned by the \texttt{evaluate\_architecture} tool.
    \item Analyze the scored architectures and select top \texttt{\{recommendation\_count\}} recommendations based on scores.
    \item Provide detailed reasoning for each recommendation.
\end{enumerate}

\vspace{5pt}
\textbf{Critical Output Format Requirements:}
\begin{itemize}[leftmargin=*, nosep]
    \item Architecture names \textbf{MUST} be in the exact format...
    \item Examples: \texttt{mix-kmer1-path\_54}, \texttt{cnn-kmer1-path\_12}, \texttt{transformer-kmer1-path\_8}.
    \item Use \textbf{EXACT} architecture names from the Architecture Retriever output.
    \item Do \textbf{NOT} modify or abbreviate architecture names.
\end{itemize}

\vspace{5pt}
\textbf{Your output format is strict (Template):}
\begin{tcolorbox}[colback=white, colframe=gray!50, boxrule=0.5pt, arc=0mm, left=2pt, right=2pt, top=2pt, bottom=2pt, fontupper=\footnotesize\ttfamily]
Executive summary\\
Brief overview of the analysis and key findings.

Top recommendations

1. Best performance\\
- Architecture: [EXACT\_ARCHITECTURE\_NAME]

2. Second best\\
- Architecture: [EXACT\_ARCHITECTURE\_NAME]

3. Third best\\
- Architecture: [EXACT\_ARCHITECTURE\_NAME]

(Continue for all \{recommendation\_count\} recommendations if more than 3)

Detailed Analysis\\
Architecture: [EXACT\_ARCHITECTURE\_NAME]\\
- Justification: [Justification from evaluate\_architecture tool]

(Continue for all architectures)

Summary:\\
Briefly summarize the findings from the scoring tool and highlight the key trade-offs identified.

Analysis Complete - Ready for implementation!
\end{tcolorbox}

\vspace{5pt}
\textbf{Important:} Always use the EXACT architecture names from the Architecture Retriever output. Do not modify, abbreviate, or change the format of architecture names.
\end{promptbox}

%% file: icml2026.bib
@misc{devlin2019bertpretrainingdeepbidirectional,
	title        = {BERT: Pre-training of Deep Bidirectional Transformers for Language Understanding},
	author       = {Jacob Devlin and Ming-Wei Chang and Kenton Lee and Kristina Toutanova},
	year         = 2019,
	url          = {https://arxiv.org/abs/1810.04805},
	eprint       = {1810.04805},
	archiveprefix = {arXiv},
	primaryclass = {cs.CL}
}

@article{openai2023gpt,
	title        = {GPT-4 technical report},
	author       = {OpenAI, R},
	year         = 2023,
	journal      = {arXiv},
	pages        = {2303--08774}
}

@misc{touvron2023llamaopenefficientfoundation,
	title        = {LLaMA: Open and Efficient Foundation Language Models},
	author       = {Hugo Touvron and Thibaut Lavril and Gautier Izacard and Xavier Martinet and Marie-Anne Lachaux and Timothée Lacroix and Baptiste Rozière and Naman Goyal and Eric Hambro and Faisal Azhar and Aurelien Rodriguez and Armand Joulin and Edouard Grave and Guillaume Lample},
	year         = 2023,
	url          = {https://arxiv.org/abs/2302.13971},
	eprint       = {2302.13971},
	archiveprefix = {arXiv},
	primaryclass = {cs.CL}
}

@article{dosovitskiy2020image,
	title        = {An image is worth 16x16 words: Transformers for image recognition at scale},
	author       = {Dosovitskiy, Alexey},
	year         = 2020,
	journal      = {arXiv preprint arXiv:2010.11929}
}

@article{liu2024visual,
	title        = {Visual instruction tuning},
	author       = {Liu, Haotian and Li, Chunyuan and Wu, Qingyang and Lee, Yong Jae},
	year         = 2024,
	journal      = {Advances in neural information processing systems},
	volume       = 36
}

@article{vaswani2017attention,
	title        = {Attention is all you need},
	author       = {Vaswani, Ashish and Shazeer, Noam and Parmar, Niki and Uszkoreit, Jakob and Jones, Llion and Gomez, Aidan N and Kaiser, {\L}ukasz and Polosukhin, Illia},
	year         = 2017,
	journal      = {Advances in neural information processing systems},
	volume       = 30
}

@misc{ho2020denoisingdiffusionprobabilisticmodels,
      title={Denoising Diffusion Probabilistic Models}, 
      author={Jonathan Ho and Ajay Jain and Pieter Abbeel},
      year={2020},
      eprint={2006.11239},
      archivePrefix={arXiv},
      primaryClass={cs.LG},
      url={https://arxiv.org/abs/2006.11239}, 
}

@misc{xiao2025proteinlargelanguagemodels,
      title={Protein Large Language Models: A Comprehensive Survey}, 
      author={Yijia Xiao and Wanjia Zhao and Junkai Zhang and Yiqiao Jin and Han Zhang and Zhicheng Ren and Renliang Sun and Haixin Wang and Guancheng Wan and Pan Lu and Xiao Luo and Yu Zhang and James Zou and Yizhou Sun and Wei Wang},
      year={2025},
      eprint={2502.17504},
      archivePrefix={arXiv},
      primaryClass={q-bio.BM},
      url={https://arxiv.org/abs/2502.17504}, 
}

@misc{zhou2024dnabert2efficientfoundationmodel,
      title={DNABERT-2: Efficient Foundation Model and Benchmark For Multi-Species Genome}, 
      author={Zhihan Zhou and Yanrong Ji and Weijian Li and Pratik Dutta and Ramana Davuluri and Han Liu},
      year={2024},
      eprint={2306.15006},
      archivePrefix={arXiv},
      primaryClass={q-bio.GN},
      url={https://arxiv.org/abs/2306.15006}, 
}

@article{
    doi:10.1073/pnas.2016239118,
    author = {Alexander Rives  and Joshua Meier  and Tom Sercu  and Siddharth Goyal  and Zeming Lin  and Jason Liu  and Demi Guo  and Myle Ott  and C. Lawrence Zitnick  and Jerry Ma  and Rob Fergus },
    title = {Biological structure and function emerge from scaling unsupervised learning to 250 million protein sequences},
    journal = {Proceedings of the National Academy of Sciences},
    volume = {118},
    number = {15},
    pages = {e2016239118},
    year = {2021},
    doi = {10.1073/pnas.2016239118},
    URL = {https://www.pnas.org/doi/abs/10.1073/pnas.2016239118},
    eprint = {https://www.pnas.org/doi/pdf/10.1073/pnas.2016239118},
}

@article{dill2012protein,
  title={The protein-folding problem, 50 years on},
  author={Dill, Ken A. and MacCallum, Jeffrey L.},
  journal={Science},
  volume={338},
  number={6110},
  pages={1042--1046},
  year={2012},
  publisher={American Association for the Advancement of Science},
  doi={10.1126/science.1219021}
}

@article{Wang_2017,
   title={Accurate De Novo Prediction of Protein Contact Map by Ultra-Deep Learning Model},
   volume={13},
   ISSN={1553-7358},
   url={http://dx.doi.org/10.1371/journal.pcbi.1005324},
   DOI={10.1371/journal.pcbi.1005324},
   number={1},
   journal={PLOS Computational Biology},
   publisher={Public Library of Science (PLoS)},
   author={Wang, Sheng and Sun, Siqi and Li, Zhen and Zhang, Renyu and Xu, Jinbo},
   editor={Schlessinger, Avner},
   year={2017},
   month=jan, pages={e1005324} }

@article{jumper2021highly,
  title={Highly accurate protein structure prediction with {AlphaFold}},
  author={Jumper, John and Evans, Richard and Pritzel, Alexander and Green, Tim and Figurnov, Michael and Ronneberger, Olaf and Tunyasuvunakool, Kathryn and Bates, Russ and {\v{Z}}{\'\i}dek, Augustin and Potapenko, Anna and others},
  journal={Nature},
  volume={596},
  number={7873},
  pages={583--589},
  year={2021},
  publisher={Nature Publishing Group}
}

@misc{dip2024patholmidentifyingpathogenicitydna,
      title={PathoLM: Identifying pathogenicity from the DNA sequence through the Genome Foundation Model}, 
      author={Sajib Acharjee Dip and Uddip Acharjee Shuvo and Tran Chau and Haoqiu Song and Petra Choi and Xuan Wang and Liqing Zhang},
      year={2024},
      eprint={2406.13133},
      archivePrefix={arXiv},
      primaryClass={cs.CL},
      url={https://arxiv.org/abs/2406.13133}, 
}

@article{Wang2024Multi,
  author    = {Wang, Ning and Bian, Jialu and Li, Yutong and Wang, Donghan and Duan, Junbo and Wan, Yue and Wang, Jike and Ma, Jianzhu and Li, Yi-fan},
  title     = {Multi-purpose {RNA} language modelling with motif-aware pretraining and type-guided fine-tuning},
  journal   = {Nature Machine Intelligence},
  volume    = {6},
  number    = {5},
  pages     = {548--557},
  year      = {2024},
  publisher = {Springer Nature},
  doi       = {10.1038/s42256-024-00836-4},
  url       = {https://doi.org/10.1038/s42256-024-00836-4}
}

@misc{zoph2017neuralarchitecturesearchreinforcement,
      title={Neural Architecture Search with Reinforcement Learning}, 
      author={Barret Zoph and Quoc V. Le},
      year={2017},
      eprint={1611.01578},
      archivePrefix={arXiv},
      primaryClass={cs.LG},
      url={https://arxiv.org/abs/1611.01578}, 
}

@misc{real2017largescaleevolutionimageclassifiers,
      title={Large-Scale Evolution of Image Classifiers}, 
      author={Esteban Real and Sherry Moore and Andrew Selle and Saurabh Saxena and Yutaka Leon Suematsu and Jie Tan and Quoc Le and Alex Kurakin},
      year={2017},
      eprint={1703.01041},
      archivePrefix={arXiv},
      primaryClass={cs.NE},
      url={https://arxiv.org/abs/1703.01041}, 
}

@misc{liu2019dartsdifferentiablearchitecturesearch,
      title={DARTS: Differentiable Architecture Search}, 
      author={Hanxiao Liu and Karen Simonyan and Yiming Yang},
      year={2019},
      eprint={1806.09055},
      archivePrefix={arXiv},
      primaryClass={cs.LG},
      url={https://arxiv.org/abs/1806.09055}, 
}

@misc{tan2020efficientnetrethinkingmodelscaling,
      title={EfficientNet: Rethinking Model Scaling for Convolutional Neural Networks}, 
      author={Mingxing Tan and Quoc V. Le},
      year={2020},
      eprint={1905.11946},
      archivePrefix={arXiv},
      primaryClass={cs.LG},
      url={https://arxiv.org/abs/1905.11946}, 
}

@inproceedings{deng2009imagenet,
  title={ImageNet: A Large-Scale Hierarchical Image Database},
  author={Deng, Jia and Dong, Wei and Socher, Richard and Li, Li-Jia and Li, Kai and Fei-Fei, Li},
  booktitle={2009 IEEE Conference on Computer Vision and Pattern Recognition (CVPR)},
  pages={248--255},
  year={2009},
  organization={IEEE}
}

@article{SIVANGI202263,
title = {NoAS-DS: Neural optimal architecture search for detection of diverse DNA signals},
journal = {Neural Networks},
volume = {147},
pages = {63-71},
year = {2022},
issn = {0893-6080},
doi = {https://doi.org/10.1016/j.neunet.2021.12.009},
url = {https://www.sciencedirect.com/science/article/pii/S0893608021004822},
author = {Kaushik Bhargav Sivangi and Chandra Mohan Dasari and Santhosh Amilpur and Raju Bhukya},
}

@misc{ma2019deepneuralarchitecturesearch,
      title={Deep Neural Architecture Search with Deep Graph Bayesian Optimization}, 
      author={Lizheng Ma and Jiaxu Cui and Bo Yang},
      year={2019},
      eprint={1905.06159},
      archivePrefix={arXiv},
      primaryClass={cs.LG},
      url={https://arxiv.org/abs/1905.06159}, 
}

@misc{white2021powerfulperformancepredictorsneural,
      title={How Powerful are Performance Predictors in Neural Architecture Search?}, 
      author={Colin White and Arber Zela and Binxin Ru and Yang Liu and Frank Hutter},
      year={2021},
      eprint={2104.01177},
      archivePrefix={arXiv},
      primaryClass={cs.LG},
      url={https://arxiv.org/abs/2104.01177}, 
}

@article{Suzek2007,
  author    = {Suzek, Baris E. and Huang, Hongzhan and McGarvey, Peter and Mazumder, Raja and Wu, Cathy H.},
  title     = {UniRef: comprehensive and non-redundant UniProt reference clusters},
  journal   = {Bioinformatics},
  year      = {2007},
  volume    = {23},
  number    = {10},
  pages     = {1282--1288},
  month     = {may},
  doi       = {10.1093/bioinformatics/btm098},
  pmid      = {17379688},
  issn      = {1367-4803}
}

@misc{xu2022peercomprehensivemultitaskbenchmark,
      title={PEER: A Comprehensive and Multi-Task Benchmark for Protein Sequence Understanding}, 
      author={Minghao Xu and Zuobai Zhang and Jiarui Lu and Zhaocheng Zhu and Yangtian Zhang and Chang Ma and Runcheng Liu and Jian Tang},
      year={2022},
      eprint={2206.02096},
      archivePrefix={arXiv},
      primaryClass={cs.LG},
      url={https://arxiv.org/abs/2206.02096}, 
}

@misc{guo2020singlepathoneshotneural,
      title={Single Path One-Shot Neural Architecture Search with Uniform Sampling}, 
      author={Zichao Guo and Xiangyu Zhang and Haoyuan Mu and Wen Heng and Zechun Liu and Yichen Wei and Jian Sun},
      year={2020},
      eprint={1904.00420},
      archivePrefix={arXiv},
      primaryClass={cs.CV},
      url={https://arxiv.org/abs/1904.00420}, 
}

@misc{ying2019nasbench101,
      title={NAS-Bench-101: Towards Reproducible Neural Architecture Search},
      author={Chris Ying and Aaron Klein and Esteban Real and Eric Christiansen and Kevin Murphy and Frank Hutter},
      year={2019},
      eprint={1902.09635},
      archivePrefix={arXiv},
      primaryClass={cs.LG},
      url={https://arxiv.org/abs/1902.09635},
}

@misc{dudziak2021brpnaspredictionbasednasusing,
      title={BRP-NAS: Prediction-based NAS using GCNs}, 
      author={Łukasz Dudziak and Thomas Chau and Mohamed S. Abdelfattah and Royson Lee and Hyeji Kim and Nicholas D. Lane},
      year={2021},
      eprint={2007.08668},
      archivePrefix={arXiv},
      primaryClass={cs.LG},
      url={https://arxiv.org/abs/2007.08668}, 
}

@misc{duan2021transnasbench101improvingtransferabilitygeneralizability,
      title={TransNAS-Bench-101: Improving Transferability and Generalizability of Cross-Task Neural Architecture Search}, 
      author={Yawen Duan and Xin Chen and Hang Xu and Zewei Chen and Xiaodan Liang and Tong Zhang and Zhenguo Li},
      year={2021},
      eprint={2105.11871},
      archivePrefix={arXiv},
      primaryClass={cs.CV},
      url={https://arxiv.org/abs/2105.11871}, 
}

@article{Sapoval2022,
  author    = {Sapoval, Nina and Aghazadeh, Ava and Nute, Michael G. and Moore, Fletcher H. and Gob Biswas, Shaurya D. and Wang, Michael F. Z. and AlQuraishi, Mohammed},
  title     = {Current progress and open challenges for applying deep learning across the biosciences},
  journal   = {Nature Communications},
  year      = {2022},
  volume    = {13},
  number    = {1},
  pages     = {1728},
  doi       = {10.1038/s41467-022-29268-7},
  url       = {https://doi.org/10.1038/s41467-022-29268-7}
}

@article{Greener2022guide,
  author    = {Greener, Joseph G. and Kandathil, Shaun M. and Moffat, Lewis and Jones, David T.},
  title     = {A guide to machine learning for biologists},
  journal   = {Nature Reviews Molecular Cell Biology},
  year      = {2022},
  volume    = {23},
  number    = {1},
  pages     = {40--55},
  month     = {Jan},
  doi       = {10.1038/s41580-021-00407-0},
  url       = {https://doi.org/10.1038/s41580-021-00407-0}
}

@article{Eisenstein2024foundation,
  author    = {Eisenstein, Michael},
  title     = {Foundation models build on {ChatGPT} tech to learn the fundamental language of biology},
  journal   = {Nature Biotechnology},
  year      = {2024},
  volume    = {42},
  number    = {9},
  pages     = {1323--1325},
  month     = {Sep},
  doi       = {10.1038/s41587-024-02400-2},
  url       = {https://doi.org/10.1038/s41587-024-02400-2}
}

@article {Vishniakov2024.12.18.628606,
	author = {Vishniakov, Kirill and Viswanathan, Karthik and Medvedev, Aleksandr and Kanithi, Praveenkumar and Pimentel, Marco AF and Rajan, Ronnie and Khan, Shadab},
	title = {Genomic Foundationless Models: Pretraining Does Not Promise Performance},
	elocation-id = {2024.12.18.628606},
	year = {2025},
	doi = {10.1101/2024.12.18.628606},
	publisher = {Cold Spring Harbor Laboratory},
	URL = {https://www.biorxiv.org/content/early/2025/06/25/2024.12.18.628606},
	eprint = {https://www.biorxiv.org/content/early/2025/06/25/2024.12.18.628606.full.pdf},
	journal = {bioRxiv}
}

@inproceedings{Bender2018UnderstandingAS,
  title={Understanding and Simplifying One-Shot Architecture Search},
  author={Gabriel Bender and Pieter-Jan Kindermans and Barret Zoph and Vijay Vasudevan and Quoc V. Le},
  booktitle={International Conference on Machine Learning},
  year={2018},
  url={https://api.semanticscholar.org/CorpusID:51763580}
}

@inproceedings{krizhevsky2012imagenet,
  title={Imagenet classification with deep convolutional neural networks},
  author={Krizhevsky, Alex and Sutskever, Ilya and Hinton, Geoffrey E},
  booktitle={Advances in neural information processing systems},
  volume={25},
  year={2012}
}

@article{hochreiter1997long,
  title={Long short-term memory},
  author={Hochreiter, Sepp and Schmidhuber, J{\"u}rgen},
  journal={Neural Computation},
  volume={9},
  number={8},
  pages={1735--1780},
  year={1997},
  publisher={MIT Press}
}

@inproceedings{gu2024mamba,
  title={Mamba: Linear-Time Sequence Modeling with Selective State Spaces},
  author={Gu, Albert and Dao, Tri},
  booktitle={The Twelfth International Conference on Learning Representations (ICLR)},
  year={2024},
  url={https://openreview.net/forum?id=AL1Oo7_aKC}
}

@inproceedings{poli2023hyena,
  title={Hyena Hierarchy: Towards Larger Convolutional Language Models},
  author={Poli, Michael and Massaroli, Stefano and Nguyen, Eric and Fu, Daniel Y. and Dao, Tri and Baccus, Stephen A. and Yoshimoto, Yoshua and Cl{\'e}ment, Christopher and L{\'o}pez-Paz, David and R{\'e}, Christopher},
  booktitle={International Conference on Machine Learning (ICML)},
  pages={27926--27950},
  year={2023},
  organization={PMLR}
}

@article {Dalla-Torre2023.01.11.523679,
	author = {Dalla-Torre, Hugo and Gonzalez, Liam and Mendoza-Revilla, Javier and Carranza, Nicolas Lopez and Grzywaczewski, Adam Henryk and Oteri, Francesco and Dallago, Christian and Trop, Evan and Sirelkhatim, Hassan and Richard, Guillaume and Skwark, Marcin and Beguir, Karim and Lopez, Marie and Pierrot, Thomas},
	title = {The Nucleotide Transformer: Building and Evaluating Robust Foundation Models for Human Genomics},
	elocation-id = {2023.01.11.523679},
	year = {2023},
	doi = {10.1101/2023.01.11.523679},
	publisher = {Cold Spring Harbor Laboratory},
	eprint = {https://www.biorxiv.org/content/early/2023/03/09/2023.01.11.523679.full.pdf},
	journal = {bioRxiv}
}

@misc{li2024vqdnaunleashingpowervector,
      title={VQDNA: Unleashing the Power of Vector Quantization for Multi-Species Genomic Sequence Modeling}, 
      author={Siyuan Li and Zedong Wang and Zicheng Liu and Di Wu and Cheng Tan and Jiangbin Zheng and Yufei Huang and Stan Z. Li},
      year={2024},
      eprint={2405.10812},
      archivePrefix={arXiv},
      primaryClass={q-bio.GN},
      url={https://arxiv.org/abs/2405.10812}, 
}

@article {Elnaggar2020.07.12.199554,
	author = {Elnaggar, Ahmed and Heinzinger, Michael and Dallago, Christian and Rehawi, Ghalia and Wang, Yu and Jones, Llion and Gibbs, Tom and Feher, Tamas and Angerer, Christoph and Steinegger, Martin and Bhowmik, Debsindhu and Rost, Burkhard},
	title = {ProtTrans: Towards Cracking the Language of Life{\textquoteright}s Code Through Self-Supervised Learning},
	elocation-id = {2020.07.12.199554},
	year = {2021},
	doi = {10.1101/2020.07.12.199554},
	URL = {https://www.biorxiv.org/content/early/2021/05/04/2020.07.12.199554},
	eprint = {https://www.biorxiv.org/content/early/2021/05/04/2020.07.12.199554.full.pdf},
	journal = {bioRxiv}
}

@misc{Microsoft_Neural_Network_Intelligence_2021,
author = {{Microsoft}},
month = jan,
title = {{Neural Network Intelligence}},
url = {https://github.com/microsoft/nni},
version = {2.0},
year = {2021}
}

@article{marcos2017principles,
  title     = {Principles for designing proteins with cavities formed by curved {β}-sheets},
  author    = {Marcos, Enrique and Basanta, Benjamin and Chidyausiku, Tamuka M and Tang, Yuefeng and Oberdorfer, Gustav and Liu, Gaohua and Swapna, GVT and Guan, Rongjin and Silva, Daniel-Adriano and Dou, Jiayi and Pereira, Jose Henrique and Xiao, Rong and Sankaran, Banumathi and Zwart, Peter H and Montelione, Gaetano T and Baker, David},
  journal   = {Science},
  volume    = {355},
  number    = {6321},
  pages     = {201--206},
  year      = {2017},
  publisher = {American Association for the Advancement of Science},
  doi       = {10.1126/science.aah7383},
  pmid      = {28082595},
  pmcid     = {PMC5588894}
}

@article{10.1093/nargab/lqac012,
    author = {Akiyama, Manato and Sakakibara, Yasubumi},
    title = {Informative RNA base embedding for RNA structural alignment and clustering by deep representation learning},
    journal = {NAR Genomics and Bioinformatics},
    volume = {4},
    number = {1},
    pages = {lqac012},
    year = {2022},
    month = {02},
    issn = {2631-9268},
    doi = {10.1093/nargab/lqac012},
    url = {https://doi.org/10.1093/nargab/lqac012},
    eprint = {https://academic.oup.com/nargab/article-pdf/4/1/lqac012/42577168/lqac012.pdf},
}

@article{gunduz2024optimized,
  author    = {G{\"u}ndüz, H. A. and Mreches, R. and Moosbauer, J. and Robertson, G. and To, X. Y. and Franzosa, E. A. and Huttenhower, C. and Rezaei, M. and McHardy, A. C. and Bischl, B. and M{\"u}nch, P. C. and Binder, M.},
  title     = {Optimized model architectures for deep learning on genomic data},
  journal   = {Communications Biology},
  year      = {2024},
  volume    = {7},
  number    = {1},
  pages     = {516},
  month     = {Apr},
  doi       = {10.1038/s42003-024-06161-1},
  pmid      = {38693292},
  pmcid     = {PMC11063068},
  note      = {Erratum in: Commun Biol. 2024 May 23;7(1):625. doi: 10.1038/s42003-024-06318-y}
}

@article {Zhang2021.02.25.432960,
	author = {Zhang, Zijun and Cofer, Evan M. and Troyanskaya, Olga G.},
	title = {AMBIENT: Accelerated Convolutional Neural Network Architecture Search for Regulatory Genomics},
	elocation-id = {2021.02.25.432960},
	year = {2021},
	doi = {10.1101/2021.02.25.432960},
	publisher = {Cold Spring Harbor Laboratory},
	URL = {https://www.biorxiv.org/content/early/2021/02/27/2021.02.25.432960},
	eprint = {https://www.biorxiv.org/content/early/2021/02/27/2021.02.25.432960.full.pdf},
	journal = {bioRxiv}
}

@article{zhang2021automated,
  title     = {An automated framework for efficiently designing deep convolutional neural networks in genomics},
  author    = {Zhang, Zhidie and Park, Chae Yeon and Theesfeld, Christopher L and others},
  journal   = {Nature Machine Intelligence},
  volume    = {3},
  number    = {5},
  pages     = {392--400},
  year      = {2021},
  month     = {May},
  publisher = {Nature Publishing Group UK London},
  doi       = {10.1038/s42256-021-00316-z}
}

@misc{sennrich2016neuralmachinetranslationrare,
      title={Neural Machine Translation of Rare Words with Subword Units}, 
      author={Rico Sennrich and Barry Haddow and Alexandra Birch},
      year={2016},
      eprint={1508.07909},
      archivePrefix={arXiv},
      primaryClass={cs.CL},
      url={https://arxiv.org/abs/1508.07909}, 
}

@article{10.1093/nar/gkm895,
    author = {The UniProt Consortium},
    title = {The Universal Protein Resource (UniProt)},
    journal = {Nucleic Acids Research},
    volume = {36},
    number = {suppl\_1},
    pages = {D190-D195},
    year = {2007},
    month = {11},
    issn = {0305-1048},
    doi = {10.1093/nar/gkm895},
    url = {https://doi.org/10.1093/nar/gkm895},
    eprint = {https://academic.oup.com/nar/article-pdf/36/suppl_1/D190/7633197/gkm895.pdf},
}

@article{steinegger2019protein,
  title={Protein-level assembly increases protein sequence recovery from metagenomic samples manyfold},
  author={Steinegger, Martin and Mirdita, Milot and S{\"o}ding, Johannes},
  journal={Nature Methods},
  volume={16},
  number={7},
  pages={603--606},
  year={2019},
  publisher={Nature Publishing Group},
  doi={10.1038/s41592-019-0437-4}
}

@article{Fishman2023.06.12.544594,
	author = {Fishman, Veniamin and Kuratov, Yuri and Petrov, Maxim and Shmelev, Aleksei and Shepelin, Denis and Chekanov, Nikolay and Kardymon, Olga and Burtsev, Mikhail},
	title = {GENA-LM: A Family of Open-Source Foundational Models for Long DNA Sequences},
	elocation-id = {2023.06.12.544594},
	year = {2023},
	doi = {10.1101/2023.06.12.544594},
	publisher = {Cold Spring Harbor Laboratory},
	URL = {https://www.biorxiv.org/content/early/2023/06/13/2023.06.12.544594},
	eprint = {https://www.biorxiv.org/content/early/2023/06/13/2023.06.12.544594.full.pdf},
	journal = {bioRxiv}
}

@article{Theodoris2023Transfer,
  author    = {Theodoris, Christos V. and Xiao, Linyi and Chopra, Anant and Chiorazzi, Michael H. and Retkwa, Reuben S. F. and Varnavides, Georgios and Hsieh, Evan M. and Vaziri, Sasan A. and Lee, Victoria S. and Ward, Casey W. and He, Jesse D. R. D. and Aalfs, Jeffrey R. and Fonoudi, Hananeh and Pico, Alexander and Gambhir, Sanjiv S. and Snyder, Michael P. and Pollard, Katherine A. and Srivastava, Deepak},
  title     = {Transfer learning enables predictions in network biology},
  journal   = {Nature},
  year      = {2023},
  month     = {Jun},
  volume    = {618},
  number    = {7995},
  pages     = {616--624},
  doi       = {10.1038/s41586-023-06139-9},
  url       = {https://doi.org/10.1038/s41586-023-06139-9}
}

@article{Avsec2021Effective,
  author  = {Avsec, {\v Z}an and Agarwal, Vikram and Visentin, Dario and Laji{\'c}, Sergej and Curk, Tomaz and Gjorgjieva, Elena},
  title   = {Effective gene expression prediction from sequence by integrating long-range interactions},
  journal = {Nature Methods},
  volume  = {18},
  number  = {10},
  pages   = {1196--1203},
  year    = {2021},
  doi     = {10.1038/s41592-021-01252-x},
  url     = {https://doi.org/10.1038/s41592-021-01252-x}
}

@article{10.1093/bib/bbae577,
    author = {Yu, Zhenhua and Liu, Furui and Li, Yang},
    title = {scTCA: a hybrid Transformer-CNN architecture for imputation and denoising of scDNA-seq data},
    journal = {Briefings in Bioinformatics},
    volume = {25},
    number = {6},
    pages = {bbae577},
    year = {2024},
    month = {11},
    issn = {1477-4054},
    doi = {10.1093/bib/bbae577},
    url = {https://doi.org/10.1093/bib/bbae577},
    eprint = {https://academic.oup.com/bib/article-pdf/25/6/bbae577/60580936/bbae577.pdf},
}

@article{Zeng2025CellFM,
  title={CellFM: a large-scale foundation model pre-trained on transcriptomics of 100 million human cells},
  author={Zeng, Y. and Xie, J. and Shangguan, N. and others},
  journal={Nat Commun},
  volume={16},
  pages={4679},
  year={2025},
  month={May},
  doi={10.1038/s41467-025-59926-5}
}

@article{
doi:10.1126/science.ade2574,
author = {Zeming Lin  and Halil Akin  and Roshan Rao  and Brian Hie  and Zhongkai Zhu  and Wenting Lu  and Nikita Smetanin  and Robert Verkuil  and Ori Kabeli  and Yaniv Shmueli  and Allan dos Santos Costa  and Maryam Fazel-Zarandi  and Tom Sercu  and Salvatore Candido  and Alexander Rives },
title = {Evolutionary-scale prediction of atomic-level protein structure with a language model},
journal = {Science},
volume = {379},
number = {6637},
pages = {1123-1130},
year = {2023},
doi = {10.1126/science.ade2574},
URL = {https://www.science.org/doi/abs/10.1126/science.ade2574},
eprint = {https://www.science.org/doi/pdf/10.1126/science.ade2574},
}

@article{
doi:10.1126/science.ado9336,
author = {Eric Nguyen  and Michael Poli  and Matthew G. Durrant  and Brian Kang  and Dhruva Katrekar  and David B. Li  and Liam J. Bartie  and Armin W. Thomas  and Samuel H. King  and Garyk Brixi  and Jeremy Sullivan  and Madelena Y. Ng  and Ashley Lewis  and Aaron Lou  and Stefano Ermon  and Stephen A. Baccus  and Tina Hernandez-Boussard  and Christopher Ré  and Patrick D. Hsu  and Brian L. Hie },
title = {Sequence modeling and design from molecular to genome scale with Evo},
journal = {Science},
volume = {386},
number = {6723},
pages = {eado9336},
year = {2024},
doi = {10.1126/science.ado9336},
URL = {https://www.science.org/doi/abs/10.1126/science.ado9336},
eprint = {https://www.science.org/doi/pdf/10.1126/science.ado9336}}

@article {Brixi2025.02.18.638918,
	author = {Brixi, Garyk and Durrant, Matthew G. and Ku, Jerome and Poli, Michael and Brockman, Greg and Chang, Daniel and Gonzalez, Gabriel A. and King, Samuel H. and Li, David B. and Merchant, Aditi T. and Naghipourfar, Mohsen and Nguyen, Eric and Ricci-Tam, Chiara and Romero, David W. and Sun, Gwanggyu and Taghibakshi, Ali and Vorontsov, Anton and Yang, Brandon and Deng, Myra and Gorton, Liv and Nguyen, Nam and Wang, Nicholas K. and Adams, Etowah and Baccus, Stephen A. and Dillmann, Steven and Ermon, Stefano and Guo, Daniel and Ilango, Rajesh and Janik, Ken and Lu, Amy X. and Mehta, Reshma and Mofrad, Mohammad R.K. and Ng, Madelena Y. and Pannu, Jaspreet and R{\'e}, Christopher and Schmok, Jonathan C. and John, John St. and Sullivan, Jeremy and Zhu, Kevin and Zynda, Greg and Balsam, Daniel and Collison, Patrick and Costa, Anthony B. and Hernandez-Boussard, Tina and Ho, Eric and Liu, Ming-Yu and McGrath, Thomas and Powell, Kimberly and Burke, Dave P. and Goodarzi, Hani and Hsu, Patrick D. and Hie, Brian L.},
	title = {Genome modeling and design across all domains of life with Evo 2},
	elocation-id = {2025.02.18.638918},
	year = {2025},
	doi = {10.1101/2025.02.18.638918},
	publisher = {Cold Spring Harbor Laboratory},
	URL = {https://www.biorxiv.org/content/early/2025/02/21/2025.02.18.638918},
	eprint = {https://www.biorxiv.org/content/early/2025/02/21/2025.02.18.638918.full.pdf},
	journal = {bioRxiv}
}

@article{Javahery1994DNA,
  title     = {DNA sequence requirements for transcriptional initiator activity in mammalian cells},
  author    = {Javahery, Ramin and Khachi, Anahid and Lo, Kenneth and Zenzie-Gregory, Barbara and Smale, Stephen T.},
  journal   = {Mol Cell Biol},
  volume    = {14},
  number    = {1},
  pages     = {116--127},
  year      = {1994},
  doi       = {10.1128/mcb.14.1.116-127.1994}
}

@article{Burke1997DPE,
  title     = {The downstream core promoter element, {DPE}, is conserved from {Drosophila} to humans and is recognized by {TAFII60} of {Drosophila}},
  author    = {Burke, Thomas W. and Kadonaga, James T.},
  journal   = {Genes Dev},
  volume    = {11},
  number    = {22},
  pages     = {3020--3031},
  year      = {1997},
  month     = {Nov},
  doi       = {10.1101/gad.11.22.3020},
  pmid      = {9367984},
  pmcid     = {PMC316699}
}

@article{deaton2011cpg,
  title={CpG islands and the regulation of transcription},
  author={Deaton, Ailsa M and Bird, Adrian},
  journal={Genes \& development},
  volume={25},
  number={10},
  pages={1010--1022},
  year={2011}
}

@article{burke1996dpe,
  title={Drosophila TFIID binds to a conserved downstream basal promoter element that is present in many TATA-box-deficient promoters},
  author={Burke, T. W. and Kadonaga, J. T.},
  journal={Genes Dev.},
  volume={10},
  number={6},
  pages={711--724},
  year={1996},
  month={Mar},
  doi={10.1101/gad.10.6.711},
  note={PMID: 8598298}
}

@article{kutach2000dpe,
  title={The downstream promoter element DPE appears to be as widely used as the TATA box in Drosophila core promoters},
  author={Kutach, Alan K and Kadonaga, James T},
  journal={Molecular and Cellular Biology},
  volume={20},
  number={13},
  pages={4754--4764},
  year={2000}
}

@misc{lan2020albertlitebertselfsupervised,
      title={ALBERT: A Lite BERT for Self-supervised Learning of Language Representations}, 
      author={Zhenzhong Lan and Mingda Chen and Sebastian Goodman and Kevin Gimpel and Piyush Sharma and Radu Soricut},
      year={2020},
      eprint={1909.11942},
      archivePrefix={arXiv},
      primaryClass={cs.CL},
      url={https://arxiv.org/abs/1909.11942}, 
}

@article{ren2020identifying,
  title={Identifying viruses from metagenomic data using deep learning},
  author={Ren, J. and Song, K. and Deng, C. and Ahlgren, N. A. and Fuhrman, J. A. and Li, Y. and Xie, X. and Poplin, R. and Sun, F.},
  journal={Quantitative Biology},
  volume={8},
  number={1},
  pages={64--77},
  year={2020},
  month={Mar},
  publisher={Springer},
  doi={10.1007/s40484-019-0187-4},
  pmid={34084563},
  pmcid={PMC8172088}
}

@misc{li2021bossnasexploringhybridcnntransformers,
      title={BossNAS: Exploring Hybrid CNN-transformers with Block-wisely Self-supervised Neural Architecture Search}, 
      author={Changlin Li and Tao Tang and Guangrun Wang and Jiefeng Peng and Bing Wang and Xiaodan Liang and Xiaojun Chang},
      year={2021},
      eprint={2103.12424},
      archivePrefix={arXiv},
      primaryClass={cs.CV},
      url={https://arxiv.org/abs/2103.12424},
}

@inproceedings{schiff2024caduceus,
      title={Caduceus: Bi-Directional Equivariant Long-Range DNA Sequence Modeling},
      author={Schiff, Yair and Kao, Chia-Hsiang and Gokaslan, Aaron and Dao, Tri and Gu, Albert and Kuleshov, Volodymyr},
      booktitle={Proceedings of the 41st International Conference on Machine Learning (ICML)},
      year={2024},
      eprint={2403.03234},
      archivePrefix={arXiv},
      primaryClass={q-bio.GN},
      url={https://arxiv.org/abs/2403.03234},
}

@misc{wu2025generator,
      title={GENERator: A Long-Context Generative Genomic Foundation Model},
      author={Wu, Wei and Li, Qiuyi and Li, Mingyang and Fu, Kun and Feng, Fuli and Ye, Jieping and Xiong, Hui and Wang, Zheng},
      year={2025},
      eprint={2502.07272},
      archivePrefix={arXiv},
      primaryClass={cs.LG},
      url={https://arxiv.org/abs/2502.07272},
}

@article{gresova2023genomic,
      title={Genomic benchmarks: a collection of datasets for genomic sequence classification},
      author={Gre{\v{s}}ov{\'a}, Katar{\'i}na and Martinek, Vlastimil and {\v{C}}ech{\'a}k, David and {\v{S}}ime{\v{c}}ek, Petr and Alexiou, Panagiotis},
      journal={BMC Genomic Data},
      volume={24},
      number={1},
      pages={25},
      year={2023},
      publisher={Springer},
      doi={10.1186/s12863-023-01123-8},
}
